\documentclass{ieeeaccess}
\usepackage{cite}
\usepackage{amsmath,amssymb,amsfonts}
\usepackage{algorithmic}
\usepackage{graphicx}
\usepackage{textcomp}

\usepackage{bm}
\makeatletter
\AtBeginDocument{\DeclareMathVersion{bold}
\SetSymbolFont{operators}{bold}{T1}{times}{b}{n}
\SetSymbolFont{NewLetters}{bold}{T1}{times}{b}{it}
\SetMathAlphabet{\mathrm}{bold}{T1}{times}{b}{n}
\SetMathAlphabet{\mathit}{bold}{T1}{times}{b}{it}
\SetMathAlphabet{\mathbf}{bold}{T1}{times}{b}{n}
\SetMathAlphabet{\mathtt}{bold}{OT1}{pcr}{b}{n}
\SetSymbolFont{symbols}{bold}{OMS}{cmsy}{b}{n}
\renewcommand\boldmath{\@nomath\boldmath\mathversion{bold}}}
\makeatother

\def\BibTeX{{\rm B\kern-.05em{\sc i\kern-.025em b}\kern-.08em
    T\kern-.1667em\lower.7ex\hbox{E}\kern-.125emX}}

\usepackage[utf8]{inputenc} 
\usepackage{amsmath}        
\usepackage{amssymb}        
\usepackage{amsfonts}       
\usepackage{multirow}       
\usepackage{comment}        
\usepackage{booktabs}       
\usepackage{enumitem}       
\usepackage{url}            
\usepackage{pifont}         
\usepackage{color,soul}     
\usepackage[table]{xcolor}  

\begin{document}
\history{\vspace{12pt}}
\doi{10.1109/ACCESS.2025.3581035}

\title{Sim-to-Real Transfer of Deep Reinforcement Learning Agents for Online Coverage Path Planning}
\author{Arvi Jonnarth\authorrefmark{1} \IEEEmembership{Member, IEEE},
Ola Johansson\authorrefmark{2}, Jie Zhao\authorrefmark{3}, and Michael Felsberg\authorrefmark{2,4}
\IEEEmembership{Senior member, IEEE}}

\address[1]{Manta Systems, Linköping, Sweden (e-mail: arvi.jonnarth@ieee.org)}
\address[2]{Department of Electrical Engineering, Linköping University, Sweden (e-mail: ola.johansson@liu.se and michael.felsberg@liu.se)}
\address[3]{Department of Information and Communication Engineering, Dalian University of Technology, China (e-mail: zj982853200@dlut.edu.cn)}
\address[4]{School of Engineering, University of KwaZulu-Natal, Durban, South Africa}

\tfootnote{During this research, Arvi Jonnarth was an industrial PhD student at Linköping University and Husqvarna Group. This work was partially supported by the Wallenberg AI, Autonomous Systems and Software Program (WASP), funded by the Knut and Alice Wallenberg (KAW) Foundation. The work was funded in part by the Vinnova project, human-centered autonomous regional airport, Dnr 2022-02678, and by the strategic research environment, ELLIIT, funded by the Swedish government. The computational resources were provided by the National Academic Infrastructure for Supercomputing in Sweden (NAISS), partially funded by the Swedish Research Council, grant agreement no.\ 2022-06725, and by the Berzelius resource, provided by the KAW Foundation at the National Supercomputer Centre (NSC).}

\markboth
{Jonnarth \headeretal: Sim-to-Real Transfer of Deep Reinforcement Learning Agents for Online Coverage Path Planning}
{Jonnarth \headeretal: Sim-to-Real Transfer of Deep Reinforcement Learning Agents for Online Coverage Path Planning}

\corresp{Corresponding author: Arvi Jonnarth (e-mail: arvi.jonnarth@ieee.org).}

\begin{abstract}
Coverage path planning (CPP) is the problem of finding a path that covers the entire free space of a confined area, with applications ranging from robotic lawn mowing to search-and-rescue. While for known environments, offline methods can find provably complete paths, and in some cases optimal solutions, unknown environments need to be planned \emph{online} during mapping. We investigate the suitability of continuous-space reinforcement learning (RL) for this challenging problem, and propose a computationally feasible egocentric map representation based on frontiers, as well as a novel reward term based on total variation to promote complete coverage. Compared to existing classical methods, this approach allows for a flexible path space, and enables the agent to adapt to specific environment characteristics. Meanwhile, the deployment of RL models on real robot systems is difficult. Training from scratch may be infeasible due to slow convergence times, while transferring from simulation to reality, i.e.\ \emph{sim-to-real transfer}, is a key challenge in itself. We bridge the sim-to-real gap through a semi-virtual environment, including a real robot and real-time aspects, while utilizing a simulated sensor and obstacles to enable environment randomization and automated episode resetting. We investigate what level of fine-tuning is needed for adapting to a realistic setting. Through extensive experiments, we show that our approach surpasses the performance of both previous RL-based approaches and highly specialized methods across multiple CPP variations in simulation. Meanwhile, our method successfully transfers to a real robot. Our code implementation can be found online.\footnote{Link to code repository:\ \url{https://github.com/arvijj/rl-cpp}}
\end{abstract}


\titlepgskip=-21pt

\maketitle

\begin{figure*}[!t]
    \centering
    \setlength{\tabcolsep}{5pt}
    \begin{tabular}{cc}
        \includegraphics[height=.3\linewidth]{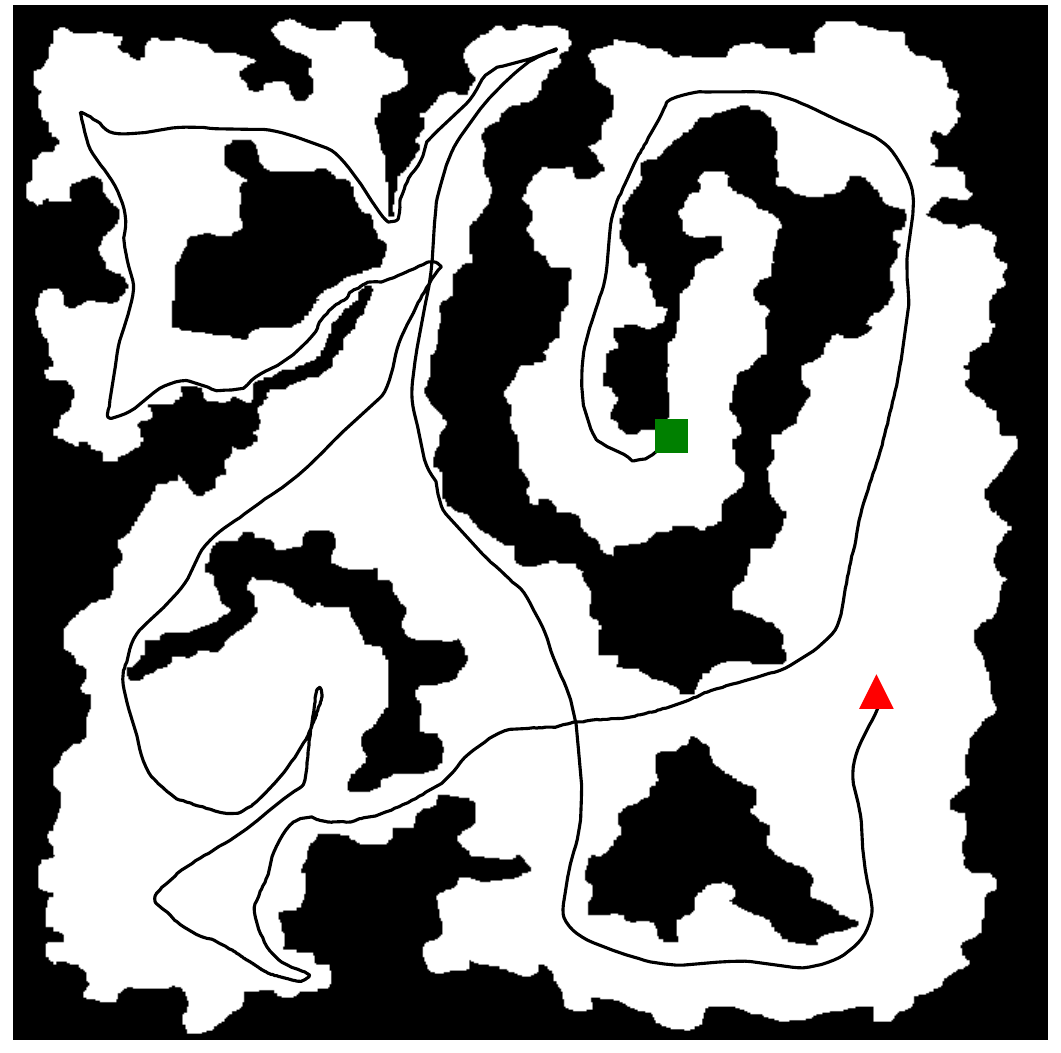} &
        \includegraphics[height=.3\linewidth]{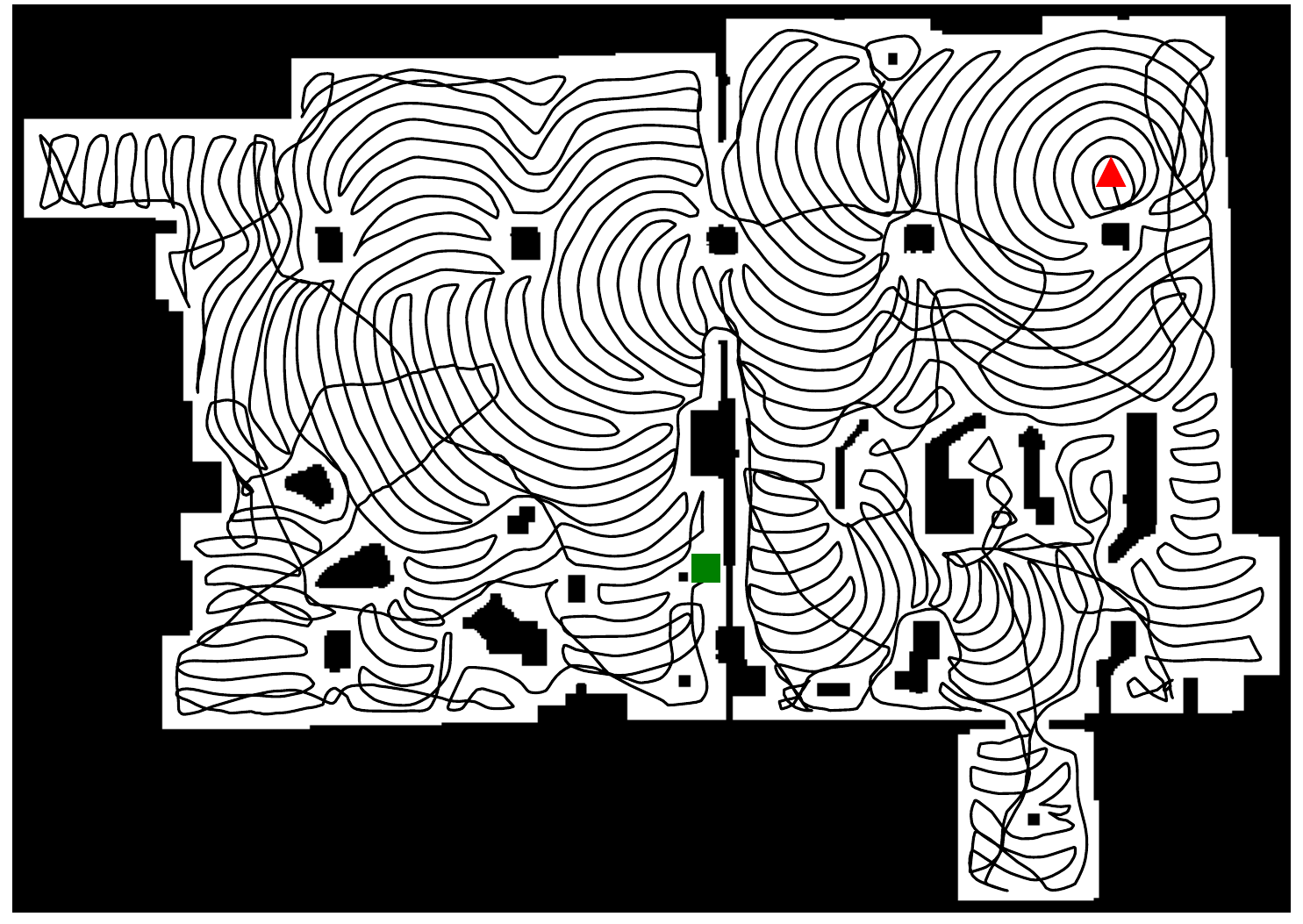} \\
    \end{tabular}
    \caption{Learned paths for exploration (left) and lawn mowing (right), including the start (red triangle) and end position (green square).}
    \label{fig_qualitative_paths}
\end{figure*}

\section{Introduction}

\PARstart{T}{ask} automation is an ever-growing field, and as automated systems become more intelligent, they are expected to solve increasingly challenging problems. One such problem is coverage path planning (CPP), where the task is to find a path that covers all of the free space of an environment. If the environment is known, an optimal path can be planned offline \cite{huang2001optimal}. In contrast, if it is unknown, the path has to be planned online during mapping of the environment, e.g.\ by a robot, and an optimal path cannot be found in the general case~\cite{galceran2013ras}. CPP has found its uses in many robotic applications, such as lawn mowing \cite{cao1988jrs}, vacuum cleaning \cite{yasutomi1988icra}, search-and-rescue \cite{jia2016ssrr}, and exploration \cite{xu2022explore}.

We aim to develop a task- and robot-agnostic method for online CPP in unknown environments. This is a challenging task, as intricacies in certain applications may be difficult to model, or even impossible, in some cases. In the real world, unforeseen discrepancies between the modeled and true environment may occur, \textit{e.g.}\ due to wheel slip. Incompleteness of the present model must be assumed to account for open world scenarios. Thus, we approach the problem from the perspective of learning through embodiment. Reinforcement learning (RL) lends itself as a natural choice, where an agent can interact with the world and adapt to specific environment conditions, without the need for an explicit model. Concretely, we consider the case where an RL model directly predicts continuous control signals for an agent from sensor data. We present an RL approach for the CPP problem in terms of a multi-scale map feature representation using frontiers, a continuous action space, a network architecture based on grouped convolutions, and a reward term based on total variation to encourage complete coverage. Our approach finds coverage paths in complex environments for different variations of the CPP problem, as can be seen in Fig. \ref{fig_qualitative_paths}.

A practical limitation for training machine learning models is the need for large amounts of real-world data, which is tedious and time-consuming to collect. In particular, training reinforcement learning agents for robotics from scratch requires access to the robot for the full duration of the training process. Meanwhile, during the early training phase, the agent is more likely to make mistakes that may damage the hardware or require human supervision for resetting the episodes. Instead, learning in simulation presents an attractive alternative. However, due to differences in the dynamics between the simulation and the real world, transferring a model from simulation to reality is challenging. Prior work reduce the sim-to-real gap by improving the simulation, e.g.\ by injecting realistic noise \cite{kaufmann2023champion,zhao2020towards}, applying domain randomization \cite{muratore2021data,tobin2017domain}, or through meta learning \cite{arndt2020meta,nagabandi2019learning}. Our goal is to transfer, for the task of coverage path planning, models trained in simulation to real environments, by fine-tuning in a realistic setting.

When training reinforcement learning models in real time on physical robots, additional considerations need to be accounted for compared to training in simulation. (1)~There exists a mismatch between the real and simulated robot kinematics, dynamics, and sensing, such as slippage and noisy localization. This leads to different transition probabilities in the real world, where the predicted actions based on simulation may be suboptimal. (2)~Due to inertia and latencies in the system, the dynamics are non-Markovian \cite{haarnoja2018soft_2}. This violates the common assumption that the Markov property holds. (3)~Since the robot keeps moving during the various computations in the training process, the state observed by the agent is not perfectly aligned with the true state. This introduces a delay, where the agent predicts actions based on outdated observations of the environment. In this work, we address these problems.

To smoothen the transition into the real world, we use a real robot in a semi-virtual environment, utilizing a highly accurate positioning system in an indoor planar setting, with a simulated sensor and obstacles. This introduces the previously mentioned real-time aspects of reinforcement learning in a controlled manner, while allowing flexibility in utilizing automated episode resetting and varying training environments without having to physically construct them. The conclusions drawn from the fine-tuning experiments in this semi-virtual setting are also relevant for the case of completely real environments, although they might need to be complemented with perception-specific findings.

To reduce latency in the training process, we perform model updates in parallel with the action and state selection \cite{yuan2022asynchronous}, and perform all computations on on-board hardware. We utilize soft actor-critic learning \cite{haarnoja2018soft} for its sample efficiency and its efficacy on continuous action spaces. To account for the non-Markovian dynamics in a lightweight manner, we include past actions in the observation space. To reduce the mismatch between the simulated and real kinematics, we measure the real-world linear and angular accelerations as well as action delay, and incorporate them into the simulation. We find that a high inference frequency enables first-order Markovian policies to transfer to a real setting. By introducing delays and action observations, higher-order Markovian models can be fine-tuned to further reduce the sim-to-real gap. Moreover, these models can operate at a lower frequency, thus reducing computational requirements for deployed systems.

This work builds on the conference paper \cite{jonnarth2024learning}, and is part of the PhD thesis \cite{jonnarth2024phd}. Our contributions are as follows:
\begin{itemize}[leftmargin=*]
    \item We propose an end-to-end deep reinforcement learning approach in continuous space for the online CPP problem. This includes a multi-scale map representation exploiting frontiers, a network architecture using grouped convolutions, and a novel reward term based on total variation.
    \item We propose to divide the sim-to-real problem into two steps with an intermediate case of a real robot in a virtual environment. By performing data collection and model updates in parallel, our approach enables sim-to-real transfer of CPP policies through real-time fine-tuning online without a separate system or stopping the robot for model updates.
    \item Finally, we conduct extensive experiments, both in simulation and in a realistic setting, including detailed ablations on our proposed map representation, network architecture, and reward function, all of which improve the coverage time. Our approach surpasses the performance of both classical and RL-based methods, while successfully transferring to a real robot.
\end{itemize}

\section{Related Work}

This work relates to coverage path planning, transferring models from simulation to the real world, and online RL in real time. We summarize the related work below.

\subsection{Coverage Path Planning}

Coverage path planning methods can roughly be categorized as planning-based or learning-based. \textit{Planning-based} methods include decomposition methods, which divide the area into cells based on e.g.\ boustrophedon cellular decomposition (BCD) \cite{choset1998coverage} or Morse decomposition \cite{acar2002morse}, where each cell is covered with a pre-defined pattern. Grid-based methods, such as Spiral-STC \cite{gabriely2002ras} and the backtracking spiral algorithm (BSA) \cite{gonzalez2005icra}, discretize the area into even smaller cells, e.g.\ squares, with a similar size to the agent. Then, a path is planned based on motion primitives connecting adjacent cells, i.e.\ to move up, down, left, or right on the grid, such that each cell is visited at least once. Frontier-based methods plan a path to a chosen point on the frontier, i.e.\ on the boundary between covered and non-covered regions. The choice of frontier point can be based on different aspects, such as the distance to the agent \cite{yamauchi1997frontier}, the path of a rapidly exploring random tree (RRT) \cite{umari2017autonomous} or the gradient in a potential field \cite{yu2021smmr}. \textit{Learning-based} methods apply machine learning techniques, typically in combination with planning-based methods, to find coverage paths. Reinforcement learning is the most popular paradigm due to the sequential nature of the task. Chen \textit{et al.} \cite{chen2019iros} use RL to find the order in which to cover the cells generated by BCD. Discrete methods \cite{piardi2019aip,kyaw2020access} use RL to learn which motion primitives to perform. RL has also been combined with frontier-based methods, either for predicting the cost of each frontier point \cite{niroui2019ral}, or for learning the control signals to navigate to a chosen point \cite{hu2020tovt}. In contrast, we learn continuous control signals end-to-end from a built map and sensor data to fully utilize the flexibility of RL. To the best of our knowledge, we are the first to do so for CPP.

\subsection{Map Representation}

To perform coverage path planning, the environment needs to be represented in a suitable manner. Similar to previous work, we discretize the map into a 2D grid with sufficiently high resolution to accurately represent the environment. This map-based approach presents different choices for the input feature representation. Saha \textit{et al.} \cite{saha2021efficient} observe such maps in full resolution, where the effort for a $d \times d$ grid is $\mathcal{O}(d^2)$, which is infeasible for large environments. Niroui \textit{et al.} \cite{niroui2019ral} resize the maps to keep the input size fixed. However, this means that the information in each grid cell varies between environments, and hinders learning and generalization for differently sized environments. Meanwhile, Shrestha \textit{et al.}~\cite{shrestha2019learned} propose to learn the map in unexplored regions. Instead of considering the whole environment at once, other works \cite{heydari2021arxiv,saha2021deep} observe a local neighborhood around the agent. While the computational cost is manageable, the long-term planning potential is limited as the space beyond the local neighborhood cannot be observed. For example, if the local neighborhood is fully covered, the agent must pick a direction at random to explore further. To avoid the aforementioned limitations, we use multiple maps in different scales, similar to Klamt and Behnke~\cite{klamt2018planning}.

\subsection{Sim-to-real Transfer}

Transferring from simulation to the real world is challenging due to mismatch in both sensing and actuation \cite{zhao2020sim}. Prior work has approached this challenge by different means. Domain randomization has been utilized to randomize physical parameters in simulation, such as mass and joint damping \cite{muratore2021data}, or to randomize textures and lighting conditions in the image domain \cite{tobin2017domain}. Other works introduce perturbations in the sensing and actuation \cite{zhao2020towards}. The motivation behind these approaches is that a highly randomized simulation would cover the real-world distribution, while avoiding a highly accurate model of it \cite{zhao2020sim}. Instead of randomizing various aspects in simulation, we train directly on the real distribution by fine-tuning in the real world. Meta learning methods aim to quickly adapt to new unseen tasks from a wide variety of training task, such as adapting to the real world from simulation \cite{arndt2020meta,nagabandi2019learning}. Another approach is to learn from expert demonstrations through imitation learning \cite{yan2017sim}, which has previously been applied to coverage path planning \cite{hu2020tovt}. In contrast, we avoid human demonstrations in order to reduce the manual effort. When it comes to robot control, Niroui \textit{et al.} \cite{niroui2019ral} deploy an RL policy for CPP trained in simulation on a differential drive robot without fine-tuning. The policy predicts the next frontier node, where a separate non-learned module navigates to it, thus being less affected by misaligned kinematics between simulation and reality. Kaufmann \textit{et al.} \cite{kaufmann2023champion} transfer a lightweight RL control policy using a pre-trained perception system for first-person-view drone racing, utilizing empirical noise models to improve the fidelity of the simulator. They collect perception and dynamics residuals in the real world based on a highly accurate positioning system, which they use to augment the simulation. In contrast to these works, we fine-tune our model online in the real world.

\subsection{Real-time Reinforcement Learning}

Compared to turn-based and simulated environments, where the time for action selection and policy updates are assumed to be zero and the next state can be advanced to instantly, RL in the real world presents new challenges where these assumptions do not hold. For example, the forward and backward passes of the policy network during action selection and model updates take time, which cannot be neglected. To take this into account, the core idea in the literature is to parallelize aspects of the environment interactions and the model training. Bakker \textit{et al.} \cite{bakker2006quasi} explore \textit{quasi-online reinforcement learning}, where a model of the environment is built online and in parallel to training the policy based on the built environment model. Ramstedt and Pal \cite{ramstedt2019real} propose to perform the action selection in parallel with the state selection. Concretely, given the current state and action to be taken, the next state and new action are computed concurrently. This approach takes into account the time for action selection by delaying the observed state by one time step. However, it does not consider the time for model updates, which is typically much larger. Other works \cite{haarnoja2019learning,wang2023real} distribute the model training for soft actor-critic (SAC) \cite{haarnoja2018soft} to a separate system, while performing only the lightweight action and state selections on the on-board target hardware. This allows the data collection to be executed independently from the model updates, where the policy and replay buffer are periodically synchronized. Within this paradigm, Wang et al.\ \cite{wang2023real} propose ReLoD, a system to efficiently distribute the learning algorithm over a resource-limited local computer and a powerful remote computer. Yuan and Mahmood \cite{yuan2022asynchronous} employ a similar approach, but they run everything on a single edge device. They divide the training process into separate processes for the action selection, state selection, batch sampling, and gradient computation. We follow this approach. However, in our experimental setup, the action selection and batch sampling were much faster than the state selection and gradient computations, so we only use two threads in order to reduce complexity and communication overhead.

\section{Reinforcement Learning of Coverage Paths}

This section presents our approach for learning coverage paths using reinforcement learning. First, we define the online CPP problem in Section \ref{sec_problem_definition}, and subsequently formulate it as a partially observable Markow decision process (POMDP) in Section \ref{sec_pomdp}. After that, we present our RL-based approach in terms of observation space in Section \ref{sec_observation_space}, action space in Section \ref{sec_action_space}, reward function in Section \ref{sec_reward_function}, and agent architecture in Section \ref{sec_agent_architecture}.

\subsection{Problem Definition and Delineations}
\label{sec_problem_definition}

The goal is to navigate with an agent of radius $r$ to visit all accessible points in a confined area without prior knowledge of its geometry. The free space includes all points inside the area that are not occupied by obstacles. A point is considered visited when the distance between the point and the agent is less than the coverage radius $d$, and the point is within the field-of-view of the agent. This definition unifies variations where $d \leq r$, which we define as the \textit{lawn mowing problem}, and where $d > r$, which we refer to as \textit{exploration}. To interactively gain knowledge about the environment, and for mapping its geometry, the agent needs some form of sensor. Both ranging sensors \cite{hu2020tovt} and depth cameras \cite{chen2018learning} have been utilized for this purpose. Following Xu \textit{et al.} \cite{xu2022explore}, we choose a simulated 2D light detection and ranging (lidar) sensor that can detect obstacles in fixed angles relative to the agent, although our proposed framework is not limited to this choice. Based on the pose of the agent, the detections are transformed to global coordinates, and a map of the environment is continuously built. As the focus of this paper is to learn coverage paths, and not to solve the localization problem, we assume known pose up to a certain noise level. However, our method may be extended to the case with unknown pose with the use of an off-the-shelf SLAM method. Finally, while our method can be further extended to account for moving obstacles and multi-agent coverage, they are beyond the scope of this paper.

\begin{figure*}[!t]
    \centering
    \includegraphics[width=0.99\linewidth]{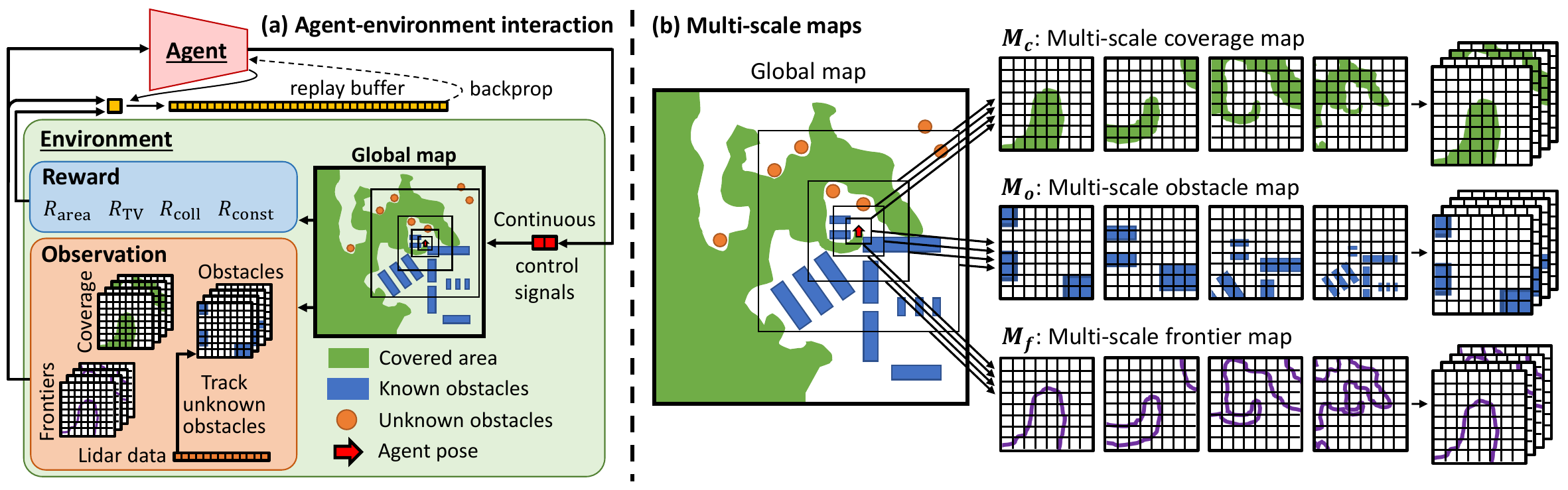}
    \caption{\textbf{(a) Agent-environment interaction:} The observation consists of multi-scale maps from (b) and lidar detections, based on which the model predicts continuous control signals for an agent. \textbf{(b) Illustration of coverage, obstacle, and frontier maps in multiple scales:} This example shows $m=4$ scales with a scale factor of $s=2$. All scales are centered at the agent, and discretized into the same pixel resolution, resulting in the multi-scale maps $M_c$, $M_o$, and $M_f$, of size $8 \times 8 \times 4$ in this example.}
    \label{fig_multi_scale_maps}
\end{figure*}

\subsection{CPP as a Markov Decision Process}
\label{sec_pomdp}

We formulate the CPP problem as a partially observable Markov decision process (POMDP), where an agent performs actions in an environment such as to maximize a reward. It is defined by the tuple $(\mathcal{S}, \mathcal{A}, T, R, \Omega, \mathcal{O})$, where $\mathcal{S}$ is the set of possible states, $\mathcal{A}$ is the set of actions, $T: \mathcal{S} \times \mathcal{A} \times \mathcal{S} \rightarrow [0, \infty)$ is the state-transition function representing the probability density of the next state $s_{t+1} \in \mathcal{S}$ given the current state $s_t \in \mathcal{S}$ and the action $a_t \in \mathcal{A}$, $R: \mathcal{S} \times \mathcal{A} \rightarrow \mathbb{R}$ is the reward function, $\Omega$ is the set of observations, and $\mathcal{O} : \mathcal{S} \times \mathcal{A} \times \Omega \rightarrow [0, \infty)$ is the observation function representing the probability density of the observation $o_t \in \mathcal{O}$ given the state $s_t \in \mathcal{S}$ and the previous action $a_{t-1} \in \mathcal{A}$. At each discrete time step $t \in \{1, ..., N\}$, the environment is in some state $s_t$, in which the agent predicts an action $a_t \sim \pi (a_t | o_t)$ according to a policy $\pi$, based on the observation $o_t$ of the current state. Subsequently, the environment transitions into a new state $s_{t+1} \sim p(s_{t+1} | s_t, a_t)$ according to $T$, while emitting a reward $r_{t+1} = R(s_{t+1}, a_t)$ and a new observation $o_{t+1} \sim p(o_{t+1} | s_{t+1}, a_t)$ according to $\mathcal{O}$. The goal for the agent is to maximize the expected future discounted reward $\mathbb{E} (\sum_{k=t}^N \gamma^{k-t} r_{k})$, with discount factor $\gamma$.

In our particular problem, the policy is a neural network that predicts continuous control signals, $a_t$, for the agent. The state $s_t$ includes the full environment geometry with all obstacles, the set of points that have been covered, as well as the pose of the agent. However, the agent only has information about what it has observed, so the observation $o_t$ consists of the part of the environment that has been mapped until time step $t$, along with covered points, agent pose, and sensor data. The reinforcement learning loop is depicted in Fig.~\ref{fig_multi_scale_maps}(a). In the setting with a real robot, the first-order Markovian property is not fulfilled, due to the dynamics including inertia and momentum. The usual approach to augment previous states in the current one becomes infeasible if the motion is implicitly represented by the agent-centered environment maps. Thus higher-order effects lead to model errors that need correction in subsequent steps. We go into further detail in Section \ref{sec_smoothening_sim_to_real_gap}.

\subsection{Observation Space}
\label{sec_observation_space}

To efficiently learn coverage paths, the observation space needs to be mapped into a suitable input feature representation for the neural network policy. To this end, we represent the visited points as a \textit{coverage map}, and the mapped obstacles and boundary of the area as an \textit{obstacle map}. The maps are discretized into a 2D grid with sufficiently high resolution to accurately represent the environment. To represent large regions in a scalable manner, we propose to use multi-scale maps, which was necessary for large environments. We make this viable through \textit{frontier maps} that preserve information about the existence of non-covered space, even in coarse scales.

\subsubsection{Multi-scale Maps}

Inspired by multi-layered maps with varying scale and coarseness levels for search-and-rescue \cite{klamt2018planning}, we propose to use multi-scale maps for the coverage and obstacles to solve the scalability issue. We extract multiple local neighborhoods with increasing size and decreasing resolution, keeping the grid size fixed. We start with a local square crop $M^{(1)}$ with side length $d_1$ for the smallest and finest scale. The multi-scale map representation $M = \{M^{(i)}\}_{i=1}^m$ with $m$ scales is constructed by cropping increasingly larger areas based on a fixed scale factor $s$. Concretely, the side length of map $M^{(i)}$ is $d_{i} = s d_{i-1}$. The resolution for the finest scale is chosen sufficiently high such that the desired level of detail is attained in the nearest vicinity of the agent, allowing it to perform precise \textit{local navigation}. At the same time, the large-scale maps allow the agent to perform \textit{long-term planning}, where a high resolution is less important. This multi-scale map representation can completely contain an area of size $d \times d$ in $\mathcal{O}(\log d)$ number of scales. The total number of grid cells is $\mathcal{O}(wh \log d)$, where $w$ and $h$ are the fixed width and height of the grids, and do not depend on $d$. This is a significant improvement over a single fixed-resolution map with $\mathcal{O}(d^2)$ grid cells. For the observation space, we use a multi-scale map $M_c$ for the coverage and $M_o$ for the obstacles. These are illustrated in Fig. \ref{fig_multi_scale_maps}(b).

\subsubsection{Frontier Maps}

When the closest vicinity is covered, the agent needs to make a decision where to explore next. However, the information in the low-resolution large-scale maps may be insufficient. For example, consider an obstacle-cluttered region where the obstacle boundaries have been mapped. A low coverage could either mean that some parts are non-covered free space, or that they are part of the interior of an obstacle. These cases cannot be distinguished if the resolution is too low. As a solution to this problem, we propose to encode a multi-scale frontier map $M_f$, which we define in the following way. In the finest scale, a non-obstacle grid cell that has not been visited is a frontier point if any of its neighbours have been visited. Thus, a frontier point is non-visited free space that is reachable from the covered area. A region where the entire free space has been visited does not induce any frontier points. In the coarser scales, a grid cell is a frontier cell if and only if it contains at least one frontier point. In this way, the existence of frontier points persists through scales. Thus, regions with non-covered free space can be deduced in any scale, based on this multi-scale frontier map representation.

\subsubsection{Egocentric Maps}

As the movement is made relative to the agent, its pose needs to be related to the map of the environment. Following Chen \textit{et al.} \cite{chen2018learning}, we use egocentric maps which encode the pose by representing the maps in the coordinate frame of the agent. Each multi-scale map is aligned such that the agent is in the center of the map, facing upwards. This allows the agent to easily map observations to movement actions, instead of having to learn an additional mapping from a separate feature representation of its position, such as a 2D one-hot map \cite{theile2020iros}.

\subsubsection{Sensor Observations}

To react on short-term obstacle detections, we include the sensor data in the input feature representation. The depth measurements from the lidar sensor are normalized to $[0, 1]$ based on its maximum range, and concatenated into a vector $S$. Note, however, that this assumes a fixed set if lidar parameters, i.e.\ for the range, field-of-view, and number of rays.

\subsection{Action Space}
\label{sec_action_space}

We let the model directly predict the control signals for the agent. This allows it to adapt to specific environment characteristics, while avoiding a constrained path space, different from a discrete action space. We consider an agent that independently controls the linear velocity $v$ and the angular velocity $\omega$, although the action space may seamlessly be changed to specific vehicle models. To keep a fixed output range, the actions are normalized to $[-1, 1]$ based on the maximum linear and angular velocities, where the sign of the velocities controls the direction of travel and rotation.

In the real-world setting, we use a differential drive wheeled robot, which is controlled by two separately driven wheels. The agent predicts the linear and angular velocities $v$ and $\omega$ for the robot, which are converted to angular velocities $\omega_R$ and $\omega_L$ for the right and left wheels,
\begin{equation}
    \label{eq_left_right_velocities}
    \omega_R = \frac{v}{r_w} + \frac{\omega b}{2 r_w}, \qquad \omega_L = \frac{v}{r_w} - \frac{\omega b}{2 r_w},
\end{equation}
where $r_w$ is the wheel radius and $b$ is the distance between the wheels.

A continuous action space is of course not the only choice, but it is more realistic than a discrete one. There are many reasons why we chose continuous actions. (1) This allows us to model a continuous pose that is not constrained by the grid discretization, and thus not constraining the path space. (2) The agent can adapt and optimize its path for a specific kinematic model. In our experiments in Section \ref{sec_experiments} we find that our approach works well in continuous settings, which hints that it may also work in other continuous control spaces for specific kinematic models. The same conclusion could not be drawn if a discrete action space was used. Our specific choice of linear and angular velocities applies directly to a wide variety of robots, e.g. differential drive systems and the Ackermann kinematic model (by using a constraint on the angular velocity). This makes it more suitable for sim-to-real transfer compared to a discrete action space. (3) Compared to grid-based discrete actions, a continuous action space introduces additional sources for error, both regarding dynamics and localization, which are more realistic.

\subsection{Reward Function}
\label{sec_reward_function}

\begin{figure}[t]
    \centering
    \includegraphics[width=0.95\linewidth]{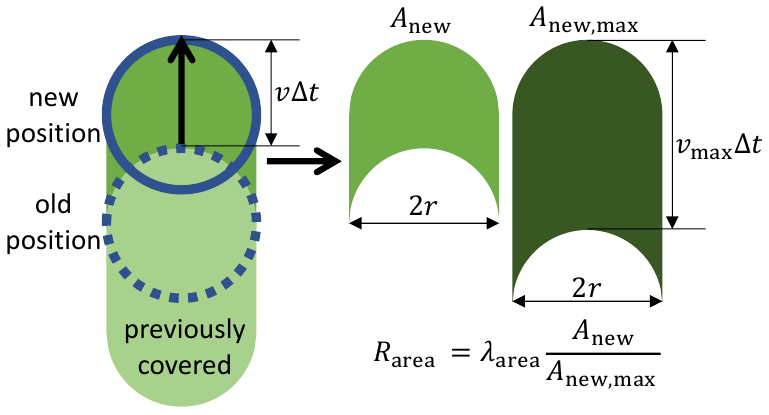}
    \caption{An illustration of the area reward $R_\mathrm{area}$, which is based on the maximum possible area that can be covered in each time step.}
    \label{fig_reward_area}
\end{figure}

As the goal is to cover the free space, a reward based on covering new ground is required. Similar to Chaplot \textit{et al.} \cite{chaplot2020Learning} and Chen \textit{et al.} \cite{chen2018learning}, we define a reward term $R_\mathrm{area}$ based on the newly covered area $A_\mathrm{new}$ in the current time step that was not covered previously. To control the scale of this reward signal, we first normalize it to $[0, 1]$ based on the maximum area that can be covered in each time step, which is related to the maximum speed $v_\mathrm{max}$. We subsequently multiply it with the scale factor $\lambda_\mathrm{area}$, which is the maximum possible reward in each step, resulting in the reward
\begin{equation}
    \label{eq_area_reward}
    R_\mathrm{area} = \lambda_\mathrm{area} \frac{A_{\mathrm{new}}}{2 r v_\mathrm{max} \Delta t},
\end{equation}
where $r$ is the agent radius and $\Delta t$ is the time step size. See Fig. \ref{fig_reward_area} for an illustration.

\begin{figure}[t]
    \centering
    \setlength{\tabcolsep}{1pt}
    \begin{tabular}{cccccccc}
        \includegraphics[width=.49\linewidth]{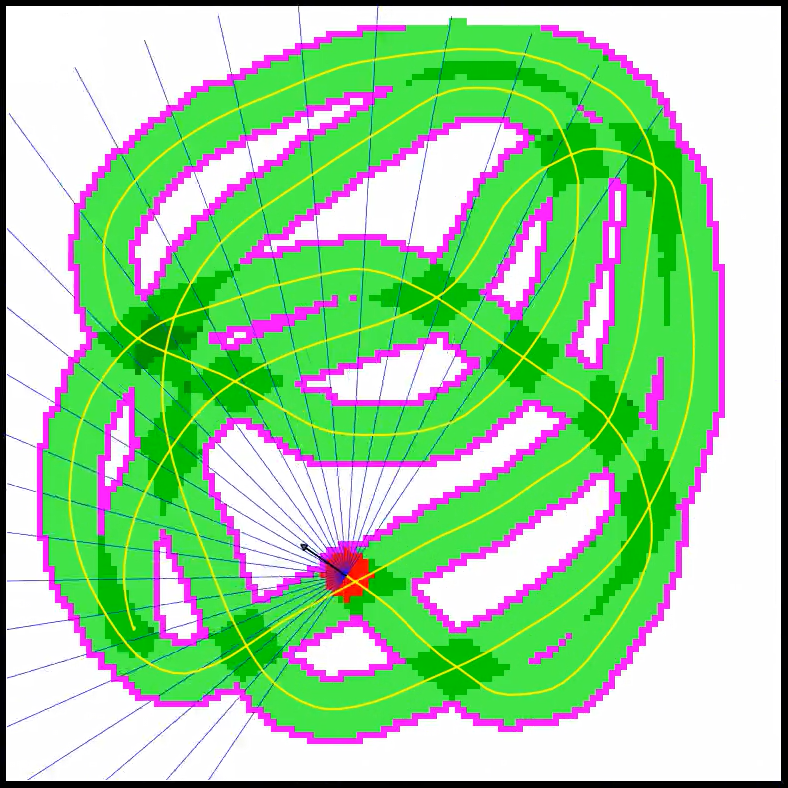} &
        \includegraphics[width=.49\linewidth]{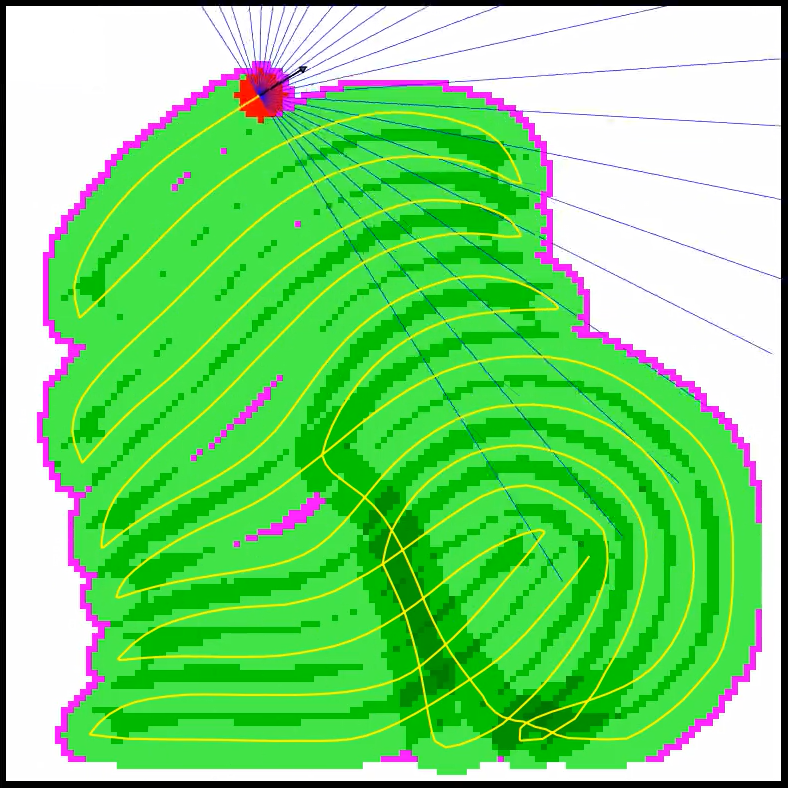} \\
    \end{tabular}
    \caption{The total variation reward term improves path quality. The figure shows the path (yellow line) without TV reward (left), and with TV reward (right), including the covered region (green), lidar rays (blue lines) and frontier points (magenta).}
    \label{fig_tv_paths}
\end{figure}

By only maximizing the coverage reward in \eqref{eq_area_reward}, the agent is discouraged from overlapping its previous path, as this reduces the reward in the short term. This leads to holes or stripes in the coverage, which we observed in our experiments, see Fig. \ref{fig_tv_paths}. These leftover parts can be costly to cover later on for reaching complete coverage. Covering the thin stripes afterward only yields a minor reward, resulting in slow convergence towards an optimal path. To reduce the leftover parts, we propose a reward term based on minimizing the total variation (TV) of the coverage map. Minimizing the TV corresponds to reducing the boundary of the coverage map, and thus leads to fewer holes. Given a 2D signal $x$, the discrete isotropic total variation, which has been used for image denoising \cite{rudin1992nonlinear}, is expressed as
\begin{equation}
    V(x) = \sum_{i,j} \sqrt{|x_{i+1,j} - x_{i,j}|^2 + |x_{i,j+1} - x_{i,j}|^2}.
\end{equation}

We consider two variants of the TV reward term, a global and an incremental. For the global TV reward $R_\mathrm{TV}^\mathrm{G}$, the agent is given a reward based on the global coverage map $C_t$ at time step $t$. To avoid an unbounded TV for large environments, it is scaled by the square root of the covered area $A_\mathrm{covered}$, as this results in a constant TV for a given shape of the coverage map, independent of scale. The incremental TV reward $R_\mathrm{TV}^\mathrm{I}$ is based on the difference in TV between the current and the previous time step. A positive reward is given if the TV is decreased, and vice versa. The incremental reward is scaled by the maximum possible increase in TV in a time step, which is twice the traveled distance. The global and incremental rewards are respectively given by
\begin{align}
    R_\mathrm{TV}^\mathrm{G}(t) & = - \lambda_\mathrm{TV}^\mathrm{G} \frac{V(C_t)}{\sqrt{A_\mathrm{covered}}}, \\
    R_\mathrm{TV}^\mathrm{I}(t) & = - \lambda_\mathrm{TV}^\mathrm{I} \frac{V(C_t) - V(C_{t-1})}{2 v_\mathrm{max} \Delta t},
\end{align}
where $\lambda_\mathrm{TV}^\mathrm{G}$ and $\lambda_\mathrm{TV}^\mathrm{I}$ are reward scaling parameters to make sure that $|R_\mathrm{TV}| < |R_\mathrm{area}|$ on average. Otherwise, the optimal behaviour is simply to stand still.

To avoid obstacle collisions, a negative reward $R_\mathrm{coll}$ is given each time the agent collides with an obstacle. Finally, a small constant negative reward $R_\mathrm{const}$ is given in each time step to encourage fast execution. Thus, our final reward function reads
\begin{equation}
    R = R_{\mathrm{area}} + R_\mathrm{TV}^\mathrm{G} + R_\mathrm{TV}^\mathrm{I} + R_{\mathrm{coll}} + R_{\mathrm{const}}.
\end{equation}
During training, each episode is terminated when the agent reaches a pre-defined goal coverage, or when it has not covered any new space in $\tau$ consecutive time steps.

\subsection{Agent Architecture}
\label{sec_agent_architecture}

Due to the multi-modal nature of the observation space, we use a map feature extractor $g_m$, a sensor feature extractor $g_s$, and a fusing module $g_f$. The map and sensor features are fused, resulting in the control signal prediction
\begin{equation}
    (v, \omega) = g_f ( g_m(M_c, M_o, M_f) , g_s(S) ).
\end{equation}

\begin{figure}[t]
    \centering
    \includegraphics[width=0.99\linewidth]{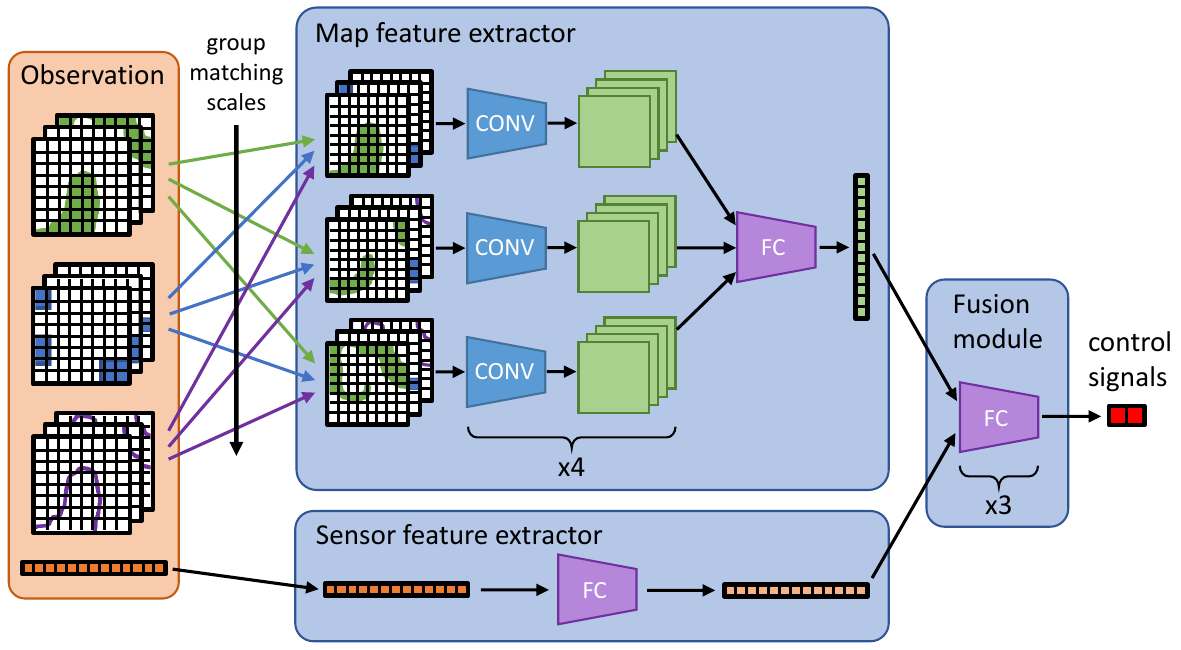}
    \caption{Our proposed SGCNN architecture consists of convolution (CONV) and fully connected (FC) layers. The scales of the multi-scale maps are convolved separately as their spatial positions are not aligned in the grid. x3/x4 refer to the number of layers.}
    \label{fig_cnn_architecture}
\end{figure}

We consider three network architectures, a simple multilayer perceptron (MLP), a standard convolutional neural network (CNN), and our proposed scale-grouped convolutional neural network (SGCNN) that independently processes the different scales in the maps, see Fig. \ref{fig_cnn_architecture}. The MLP is mainly used as a benchmark to evaluate the inductive priors in the CNN-based architectures. For the MLP, the feature extractors, $g_m$ and $g_s$, are identity functions that simply flatten the inputs, and the fusing module consists of three fully connected (FC) layers. The CNN-based architectures use convolutional layers in the map feature extractor followed by a single FC layer. In SGCNN, we group the maps in $M_c$, $M_o$ and $M_f$ by scale, as the pixel positions between different scales do not correspond to the same world coordinate. Each scale is convolved separately using grouped convolutions. This ensures that each convolution kernel is applied to grids where the spatial context is consistent across channels. The sensor feature extractor is a single FC layer, and the fusing module consists of three FC layers. More details can be found in Appendix \ref{supp_sec_agent_architecture}.

\section{Transferring CPP Agents to the Real World}

Training from scratch directly on the real robot is infeasible due to a slow convergence of the training process. In Section~\ref{sec_simulation_experiments}, we find that the number of training iterations required is in the order of millions, which would translate to non-stop training for weeks to months in the real world. Meanwhile, it would require some form of manual interaction. Thus, we aim to transfer CPP agents trained in simulation, and fine-tune them on a real robot. In this section, we describe our proposed approach for sim-to-real transfer.

\subsection{Learning CPP on the Real System}

Even if we will not train the agent from scratch in the real setting, we need to be able to fine-tune the transferred system by RL. In common RL-libraries, such as Stable-Baselines3~\cite{raffin2021stable}, Tianshou \cite{weng2022tianshou}, and Spinning Up \cite{SpinningUp2018}, the data collection and model updates are performed serially. While this is feasible in simulation and turn-based environments where the environment can be paused during model updates and time can be advanced instantly to the next state after action selection, it is not practical for real-time robotic applications where the agent is trained online.

In this case, we need to wait after action selection to observe the next state, while the execution remains idle. After the environment interaction, a batch is sampled from the replay buffer, gradients are computed, and the model is updated. However, this takes time, which is not negligible, especially on low-performance edge devices and embedded systems. During the gradient update step, the robot keeps following its previous action. As a result, the agent ends up in a state which deviates from the recorded one used for action selection. Following previous work \cite{haarnoja2019learning,yuan2022asynchronous}, we perform the model updates in parallel with the data collection, utilizing computational resources which would otherwise remain idle while waiting for the next state.

The training process can be divided into four main computational steps, where the respective measured times are given in Table \ref{tab_times}:

\textbf{(1)~Action selection:} Given the current observation, we sample an action from the policy. This corresponds to a forward pass of the policy network.

\textbf{(2)~State selection:} After sampling an action, the control signals are sent to the robot platform, and the new state is observed. In simulation, this would occur instantly, while in the real world we need to let time pass to observe how the action affected the environment. After state selection, the reward is computed, and the action, previous state, new state, and reward are added to the replay buffer.

\textbf{(3)~Batch sampling:} A training batch is sampled from the replay buffer.

\textbf{(4) Model update:} Gradients are computed based on the training batch, and the model weights are updated. This is the most computationally intensive part.

\begin{table}[t]
    \centering
    \caption{Measured times of the RL-step on the real system. Timing for (2) is application dependent, and chosen based on the optimal training step size. The overhead effectively becomes part of (1) resulting in an action delay of 50 ms.}
    \begin{tabular}{lccccc}
        \toprule
        Step & (1) & (2) & (3) & (4) & Overhead \\
        \midrule
        Time & 12 ms & 450 ms & 15 ms & 80 ms & 38 ms \\
        \bottomrule
    \end{tabular}
    \label{tab_times}
\end{table}

\begin{figure}[t]
    \centering
    \includegraphics[width=0.99\linewidth]{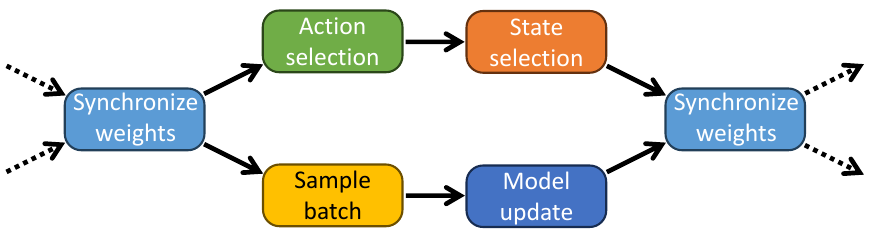}
    \caption{During online training we perform environment interactions (top) and model training (bottom) in parallel. Either branch can be executed multiple times before weight synchronization.}
    \label{fig_real_time_rl}
\end{figure}

Since the action selection and batch sampling were fast compared to the state selection and gradient computations, we only use two threads to reduce overhead and complexity. Our computational graph is shown in Fig. \ref{fig_real_time_rl}, which consists of an environment interaction thread (top) and a model update thread (bottom). Since both threads interact with the replay buffer, we use a mutex to avoid conflicts. Luckily, the environment interaction thread adds an entry at the end of the cycle, while the model update thread samples from it early, so the risk of the threads blocking each other is low.

Since the model update thread modifies the weight tensors during a large portion of the runtime, i.e.\ both during the backward pass to update the gradients and during the optimizer step to update the parameters, a mutex is not feasible as it would block execution too often. Instead, we keep a copy of the model, which is only accessed by the model update thread. The weights are synchronized when both threads have finished execution.

With this approach, both the environment interaction and the model update can be performed multiple times before model synchronization. This is useful when the computation time for the model update exceeds that of the action and state selection. In this case, the number of environment interaction steps should be chosen to match the computation time for the model update. Meanwhile, if the model update is fast compared to the environment interaction, multiple updates can be performed during one environment interaction step, which was the case in our experiments.

Apart from the four main computational steps, the time for simulating the sensor, creating the observation, and synchronizing the weights results in additional overhead. This effectively becomes part of the action selection as the overhead delays the observed state, and results in an \textit{action delay}.

\begin{figure}[t]
    \centering
    \includegraphics[width=0.99\linewidth]{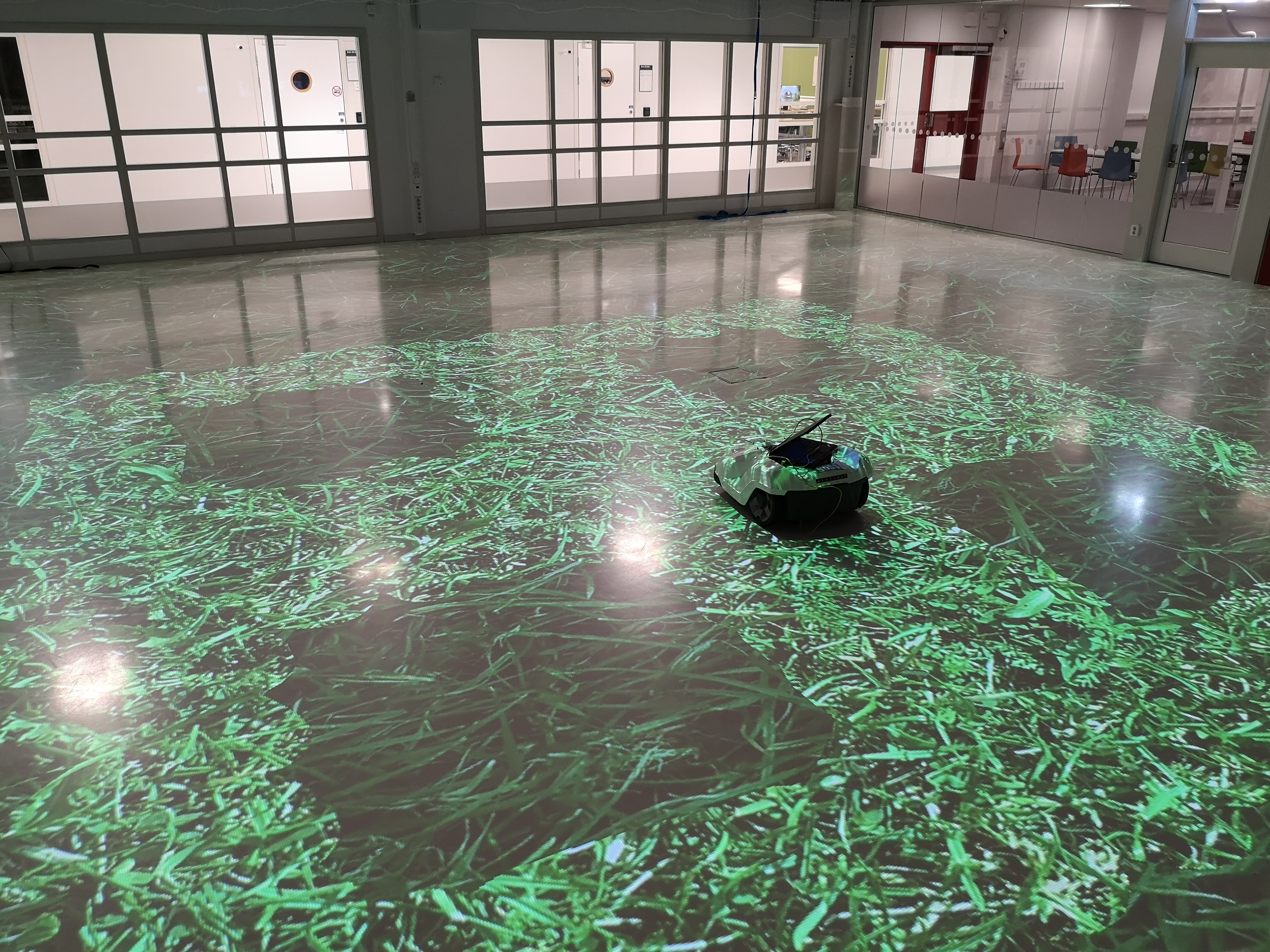}
    \caption{A picture of the robot having covered an environment with four square obstacles in our semi-virtual setting.}
    \label{fig_completed_four_obstacles}
\end{figure}

\subsection{Closing the Sim-to-Real Gap}
\label{sec_smoothening_sim_to_real_gap}




With the approach in the previous section, we could directly move to a real setting \textit{in-the-wild}. However, this would make training and fine-tuning cumbersome as manual interaction would be required to supervise the training process. In order to fully automatically (continue to) train the system, we propose to use a semi-virtual setup, see Fig.~\ref{fig_completed_four_obstacles}. In this setup, we use the real robot with its kinematics and dynamics, but simulate its lidar sensor and obstacles, and localize the robot using a positioning system. The environment, including obstacles and coverage, e.g.\ from mowing, is visualized by a projection onto the ground.

\begin{table*}[!t]
    \centering
    \caption{Physical dimensions for the three different CPP variations in simulation.}
    \begin{tabular}{lccc}
        \toprule
        & Omnidirectional & Non-omnidirectional & \multirow{2}{*}{Lawn mowing} \\
        & exploration & exploration & \\
        \midrule
        Coverage radius & $7$ m & $3.5$ m & $0.15$ m \\
        Agent radius & $0.08$ m & $0.15$ m & $0.15$ m \\
        Maximum linear velocity & $0.5$ m/s & $0.26$ m/s & $0.26$ m/s \\
        Maximum angular velocity & $1$ rad/s & $1$ rad/s & $1$ rad/s \\
        Simulation step size & $0.5$ s & $0.5$ s & $0.5$ s \\
        Lidar rays & $20$ & $24$ & $24$ \\
        Lidar range & $7$ m & $3.5$ m & $3.5$ m \\
        Lidar field-of-view & $360^\circ$ & $180^\circ$ & $180^\circ$ \\
        \bottomrule
    \end{tabular}
    \label{tab_physical_dimensions}
\end{table*}

The main purpose of using such an environment is to evaluate, in a controlled manner, the ability of the learning algorithm to generalize to a real robot in a real-time setting. We can study how the agent adapts to real-world aspects, such as wheel slip, that occur here. Moreover, as we will see in Section \ref{sec_simulation_experiments}, the proposed RL approach is robust to sensing and localization noise, and can be extended to the case of unknown pose, e.g.\ by estimating the pose using an off-the-shelf SLAM method. Thus, using our semi-virtual environment, we can draw conclusions about how we expect our approach to generalize to a fully realistic setting.

As mentioned, this semi-virtual setup enables fully automatic RL with the real robot. In particular, if the robot would drive towards one of the walls, both the robot and the environment can be moved back to the middle of the room. Furthermore, since we can automatically change or generate new environments each episode, we can train general CPP policies for completely unseen environments, and not just for a single fixed environment. Finally, the setup can also be used for fully automatic benchmarking of the learned model.

To further reduce the sim-to-real gap, we improve the fidelity of the simulator by taking into consideration the latencies induced by inertia and action delay. We measure the maximum linear acceleration, maximum angular acceleration, and action delay of the real system, and include these aspects in the simulated kinematics and dynamics.

To account for higher-order Markovian dynamics, we include information from previous time steps in the observation space.
While the common approach is to stack several previous observations \cite{haarnoja2019learning,mnih2013playing}, it is less feasible in our setting for multiple reasons.
(1)~Since the pose is embedded in the egocentric maps, rapid rotations significantly alter the observed state, making it difficult to learn the dynamics. (2)~It significantly increases the model size and processing time, which are critical aspects for real-time applications. (3)~It severely limits the replay buffer size, as the map observations are fairly large. Instead, we use a history of the previous actions, following previous work \cite{haarnoja2019learning,mnih2013playing}. This is lightweight and avoids the listed problems, and should be sufficient for learning the dynamics of the robot. Another option would be to use velocity estimates, although they can be highly noisy, in contrast to action observations. Using action observations, the agent can learn to estimate the velocity if necessary.

\subsection{Optimal Strategy for Going Sim-to-Real}

When transferring the CPP model from the simulation to the semi-virtual environment, different levels of fine-tuning can be performed and the training in simulation can happen with or without higher-order dependencies in the Markov process. 
However, training in simulation with a first-order assumption with subsequent fine-tuning does not make too much sense because fine-tuning will always be subject to higher-order effects in the real system.

Thus, one special case is the transfer without fine-tuning, as this can use a first-order model trained in simulation with arbitrary time-steps under benchmarking. In contrast, the higher-order model implies a certain step-length during benchmarking (and fine-tuning).

Key questions are thus:

\begin{enumerate}[label={(\arabic*)}]
    \item How does the first-order model trained solely in simulation work on the real robot, dependent on the step-length?
    \item How does the higher-order model trained solely in simulation work on the real robot and in comparison to (1)?
    \item How does the model with fine-tuning on the real robot perform in comparison to (2) and depend on the number of training steps?
\end{enumerate}
The working hypothesis is that (3) outperforms (2), and that (1) approximates (2) with sufficiently small step size.

\begin{figure*}[!t]
    \centering
    \setlength{\tabcolsep}{0.5pt}
    \setlength{\fboxsep}{0pt}%
    \setlength{\fboxrule}{0.5pt}%
    \begin{tabular}{cccccccc}
        \fbox{\includegraphics[width=.12\linewidth]{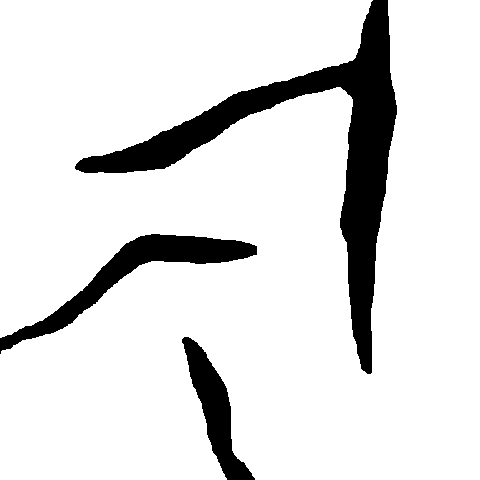}} &
        \fbox{\includegraphics[width=.12\linewidth]{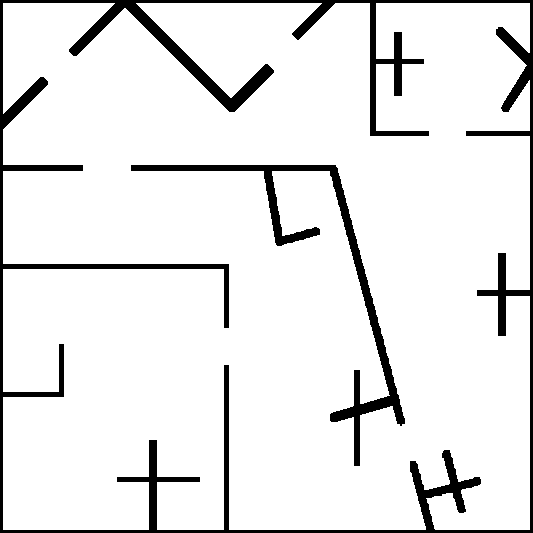}} &
        \fbox{\includegraphics[width=.12\linewidth]{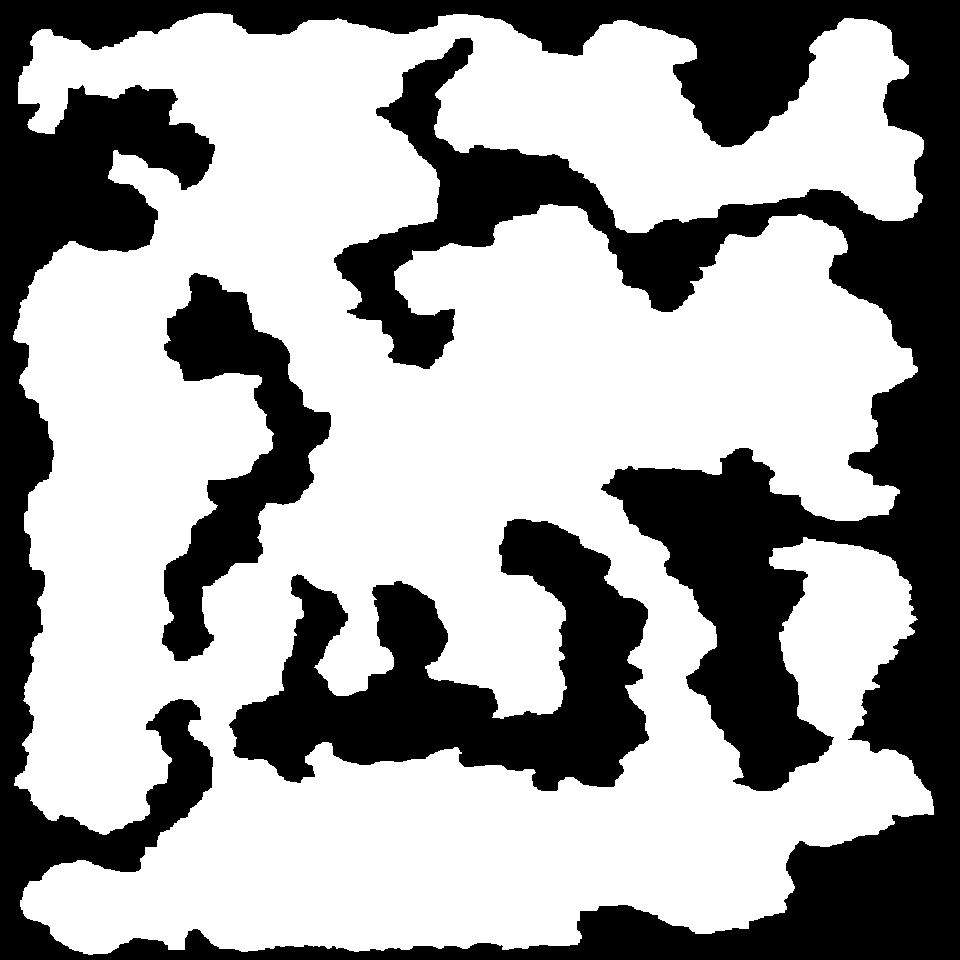}} &
        \fbox{\includegraphics[width=.12\linewidth]{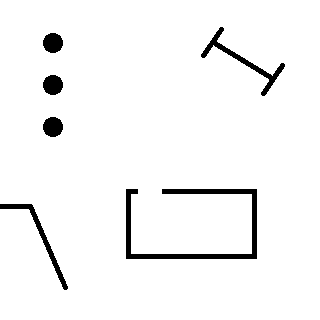}} &
        \fbox{\includegraphics[width=.12\linewidth]{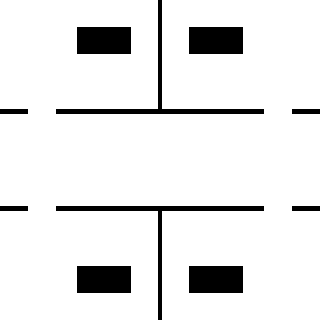}} &
        \fbox{\includegraphics[width=.12\linewidth]{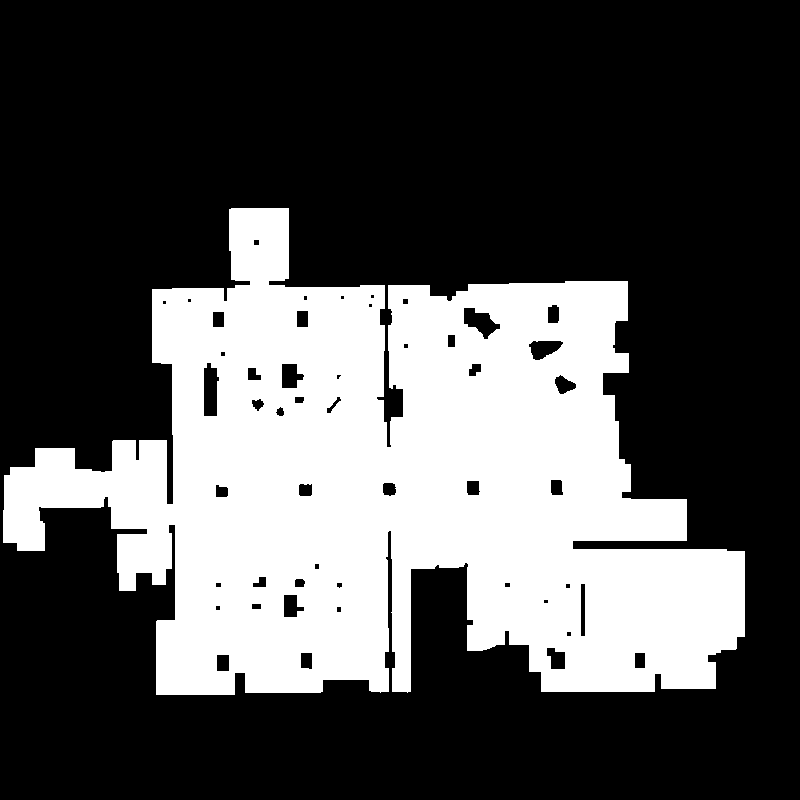}} &
        \fbox{\includegraphics[width=.12\linewidth]{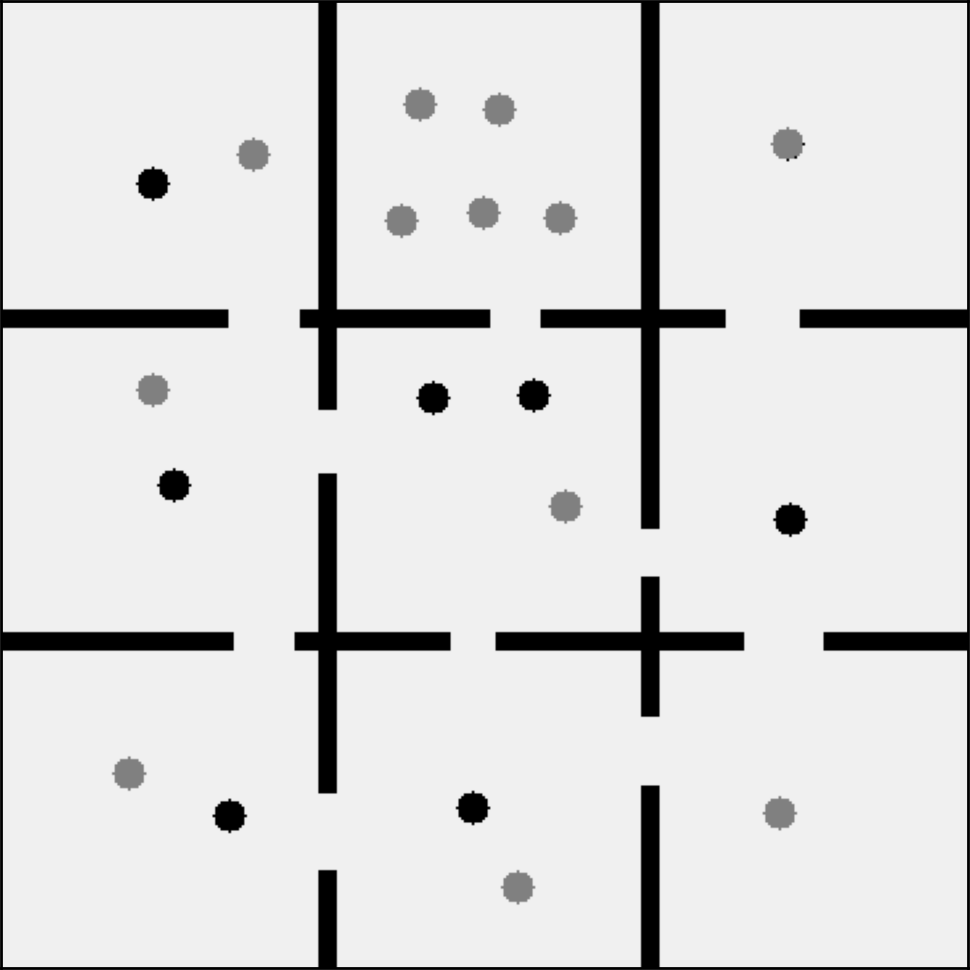}} &
        \fbox{\includegraphics[width=.12\linewidth]{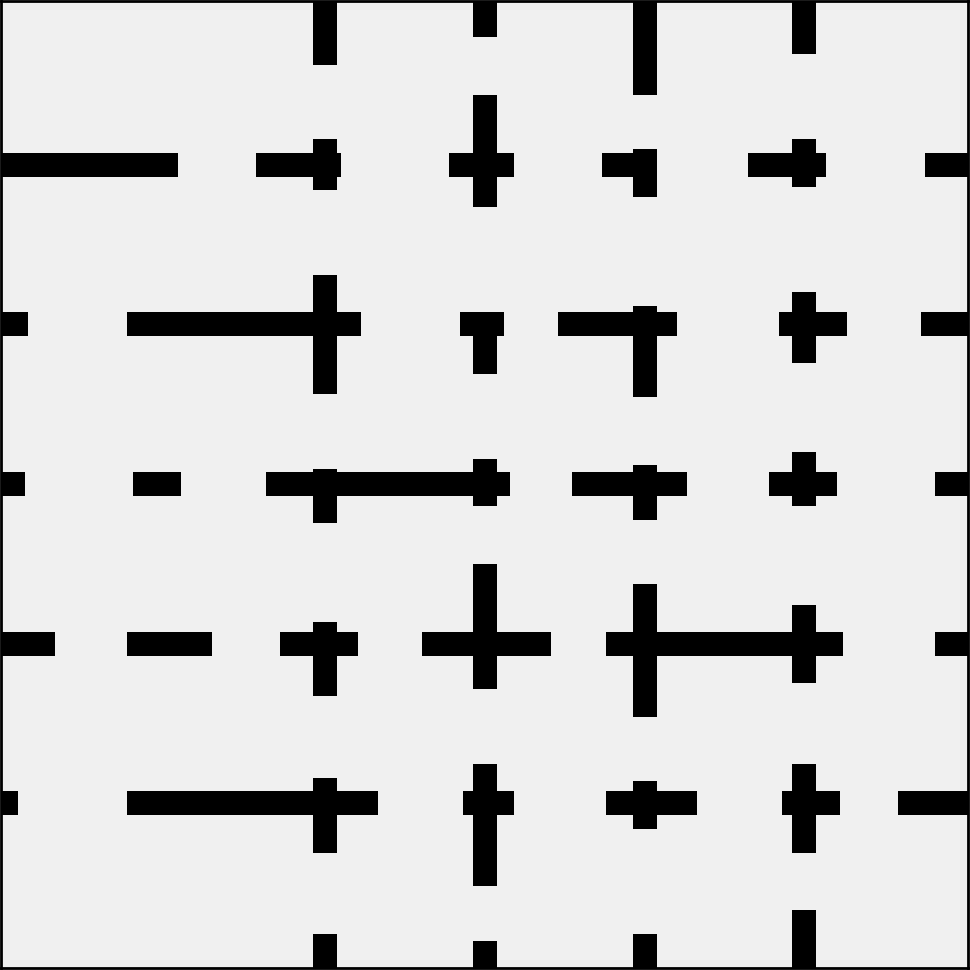}} \\
        (a) & (b) & (c) & (d) & (e) & (f) & (g) & (h) \\
    \end{tabular}
    \caption{Examples of exploration maps (a-c), lawn mowing maps (d-f), and randomly generated maps (g-h).}
    \label{fig_maps}
\end{figure*}

\section{Experiments}
\label{sec_experiments}

In this section, we present our experimental results, both in terms of rigorous comparisons and ablations in simulation, as well as sim-to-real transfer to our semi-virtual setting.

\subsection{Implementation Details}
\label{section_implementation_details}

\textbf{Simulation training details.} We first evaluate our RL approach in a simulated 2D environment. We utilize soft actor-critic learning \cite{haarnoja2018soft_2}, and train for $2$-$8$M iterations with learning rate $2 \cdot 10^{-5}$, batch size $256$, replay buffer size $5 \cdot 10^5$, and discount factor $\gamma = 0.99$. We evaluate our method on three settings: \textit{omnidirectional} and \textit{non-omnidirectional exploration}, as well as on the \textit{lawn mowing} task. For physical dimensions in the different settings, see Table \ref{tab_physical_dimensions}. The training time for one agent varied between $25$ to $150$ hours on a T4 GPU and a 6226R CPU.

\textbf{Real-world training details.} In our real-world experiments, we evaluate our approach on the \textit{lawn mowing} task. The fine-tuning on the physical robot is conducted using a Husqvarna Research Platform (HRP) (MIT software licence) \cite{husqvarna2017hrp}. This is a robotic lawnmower equipped with special firmware that allows it to be controlled through ROS. We equip the HRP with an Nvidia Jetson AGX Orin development kit. The wheel dimensions in \eqref{eq_left_right_velocities} are $r_w=12.25$ cm and $b=46.5$ cm. The training algorithms, based on the Stable-Baselines3 implementation (MIT license) \cite{raffin2021stable}, are executed on the Jetson, with control signals sent to the HRP. The experimental environment is a $12 \times 12$ meter indoor research arena. The agent receives its position from a high-precision Qualisys motion capture system \cite{qualisys2023qtm}, which comprises 12 Oqus 700+ and 8 Arqus A12 cameras. We use a $360^\circ$ simulated lidar field-of-view, $24$ lidar rays, $3.5$~m lidar range, $0.15$ m coverage radius, $0.26$ m/s maximum linear velocity, and $1$ rad/s maximum angular velocity. To improve the fidelity of the simulator during our sim-to-real experiments, we include a $0.5$ m/s$^2$ maximum linear acceleration, $2$ rad/s$^2$ maximum angular acceleration, $50$ ms action delay, and include the $10$ latest actions in the observation. For the real-world fine-tuning, we lower the learning rate to $5 \cdot 10^{-6}$ to avoid potential instability issues. In order not to give a disproportionately large weight to the first environment steps, we perform pure data collection during the first $5000$ steps, i.e.\ without any model updates.

\textbf{Environment details.} Based on initial experiments, we find suitable reward parameters. The episodes are prematurely truncated if $\tau = 1000$ consecutive steps have passed without the agent covering any new space. We set the maximum coverage reward $\lambda_\mathrm{area} = 1$, the incremental TV reward scale $\lambda_\mathrm{TV}^\mathrm{I} = 0.2$ for exploration and $\lambda_\mathrm{TV}^\mathrm{I} = 1$ for lawn mowing, the collision reward $R_\mathrm{coll} = -10$, and the constant reward $R_\mathrm{const} = -0.1$. Note that $\lambda_\mathrm{TV}^\mathrm{I}$ is the only hyperparameter that differs between the two settings. The global TV reward scale was set to $\lambda_\mathrm{TV}^\mathrm{G} = 0$ as it did not contribute to a performance gain in our ablations in Section \ref{sec_ablation_study}. For the multi-scale maps, we use $m = 4$ scales with $32 \times 32$ pixel resolution, a scale factor of $s = 4$, and $0.0375$ meters per pixel for the finest scale. Thus, the maps span a square with side length $76.8$~m. For more details regarding the choice of hyperparameters, see Appendix \ref{supp_sec_hyperparameters}. We apply curriculum learning, progressively increasing the environment complexity, and the goal coverage rate from $90\%$ to $99\%$ during training. For more details, see Appendix \ref{supp_sec_progressive_training}. To increase the variation of encountered scenarios, we use both a fixed set of maps and procedurally generated maps, by randomizing grid-like floor plans. See Fig. \ref{fig_maps} for examples, and Appendix \ref{supp_sec_random_map_generation} for more details. During real-world fine-tuning we use both fixed and randomized training maps that fit within the research arena, and set the goal coverage rate to $99\%$ for episode termination.

\begin{table*}[!t]
    \setlength{\tabcolsep}{5pt}
    \centering
    \caption{Time in seconds for reaching $90\%$ and $99\%$ coverage on Explore-Bench in simulation. Our method surpasses both frontier-based methods and a recent RL-based approach using active neural SLAM.}
    \begin{tabular}{lcccccccccccccc}
        \toprule
        Map $\rightarrow$ & \multicolumn{2}{c}{Loop} & \multicolumn{2}{c}{Corridor} & \multicolumn{2}{c}{Corner} & \multicolumn{2}{c}{Rooms} & \multicolumn{2}{c}{Comb.\ 1} & \multicolumn{2}{c}{Comb.\ 2} & \multicolumn{2}{c}{Total} \\
        \cmidrule(l{\tabcolsep}r{\tabcolsep}){2-3}
        \cmidrule(l{\tabcolsep}r{\tabcolsep}){4-5}
        \cmidrule(l{\tabcolsep}r{\tabcolsep}){6-7}
        \cmidrule(l{\tabcolsep}r{\tabcolsep}){8-9}
        \cmidrule(l{\tabcolsep}r{\tabcolsep}){10-11}
        \cmidrule(l{\tabcolsep}r{\tabcolsep}){12-13}
        \cmidrule(l{\tabcolsep}r{\tabcolsep}){14-15}
        Method $\downarrow$ & $T_{90}$ & $T_{99}$ & $T_{90}$ & $T_{99}$ & $T_{90}$ & $T_{99}$ & $T_{90}$ & $T_{99}$ & $T_{90}$ & $T_{99}$ & $T_{90}$ & $T_{99}$ & $T_{90}$ & $T_{99}$ \\
        \midrule
        Distance frontier & 124 & 145 & 162 & 169 & 210 & 426 & 159 & 210 & 169 & 175 & 230 & 537 & 1054 & 1662 \\
        RRT frontier & 145 & 180 & 166 & 170 & 171 & 331 & 176 & 211 & 141 & 192 & 249 & 439 & 1048 & 1523 \\
        Potential field frontier & 131 & 152 & 158 & 162 & 133 & 324 & 156 & 191 & 165 & 183 & 224 & 547 & 967 & 1559 \\
        Active Neural SLAM & 190 & 214 & 160 & 266 & 324 & 381 & 270 & 315 & 249 & 297 & 588 & 755 & 1781 & 2228 \\
        \textbf{Ours} & \textbf{89} & \textbf{101} & \textbf{98} & \textbf{128} & \textbf{80} & \textbf{291} & \textbf{82} & \textbf{99} & \textbf{87} & \textbf{93} & \textbf{120} & \textbf{349} & \textbf{556} & \textbf{1061} \\
        \bottomrule
    \end{tabular}
    \label{table_explore_bench}
\end{table*}

\textbf{First- vs higher-order policies.} In our experiments, we train two types of policies. For the first one, inertia and action delay are excluded from the simulation, i.e.\ the agent reaches the desired linear and angular velocities instantaneously, and the action takes effect immediately. Additionally, the observation space does not include any information about previous states. The action of this model is independent of the previous action sequence and only depends on the current state. Thus, we refer to it as a \textit{first-order} policy. For the second type of policy, we consider inertia by limiting the maximum linear and angular accelerations, and also include an action delay. This captures higher-order Markovian dynamics, as the next state depends on a finite set of previous states and actions. Policies trained in this setting also observe the 10 previous actions, and we refer to them as \textit{higher-order} policies.

\textbf{Evaluation.} To evaluate the various RL policies, we measure the times $T_{90}$ and $T_{99}$ to reach $90\%$ and $99\%$ coverage, respectively. We use these as the main metrics during ablations and comparisons with other methods, as the coverage time is usually important in CPP tasks, and common for benchmarking in the literature \cite{xu2022explore}. Some applications also value path length and accumulated rotations \cite{chen2019iros}, so we also provide these. For further analysis, we also measure the collision frequency, the average speed, and the learned entropy in our fine-tuning experiments. The evaluation is performed on maps that are not seen during training. For details on training and evaluation maps used in simulation and in the real world, see Appendix~\ref{supp_sec_maps}. Qualitative results in the form of learned paths can be found in Fig. \ref{fig_more_qualitative_paths}, and videos can be found online.\footnote{Videos can be found online at this link: \url{https://drive.google.com/drive/folders/1J0vpOBuRhCHOxsnAE1Vd1cZ6-qHPlvUh?usp=sharing}}

\subsection{Simulation Experiments}
\label{sec_simulation_experiments}

Before going to a realistic setting, we perform comparisons and ablation studies in simulation in this section. Here, we train first-order policies unless stated otherwise.

\begin{table*}[!t]
    \setlength{\tabcolsep}{5pt}
    \centering
    \caption{Time in minutes for reaching $90\%$ and $99\%$ coverage on the lawn mowing task in simulation.}
    \begin{tabular}{clcccccccccccccc}
        \toprule
        & Map $\rightarrow$ & \multicolumn{2}{c}{Map 1} & \multicolumn{2}{c}{Map 2} & \multicolumn{2}{c}{Map 3} & \multicolumn{2}{c}{Map 4} & \multicolumn{2}{c}{Map 5} & \multicolumn{2}{c}{Map 6} & \multicolumn{2}{c}{Total} \\
        \cmidrule(l{\tabcolsep}r{\tabcolsep}){3-4}
        \cmidrule(l{\tabcolsep}r{\tabcolsep}){5-6}
        \cmidrule(l{\tabcolsep}r{\tabcolsep}){7-8}
        \cmidrule(l{\tabcolsep}r{\tabcolsep}){9-10}
        \cmidrule(l{\tabcolsep}r{\tabcolsep}){11-12}
        \cmidrule(l{\tabcolsep}r{\tabcolsep}){13-14}
        \cmidrule(l{\tabcolsep}r{\tabcolsep}){15-16}
        Setting & Method $\downarrow$ & $T_{90}$ & $T_{99}$ & $T_{90}$ & $T_{99}$ & $T_{90}$ & $T_{99}$ & $T_{90}$ & $T_{99}$ & $T_{90}$ & $T_{99}$ & $T_{90}$ & $T_{99}$ & $T_{90}$ & $T_{99}$ \\
        \midrule
        \multirow{2}{*}{\textit{Offline}} & TSP & 45 & 49 & 45 & 51 & 43 & 48 & 55 & 61 & 27 & 30 & 110 & 122 & 325 & 361 \\
        & BSA & \textbf{30} & \textbf{35} & \textbf{29} & \textbf{35} & \textbf{31} & \textbf{36} & \textbf{34} & \textbf{41} & \textbf{17} & \textbf{23} & \textbf{88} & \textbf{100} & \textbf{229} & \textbf{270} \\
        \midrule
        \multirow{2}{*}{\textit{Online}} & TSP & 62 & 69 & 70 & 77 & 67 & 75 & 70 & 78 & 37 & 41 & 142 & 158 & 448 & 498 \\
        & \textbf{Ours} & \textbf{44} & \textbf{60} & \textbf{40} & \textbf{50} & \textbf{43} & \textbf{49} & \textbf{40} & \textbf{69} & \textbf{25} & \textbf{32} & \textbf{118} & \textbf{149} & \textbf{310} & \textbf{409} \\
        \bottomrule
    \end{tabular} 
    \label{table_mowing}
\end{table*}

\subsubsection{Omnidirectional Exploration}

In omnidirectional exploration, the agent observes its surroundings in all directions through a 360$^\circ$ field-of-view lidar sensor. For this setting, we evaluate our method on Explore-Bench \cite{xu2022explore}, which is a recent benchmark that implements challenging environments, where four methods have been evaluated by the authors. These include three frontier-based methods, namely a distance-based frontier method \cite{yamauchi1997frontier}, an RRT-based frontier method \cite{umari2017autonomous}, and a potential field-based frontier method \cite{yu2021smmr}. The fourth method is an RL-based approach, where Xu \textit{et al.} \cite{xu2022explore} train an RL model to determine a global goal based on active neural SLAM \cite{chaplot2020Learning}. The benchmark contains six environments; loop, corridor, corner, rooms, combination 1 (rooms with corridors), and combination 2 (complex rooms with tight spaces), which can be found in Appendix \ref{supp_sec_maps}. The results are presented in Table \ref{table_explore_bench}. Our approach surpasses the performance of both the frontier-based methods and Active Neural SLAM. This shows that learning control signals end-to-end with RL is, in fact, a suitable approach to CPP. Furthermore, Fig. \ref{fig_qualitative_paths} (left) shows that the agent has learned an efficient exploration path in a complex and obstacle-cluttered environment.

\subsubsection{Non-Omnidirectional Exploration}

\begin{figure}[t]
    \centering
    \includegraphics[width=0.8\linewidth]{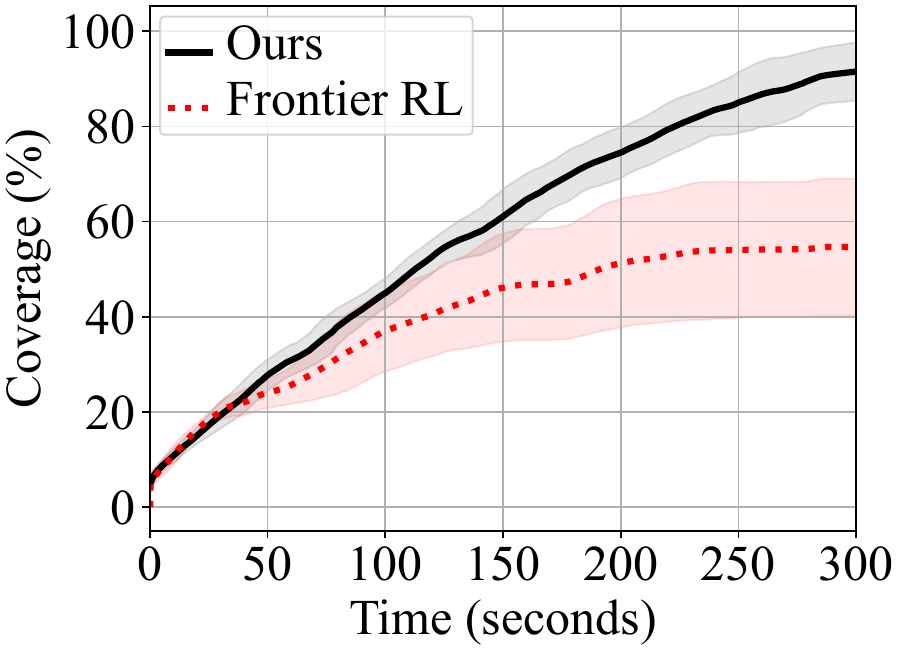}
    \caption{Simulation coverage over time in non-omnidirectional exploration.}
    \label{fig_non_omni}
\end{figure}

We further evaluate our method on non-omnidirectional exploration, with a 180$^\circ$ lidar field-of-view. We reimplement and compare with the recent frontier-based RL method of Hu \textit{et al.} \cite{hu2020tovt}, which was trained specifically for this setting. Their method uses RL to navigate to a chosen frontier node. The coverage over time is presented in Fig. \ref{fig_non_omni}. Our method outperforms the frontier-based RL approach, demonstrating that end-to-end learning of control signals is superior to a multi-stage approach for adapting to a specific sensor setup.

\subsubsection{Lawn Mowing}

For the lawn mowing task, we compare with the backtracking spiral algorithm (BSA) \cite{gonzalez2005icra}, which is a common benchmark for this CPP variation. Note however that BSA is an offline method and does not solve the mapping problem. As such, we do not expect our approach to outperform it in this comparison. Instead, we use it to see how close we are to a good solution. We further implement and compare with an offline and online version of a baseline that combines A* \cite{hart1968formal} with a traveling salesman problem (TSP) solver on nodes in a grid-like configuration. This has previously been proposed for the lawn mowing task in the offline setting for small environments \cite{bormann2018indoor}. To make it feasible for our larger environments, a heuristic was required for distant nodes, and periodic replanning was used for the online case, see Appendix \ref{supp_sec_implementation_compared_methods}. Table \ref{table_mowing} shows $T_{90}$ and $T_{99}$ for six maps numbered 1-6, which can be found in Appendix \ref{supp_sec_maps}. Compared to offline BSA, our method takes $35\%$ and $51\%$ more time to reach $90\%$ and $99\%$ coverage respectively. This is an impressive result considering the challenge of simultaneously mapping the environment. Moreover, our approach outperforms the online version of the TSP-based solution, and even surpasses the offline version for $90\%$ coverage. The limiting factors for TSP are likely the grid discretization and suboptimal replanning, which lead to overlap, see Appendix~\ref{supp_sec_implementation_compared_methods}.

\subsubsection{Ablation Study}
\label{sec_ablation_study}

\begin{figure}[t]
    \centering
    \includegraphics[width=0.8\linewidth]{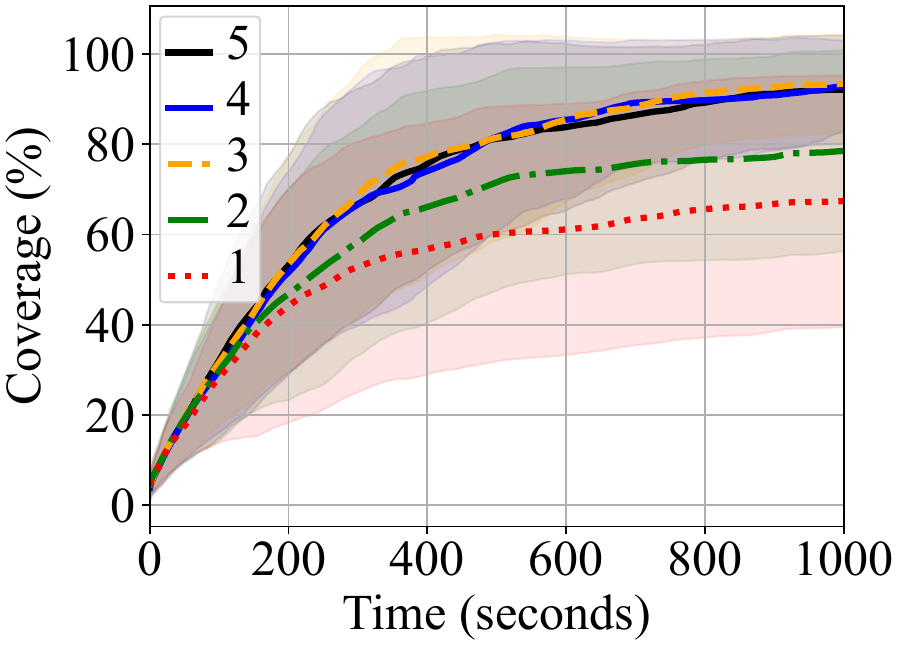}
    \caption{Exploration coverage for different number of scales in simulation.}
    \label{fig_num_maps}
\end{figure}

In Tables \ref{tab_training_step}, \ref{table_ablation}, and \ref{table_noise}, as well as in Fig. \ref{fig_num_maps}, we explore the impact of different components of our approach via a series of ablations. Since some baselines in Table \ref{table_ablation} struggled to reach $90\%$ or $99\%$ coverage, we provide the coverage at fixed times, instead of $T_{90}$ and $T_{99}$.

\begin{table}[t]
    \setlength{\tabcolsep}{8pt}
    \centering
    \caption{Training step size comparison in ms. $T_{90}$ and $T_{99}$: Average time in minutes for reaching $90\%$ and $99\%$ coverage in simulation.}
    \begin{tabular}{lcccc}
        \toprule
        & \multicolumn{2}{c}{First-order} & \multicolumn{2}{c}{Higher-order} \\
        \cmidrule(l{\tabcolsep}r{\tabcolsep}){2-3}
        \cmidrule(l{\tabcolsep}r{\tabcolsep}){4-5}
        $\Delta t$ & $T_{90}$ & $T_{99}$ & $T_{90}$ & $T_{99}$ \\
        \midrule
        150 & 7.1 & 12.7 & 7.6 & 13.7 \\
        250 & 6.4 & 9.3 & 6.4 & 10.1 \\
        500 & \textbf{6.0} & \textbf{9.1} & \textbf{5.8} & \textbf{8.5} \\
        1000 & 6.1 & 9.6 & 6.4 & 9.1 \\
        \bottomrule
    \end{tabular}
    \label{tab_training_step}
\end{table}

\begin{table}[t]
    \setlength{\tabcolsep}{3pt}
    \centering
    \caption{Coverage ($\%$) at 1500 and 1000 seconds for mowing (Mow) and exploration (Exp) respectively, comparing agent architecture (NN), TV rewards, and frontier map observation ($M_f$).}
    \begin{tabular}{ccccccc}
        \toprule
        \multicolumn{5}{c}{Settings} & \multicolumn{2}{c}{Coverage} \\
        \cmidrule(l{\tabcolsep}r{\tabcolsep}){1-5}
        \cmidrule(l{\tabcolsep}r{\tabcolsep}){6-7}
        &$R_\mathrm{TV}^{I}$ & $R_\mathrm{TV}^{G}$ & $M_f$ & NN & Mow & Exp \\
        \midrule
        \ding{172}&\checkmark & & \checkmark & MLP & 81.4 & 27.3 \\
        \ding{173}&\checkmark & & \checkmark & CNN & 93.2 & 88.4 \\
        \ding{174}&& & \checkmark & SGCNN & 85.1 & 91.5 \\
        \ding{175}&\checkmark & & & SGCNN & 72.6 & 80.6 \\
        \ding{176}&\checkmark & \checkmark & \checkmark & SGCNN & 96.6 & 89.0 \\
        \ding{177}&\checkmark & & \checkmark & SGCNN & \textbf{97.8} & \textbf{93.0} \\
        \bottomrule
    \end{tabular}
    \label{table_ablation}
\end{table}

\textbf{Optimal training step size.} In Table \ref{tab_training_step}, we investigate the impact of the training step size in simulation, both for first-order and higher-order policies. These policies were evaluated on the real-world evaluation maps, see Appendix \ref{supp_sec_maps}. In both cases, $500$ ms seems to be the best choice, considering $90\%$ and $99\%$ coverage times. Thus, we choose $500$ ms for our experiments. If the step size is small, it may take longer to train in simulation since the computation time per step is the same. More steps are needed to complete the same number of episodes. In Table \ref{tab_training_step}, we used the same number of training steps, which meant that the number of episodes were lower for the smaller step sizes. Meanwhile, with a lower step size, future rewards are more strongly discounted, and thus, long-term planning might be limited. It is possible that this could be remedied by tuning hyperparameters, such as the discount factor, $\gamma$. We simply used the default value of $\gamma=0.99$ for SAC \cite{haarnoja2018soft}. Finally, a lower step size requires a higher number of good consecutive actions to a achieve a good result, which might inhibit learning. On the other hand, if the step size is too large, the short-term control is limited, as each action is applied over a longer time period. This requires the agent to be more precise in its actions, which makes the problem harder. A step size of $500$ ms seems to strike a good balance. Based on this result, the state selection wait time in Table \ref{tab_times} was chosen. Finally, a step size of $500$ ms allows us to perform $4$ model updates for each environment step in the real setting.

\textbf{TV rewards.} In Table \ref{table_ablation}, we find that the incremental TV reward has a major impact on the lawn mowing problem, increasing the coverage from $85.1\%$ to $97.8\%$ (\ding{174} vs.\ \ding{177}). It also affects exploration, but not to the same extent. This is reasonable, as the total variation in the coverage map is lower in this case. When visualizing the learned paths, we found that by increasing $\lambda_\mathrm{TV}^\mathrm{I}$ beyond the optimal value with respect to coverage time, the agent learned more compact paths. For example, it would zig-zag rectangular regions parallel to the short edge instead of the long edge, and in general try to prioritize the nearest empty space. This behavior resulted in longer coverage times due to an increased number of turns. However, these paths could be considered more visually pleasing, and in turn be more appealing in some applications. Finally, as global TV has less temporal variation, and behaves more like a constant reward, it turns out not to be beneficial for CPP (\ding{176} vs.\ \ding{177}).

\textbf{Agent architectures.} The inductive prior of the CNN architecture enables a better understanding of the environment, and outperforms the MLP baseline (\ding{173} vs.\ \ding{172}) in Table~\ref{table_ablation}. Meanwhile, our proposed SGCNN architecture, which groups the different scales and convolves them separately, further improves the performance compared to a naive CNN (\ding{177} vs.\ \ding{173}), which treats all scales as the same spatial structure. We believe that the reason for the increased performance is that, with SGCNN, the agent is not faced with the additional challenge of discerning how the channels are spatially related. Thus, the representation learning becomes easier, and the agent can learn coverage paths more efficiently.

\textbf{Multi-scale frontier maps.} We further find in Table \ref{table_ablation} that our proposed frontier map representation is crucial for achieving a high performance with the multi-scale map approach (\ding{175} vs.\ \ding{177}). The frontier representation efficiently highlights the locations of non-visited regions, which we believe is especially useful for long-term planning. Without this input feature, too much information is lost in the coarser scales, thus hindering long-term planning.

\textbf{Number of scales.} In Fig. \ref{fig_num_maps}, we compare the coverage for different number of scales on the exploration task. It shows that only using one or two scales is not sufficient, where at least three scales are required to represent a large enough environment to enable long-term planning. Meanwhile, increasing the number of scales even further does not hamper the performance, which shows that our approach can be applied to even larger environments than what have been considered in our experiments. The discrepancy was not as large for the lawn mowing problem, as it is more local in nature with a lower coverage radius.

\textbf{Robustness to noise.} As the real world is noisy, we evaluate how robust our method is to noise in Table \ref{table_noise}. We apply Gaussian noise to the position, heading, and lidar measurements during both training and evaluation. Thus, both the maps and the sensor data perceived by the agent contain noise. We consider three levels of noise and present $T_{90}$ and $T_{99}$ for the lawn mowing task. The result shows that, even under high levels of noise, our method still functions well. Our approach surpasses the TSP solution under all three noise levels.

\begin{table}[t]
    \setlength{\tabcolsep}{5pt}
    \centering
    \caption{Coverage time in minutes for different levels of noise on the lawn mowing task in simulation.}
    \begin{tabular}{ccccc}
        \toprule
        \multicolumn{3}{c}{Noise standard deviation} & \multicolumn{2}{c}{Time} \\
        \cmidrule(l{\tabcolsep}r{\tabcolsep}){1-3}
        \cmidrule(l{\tabcolsep}r{\tabcolsep}){4-5}
        Position & Heading & Lidar & $T_{90}$ & $T_{99}$ \\
        \midrule
        0.01 m & 0.05 rad & 0.05 m & 310 & 409 \\
        0.02 m & 0.1 rad & 0.1 m & 338 & 486 \\
        0.05 m & 0.2 rad & 0.2 m & 317 & 434 \\
        \bottomrule
    \end{tabular}
    \label{table_noise}
\end{table}

\subsubsection{Collision Statistics}
\label{sec_collision}

In most real-world applications, it is vital to minimize the collision frequency, as collisions can inflict harm on humans, animals, the environment, and the robot. To gain insights into the collision characteristics of our trained agents, we tracked the collisions during evaluation. For exploration, it varied between once every $100$-$1000$ seconds in simulation, which is roughly once every $50$-$500$ meters. For lawn mowing, it varied between once every $150$-$250$ seconds in simulation, which is roughly once every $30$-$50$ meters. These include all forms of collisions, even low-speed side collisions. The vast majority of the collisions were near-parallel, and we did not observe any head on collisions. The practical implications are very different between these cases. In the real setting, the collision frequency on the lawn mowing task on the six evaluation maps was measured to be once every $6.2$ minutes, or once every $65.4$ meters, on average. Again, the collisions were mainly from the side of the robot, and not head on.

\subsection{Real-world Experiments}

In this section, we fine-tune our CPP agent in our semi-virtual environment, and evaluate how different policies generalize to the real world.

\subsubsection{Comparison of First-order CPP Policies}

We first investigate how well policies that assume a first-order Markov process transfer to our semi-virtual environment. In this experiment, the observation space does not include any previous actions, and the policy is trained for $8$M steps in a simulated environment without a limit on the linear and angular accelerations, and does not include any action delay. We evaluate the policy in simulation and in the real environment before and after fine-tuning. Since this policy does not depend on the previous action sequence, we can run it at any frequency. We evaluate it at the training frequency and as fast as possible, i.e.\ without waiting during state selection, which was measured to 15 ms per time step.

By comparing coverage times between simulation and reality in Table \ref{table_mowing_1}, we see that the real results are slightly worse at the lower frequency. This is expected due to differences in kinematics and other sources of error. When increasing the inference frequency in the real world, we observe faster coverage times, even surpassing the simulation results. This is likely the case due to the fact that the agent can update its actions faster and quickly adapt to errors. Thus, it can navigate more efficiently and produce a smoother pattern. For the performance on the individual evaluation maps, see Appendix \ref{supp_sec_extended_coverage_real}.

However, after fine-tuning, we find that the performance degrades. Our hypothesis is that as the observation includes no higher-order information, the optimal action cannot be deduced. For example, if the robot completely changes the direction of rotation, it takes some time before the new rotation takes effect. The observation may not change much, while the optimal action does. Thus, the performance degrades, which may take a long time to recover from. This is problematic due to the lower \textit{training} frequency. Running the \textit{inference} at a high frequency partly circumvents this problem, explaining the high performance of the model without fine-tuning. While we did not evaluate the fine-tuned policy on a lower frequency, we expect it to perform worse as in the non fine-tuning case.

\begin{table}[t]
    \setlength{\tabcolsep}{5pt}
    \centering
    \caption{\textbf{First-order policy.} Average time in minutes for reaching $90\%$ and $99\%$ coverage on six evaluation maps. \textit{Env}: Evaluation environment. \textit{Steps}: Fine-tuning steps. Inference step size is in ms. Note: Training step size is always 500 ms.}
    \begin{tabular}{lllcc}
        \toprule
        Env & Steps & $\Delta t$ & $T_{90}$ & $T_{99}$ \\
        \midrule
        Sim & 0 & 500 & 6.0 & 9.1 \\
        Real & 0 & 500 & 7.0 & 10.6 \\
        \midrule
        Real & 0 & 15 & 5.4 & 8.3 \\
        Real & 80k & 15 & 6.3 & 10.9 \\
        \bottomrule
    \end{tabular}
    \label{table_mowing_1}
\end{table}

\begin{table}[t]
    \setlength{\tabcolsep}{5pt}
    \centering
    \caption{\textbf{Higher-order policy.} Average time in minutes for reaching $90\%$ and $99\%$ coverage on six evaluation maps. \textit{Env}: Evaluation environment. \textit{Steps}: Fine-tuning steps. Inference step size is in ms. Note: Training step size is always 500 ms.}
    \begin{tabular}{lllcc}
        \toprule
        Env & Steps & $\Delta t$ & $T_{90}$ & $T_{99}$ \\
        \midrule
        Sim & 0 & 500 & 5.8 & 8.5 \\
        Real & 0 & 500 & 7.3 & 10.3 \\
        Real & 60k & 500 & 6.7 & 10.2 \\
        \bottomrule
    \end{tabular}
    \label{table_mowing_2}
\end{table}

\subsubsection{Comparison of Higher-order CPP Policies}

Next, we conduct a new experiment to evaluate how higher-order policies compare. We train a new model for $2$M steps in simulation, where we include the past 10 actions in the observation, and limit the maximum linear and angular accelerations based on realistic estimates. Furthermore, we include the measured real-world overhead as an action delay in simulation to better align the simulation with reality. Since the 10 actions correspond to specific points back in time, we keep the inference time step the same as during training.

The new model can learn higher-order Markovian dynamics, and in Table \ref{table_mowing_2} we see that it transfers well when directly deployed from simulation, surpassing the first-order model for $99\%$ coverage under this lower inference frequency. Again, the real results are somewhat worse compared to the simulation results, as expected. Comparing Tables \ref{table_mowing_1} and \ref{table_mowing_2}, the higher-order policy surpasses the first-order policy in both $T_{90}$ and $T_{99}$ at $500$ ms after fine-tuning. For the performance on the individual evaluation maps, see Appendix \ref{supp_sec_extended_coverage_real}.

\begin{figure}[t]
    \centering
    \includegraphics[height=0.65\linewidth]{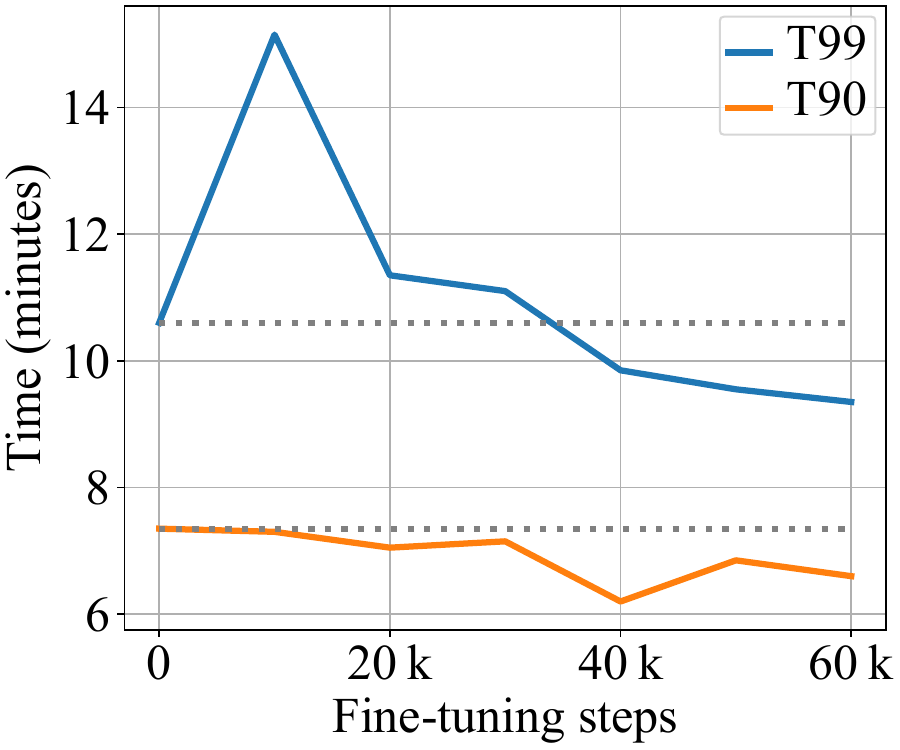}
    \caption{Average coverage times during fine-tuning on two evaluation maps. Dashed line: No fine-tuning.}
    \label{fig_fine_tune}
\end{figure}

\begin{figure}[t]
    \centering
    \includegraphics[height=0.65\linewidth]{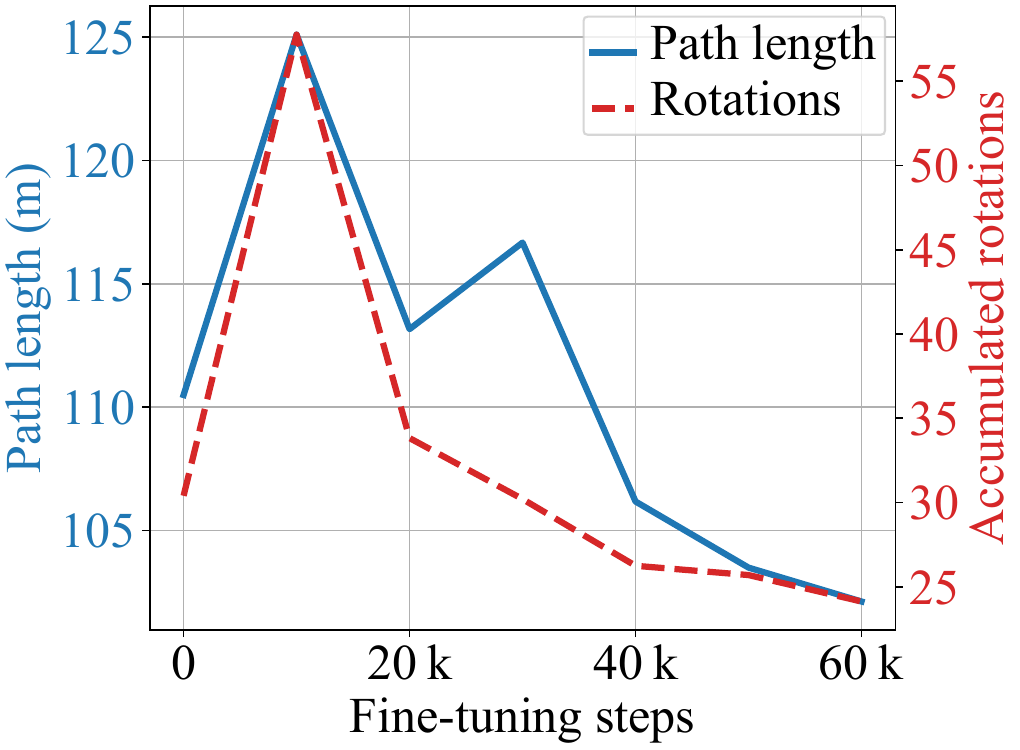}
    \caption{Average path length and accumulated full rotations during fine-tuning on two evaluation maps.}
    \label{fig_length_and_turns}
\end{figure}

In Fig. \ref{fig_fine_tune}, where we record the performance on a subset of the maps during fine-tuning, we observe that the performance degrades initially, but in contrast to the first-order model, the higher-order model surpasses its original performance. $T_{99}$ degrades heavily in the early stage, while $T_{90}$ remains more or less constant. This suggests that the agent initially sacrifices long-term planning in favour of low-level control. After it has adapted to the low-level controls, then the long-term planning also improves. Note that $T_{90}$ increases slightly after 40k iterations. It could be the case that the agent prioritizes the long-term goal over short-term performance towards the end of the fine-tuning, e.g.\ by choosing a less greedy path early on that reduces the final coverage time. Considering the general trend in the second half of Fig. \ref{fig_fine_tune}, we expect the performance to continue to improve with further fine-tuning steps.

To further analyze the fine-tuned policy, we measure the path length, accumulated rotations, the average speed, and the learned entropy coefficient. Fig. \ref{fig_length_and_turns} shows the path length and accumulated full rotations at $99\%$ coverage. The number of rotations were computed by summing the absolute differences in heading angle in every time step. Similar to the coverage time, both metrics degrade initially, and then improve over time, surpassing the initial performance. This shows that the agent finds a more efficient path after fine-tuning. It reduces the number of turns, which take time, while it overlaps its previous path less, thus reducing the total path length.

\begin{figure}[t]
    \centering
    \includegraphics[height=0.65\linewidth]{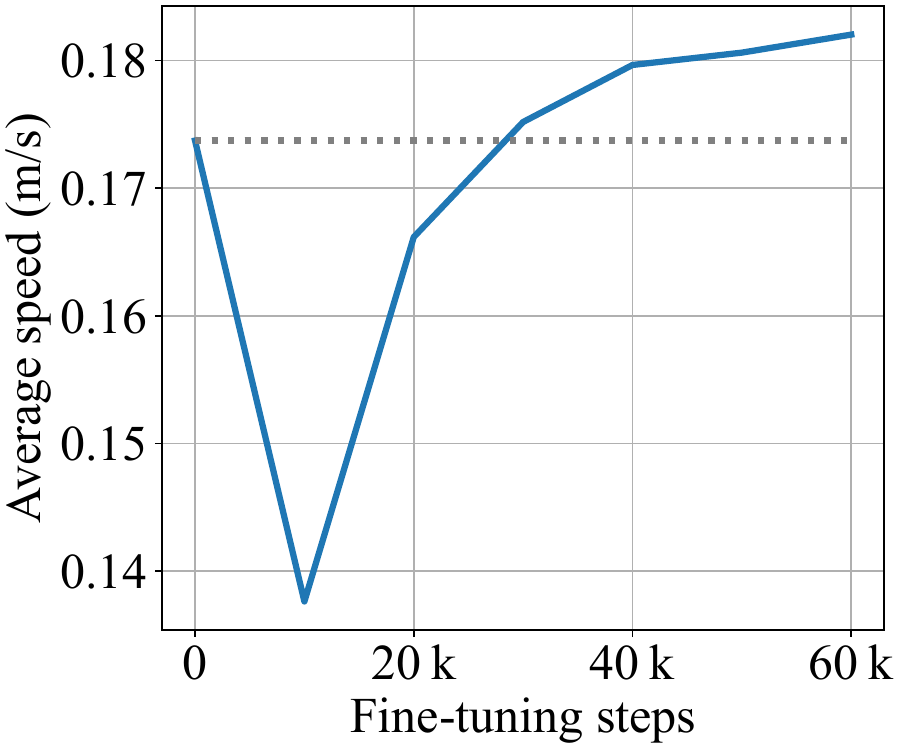}
    \caption{Average speed during fine-tuning on two evaluation maps. Dashed line: No fine-tuning.}
    \label{fig_speed}
\end{figure}

\begin{figure}[t]
    \centering
    \includegraphics[height=0.65\linewidth]{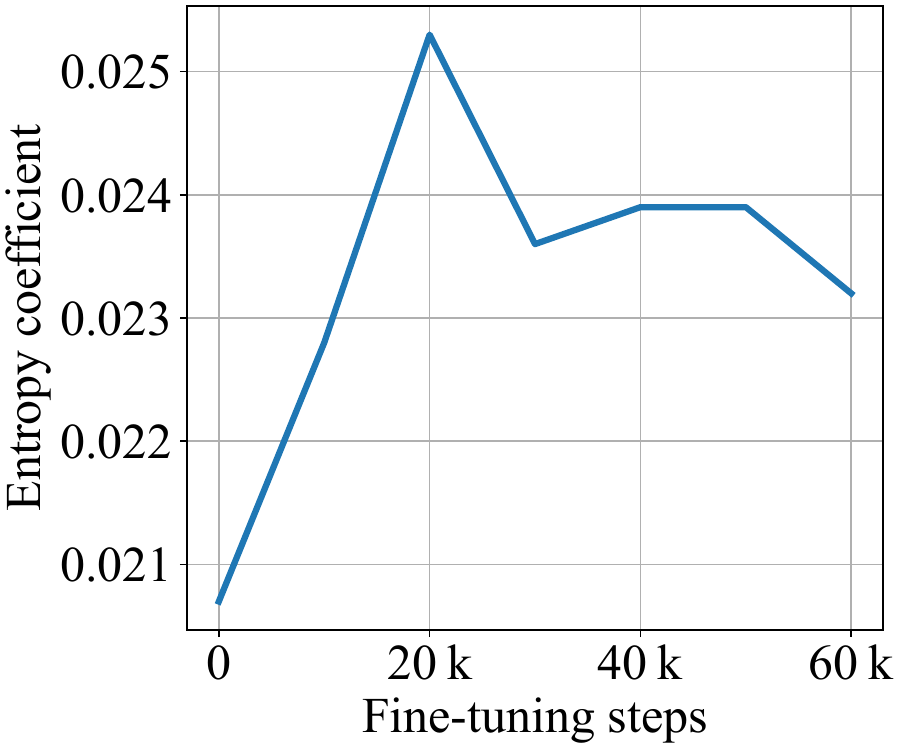}
    \caption{The learned entropy coefficient in soft actor-critic \cite{haarnoja2018soft_2} during fine-tuning.}
    \label{fig_entropy}
\end{figure}

\begin{figure*}[!t]
    \centering
    \setlength{\tabcolsep}{0.5pt}
    \setlength{\fboxsep}{0pt}%
    \setlength{\fboxrule}{0.5pt}%
    \begin{tabular}{cccccc}
        \includegraphics[width=.22\linewidth]{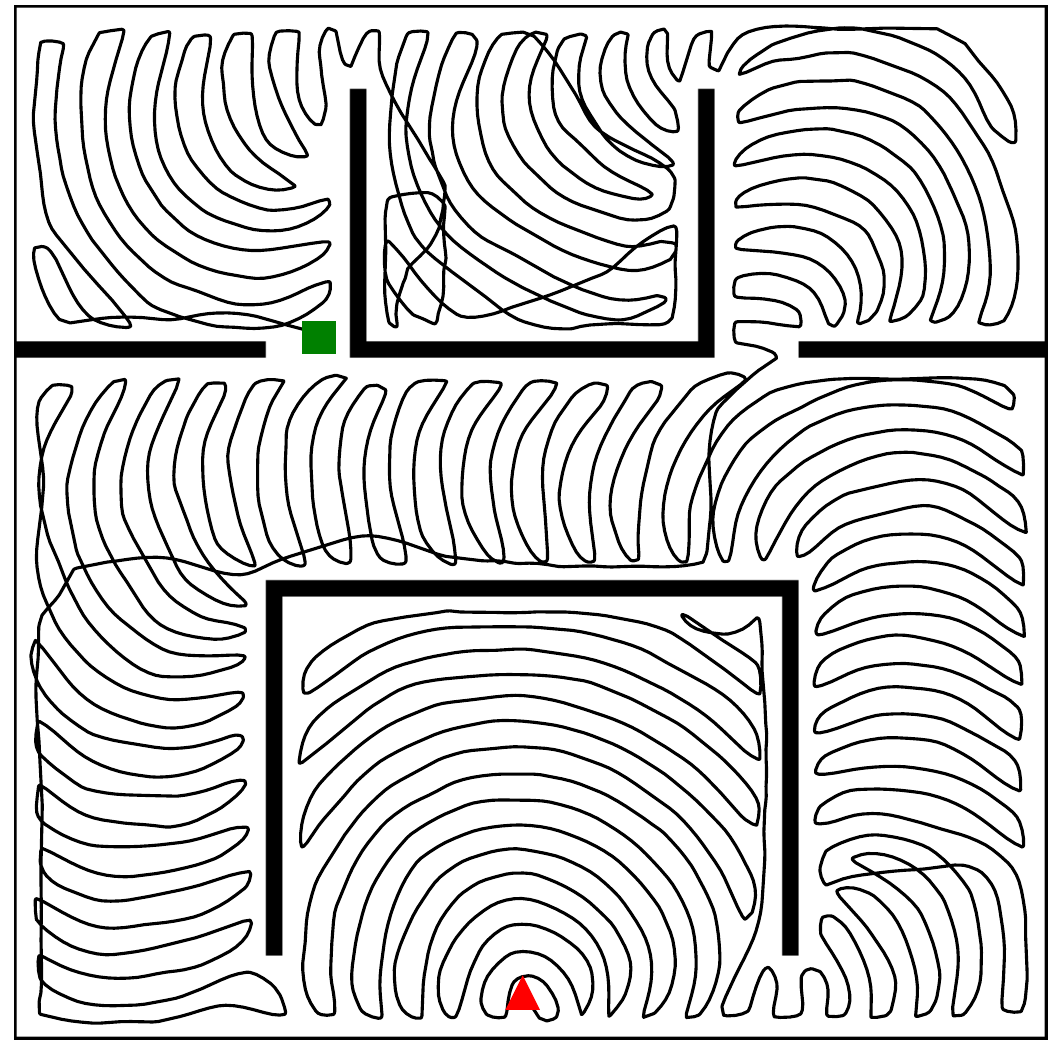} &
        \includegraphics[width=.22\linewidth]{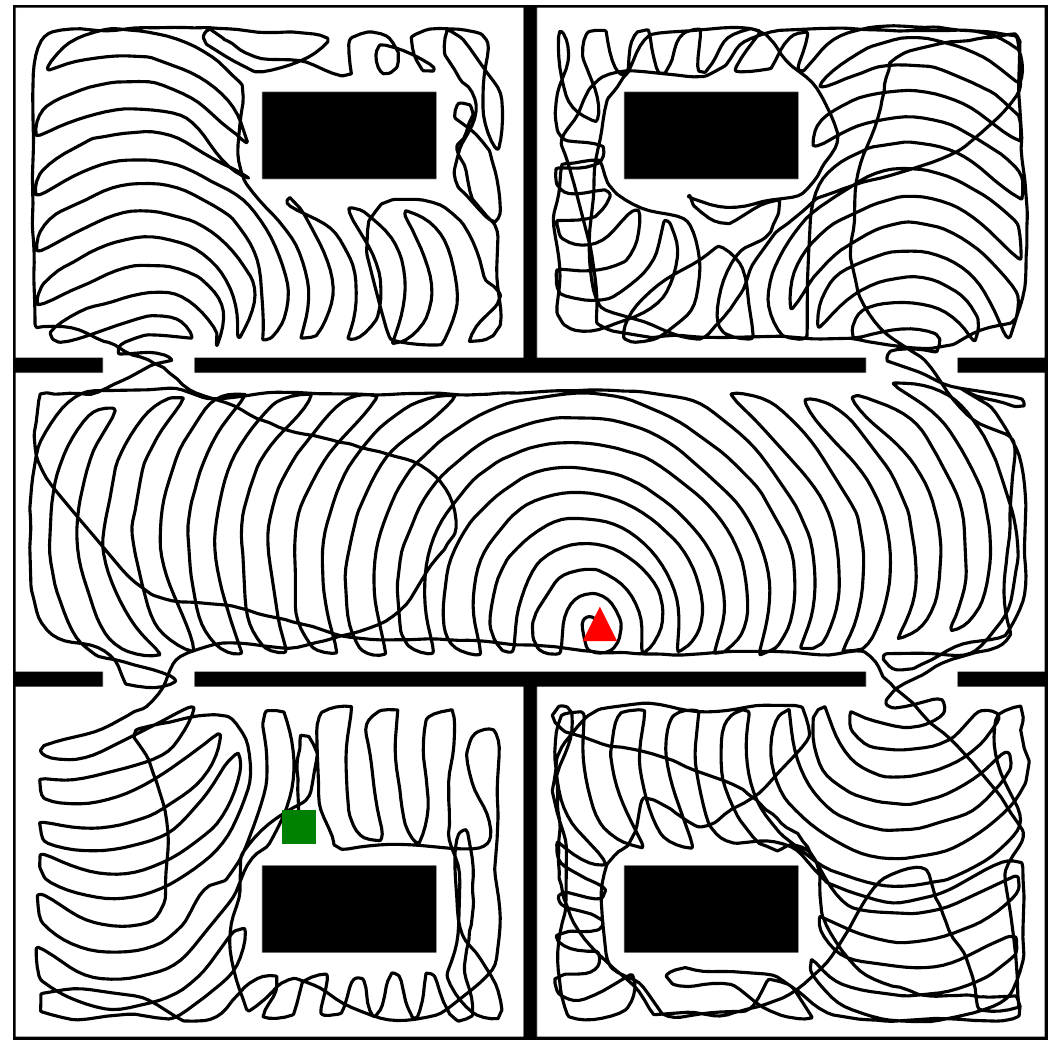} &
        \includegraphics[width=.22\linewidth]{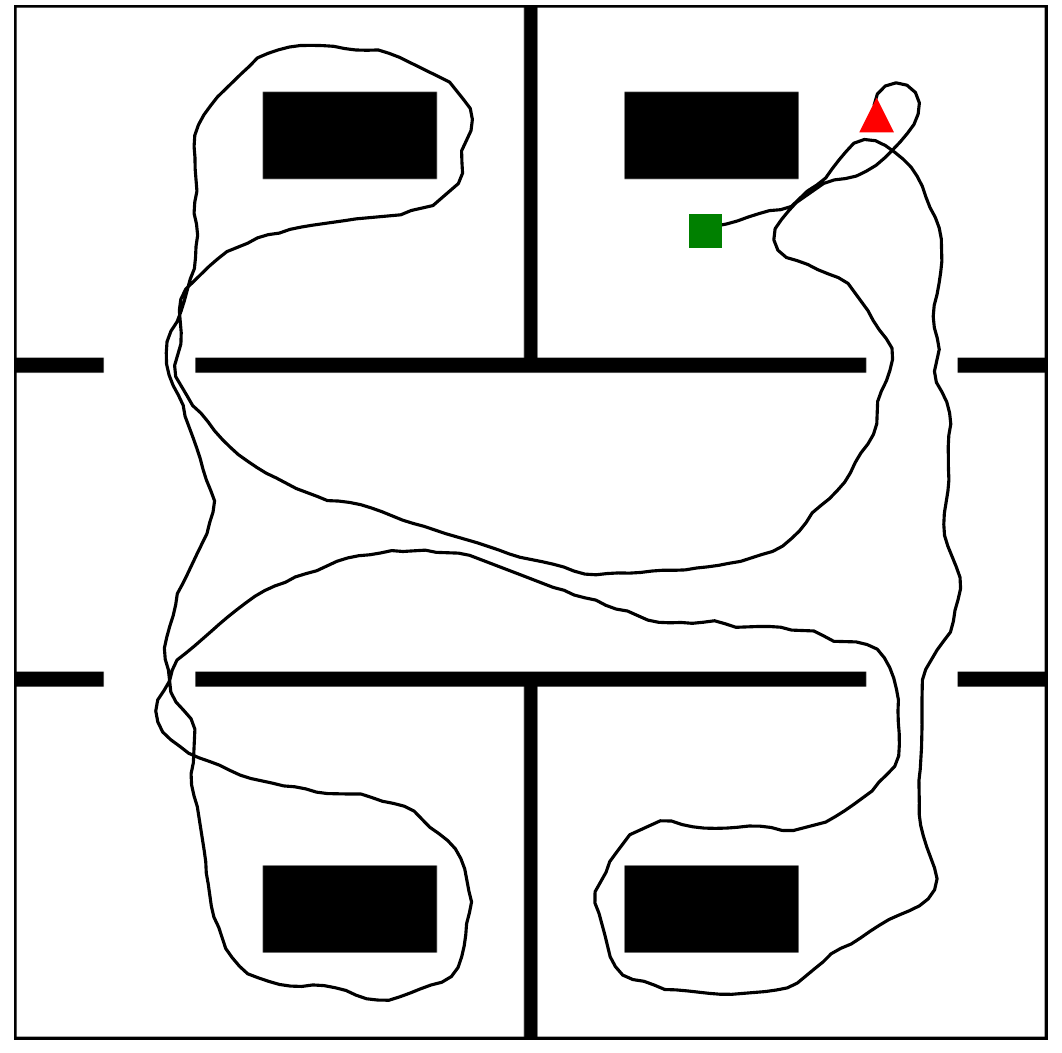} &
        \includegraphics[width=.22\linewidth]{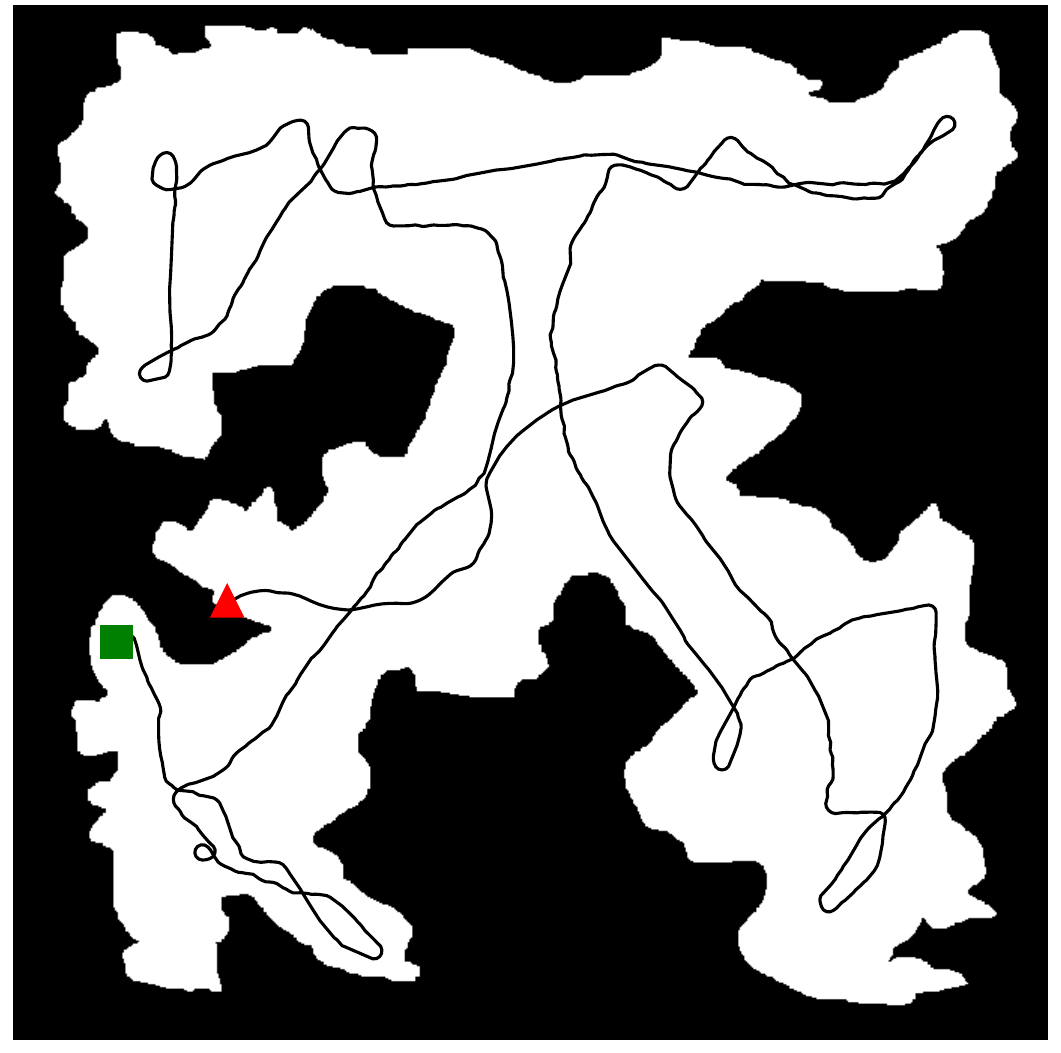} \\
        \includegraphics[width=.22\linewidth]{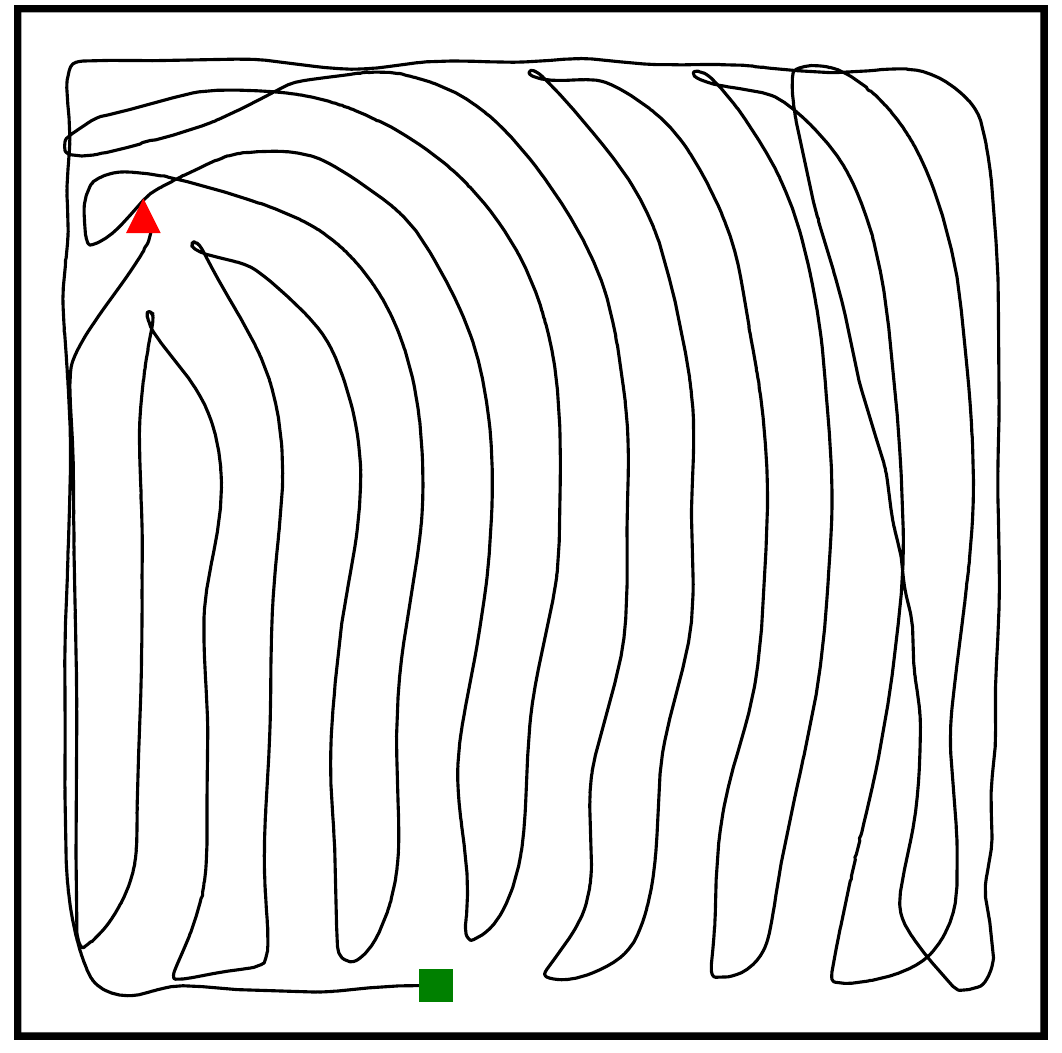} &
        \includegraphics[width=.22\linewidth]{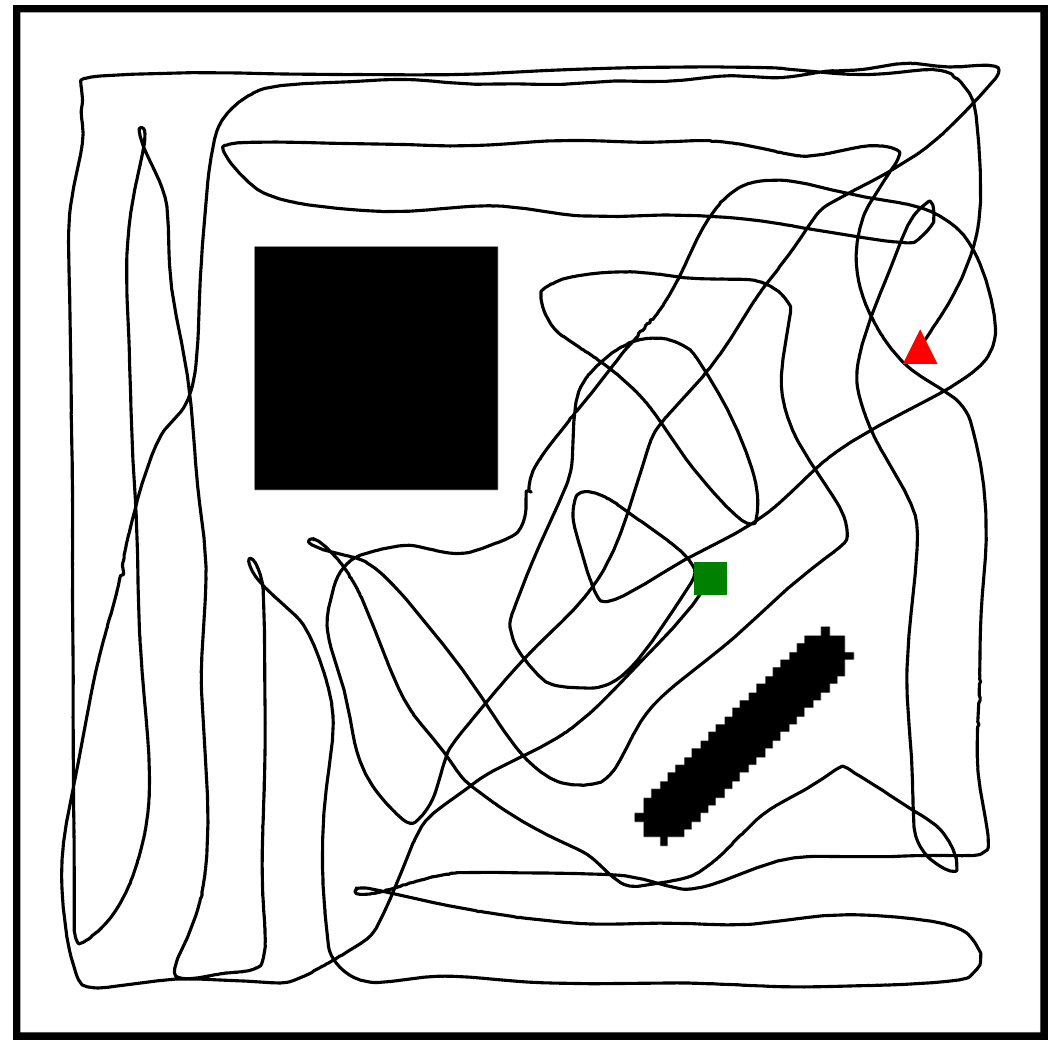} &
        \includegraphics[width=.22\linewidth]{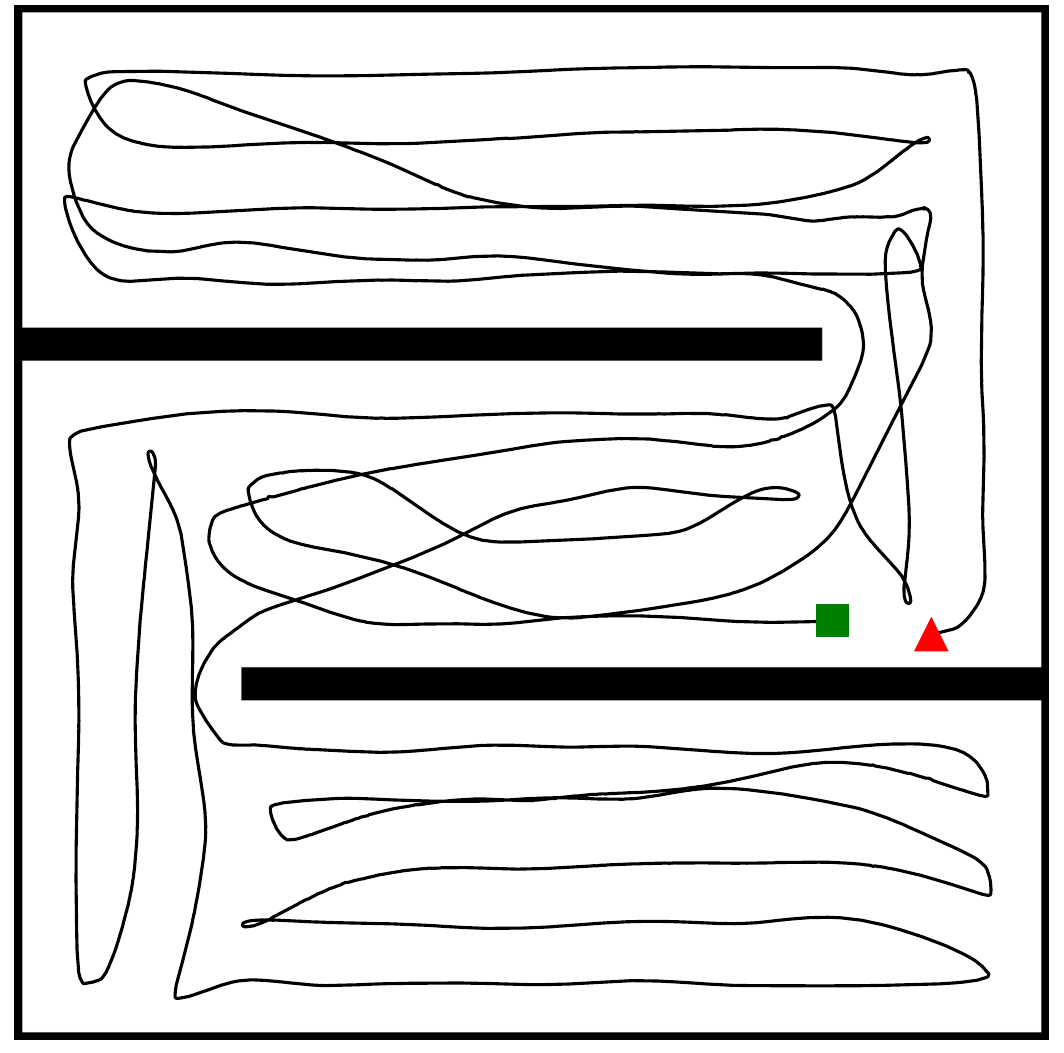} &
        \includegraphics[width=.22\linewidth]{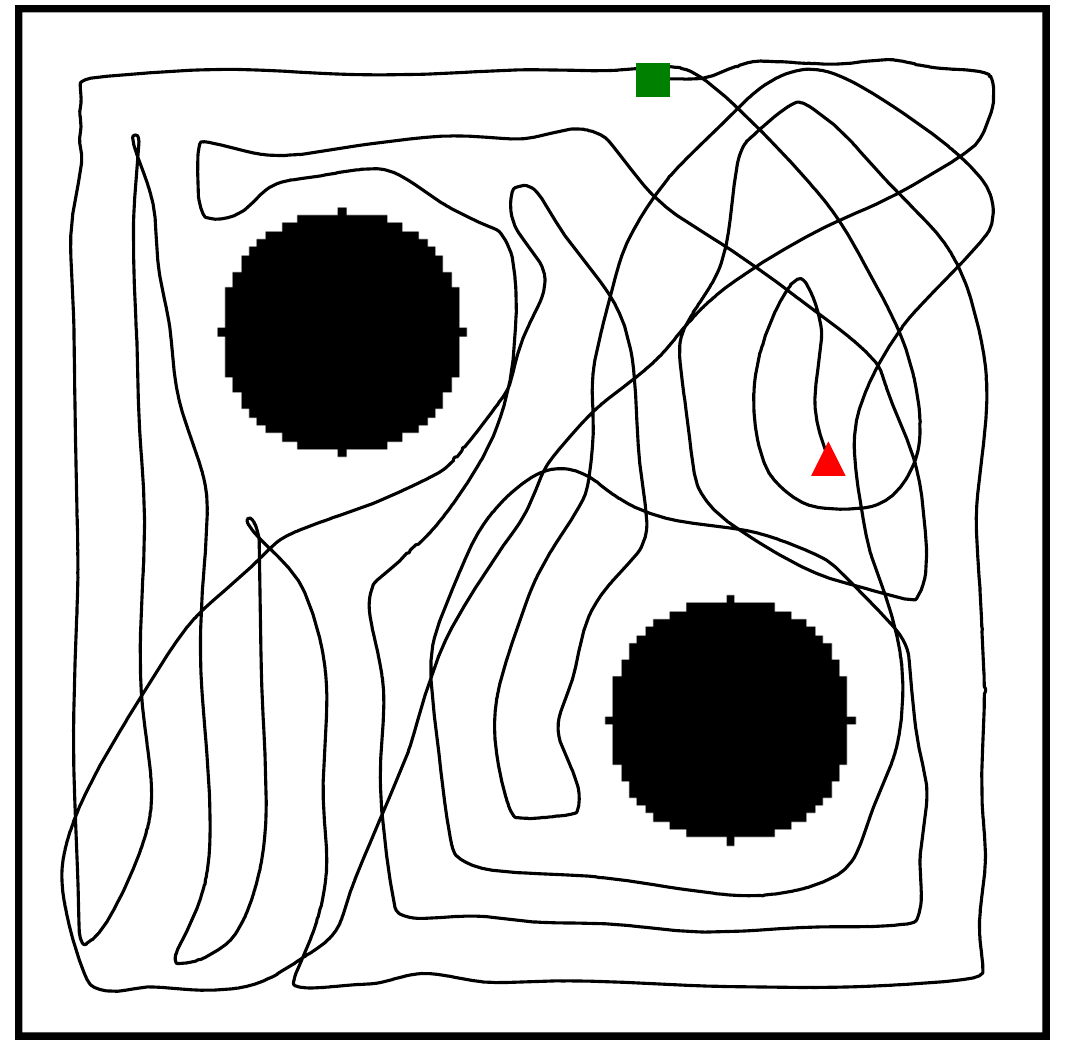} \\
        (a) & (b) & (c) & (d) & (e) \\
    \end{tabular}
    \caption{Learned paths for a fully trained agent, including the starting position (red triangle) and the end position (green square). Top row: Paths in simulation on the lawn mowing task (a-b) and on exploration (c-d). Bottom row: Paths in our semi-virtual environment on the lawn mowing task (a-d).}
    \label{fig_more_qualitative_paths}
\end{figure*}

In Fig. \ref{fig_speed}, we analyse the average speed of the agent during fine-tuning. The initial drop in performance is also evident here. In the initial fine-tuning phase, the agent attains a lower average speed, as a result of e.g.\ more collisions and many small turns, while it adjusts to the new environment dynamics. Over time, it learns to control the robot in a more efficient manner, allowing it to keep a higher average speed with less turns, which results in lower coverage times.

Fig. \ref{fig_entropy} shows the learned entropy coefficient in the soft actor-critic framework \cite{haarnoja2018soft_2}. It increases initially, suggesting that the agent prioritizes exploration over exploitation during the distribution shift to the real world. After the initial peak, the entropy coefficient decreases, putting more emphasis on exploitation. However, it does not reach the same level as in simulation, i.e.\ at $0$ fine-tuning steps. This either suggests that the real environment contains more uncertainty, or that the learning process has not entirely converged yet. It is likely a combination of the two.

\subsubsection{Generalization}

To evaluate the generalization capability of our approach, we perform a new exploration experiment, this time on a new robot platform. The new robot consists of an MiR100 platform integrated with a UR5e robotic arm (not used). The MiR platform has two driving wheels in the middle to control rotational and lateral movement, and four passive rotating wheels in each corner for stability. This allows for the linear and angular velocities to be controlled independently, and thus the predicted control signals can be passed directly to the robot platform, after proper renormalization. To further test the generalization capability, we change the sensor setup by severely limiting the visibility of the agent. We use the non-omnidirectional setup, and reduce the lidar field-of-view to $90^\circ$, i.e.\ $\pm45^\circ$, and the range to $2$ meters. The inference is executed as fast as possible, i.e.\ with a step size of $15$ ms.

\begin{figure}[t]
    \centering
    \includegraphics[width=0.57\linewidth]{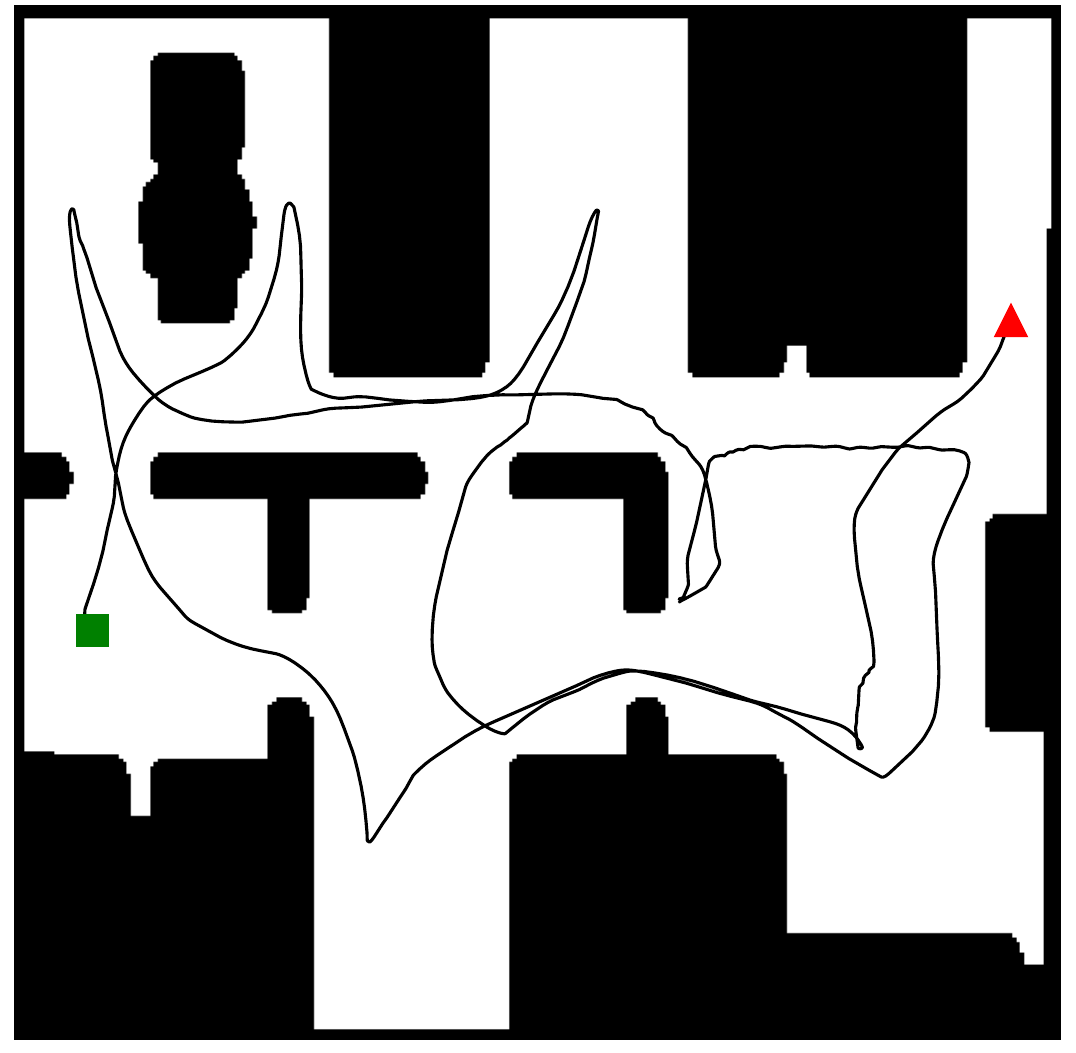}
    \caption{Coverage path on the MiR100 platform, including the start (red triangle) and end position (green square).}
    \label{fig_mir_path}
\end{figure}

\begin{table}[t]
    \setlength{\tabcolsep}{5pt}
    \centering
    \caption{Coverage time in seconds in simulation and on the MiR100 platform on exploration with limited field-of-view ($90^\circ$) and range ($2$ meters). The mean and standard deviation is computed over five runs.}
    \begin{tabular}{llccc}
        \toprule
        Environment & Robot & $T_{90}$ & $T_{99}$ \\
        \midrule
        Simulation   & -     & 163$\pm$\scriptsize{17} & 290$\pm$\scriptsize{27} \\
        Semi-virtual & MiR & 168$\pm$\scriptsize{23} & 339$\pm$\scriptsize{57} \\
        \bottomrule
    \end{tabular}
    \label{table_generalization}
\end{table}

We perform the evaluation on a suitable map for this setting, which is shown in Fig.\ \ref{fig_mir_path} together with a learned path. Table \ref{table_generalization} compares $T_{90}$ and $T_{99}$ between simulation and the MiR platform. The real-world results are slightly worse, which can be expected when considering the discrepancy between the two settings. For example, the maximum linear and angular accelerations were lower on the MiR platform compared to the HRP, which was not accounted for in the simulation. However, the coverage results are within a reasonable margin, which demonstrates that our approach can indeed generalize to different sensor setups and robot platforms.

\subsubsection{Energy Efficiency}

Due to the lightweight nature of our proposed SGCNN architecture, it is well suited for real-time applications. With only $800$k parameters, it is significantly smaller and faster than popular computer vision architectures, see Appendix~\ref{supp_sec_agent_architecture}. Meanwhile, our full computational framework is energy efficient. When measuring the energy consumption on the battery powered Jetson module, it never exceeded 15 W during training, and drawing roughly 10 W during inference. Note that, while training requires more computational power, such as a GPU, the computational requirements at inference time are much lower, and can even be done on a CPU. In total, the fine-tuning took roughly $8$ hours, which is only a small fraction of the weeks it would have taken to train from scratch.

\section{Conclusions}

We present a method for online coverage path planning in unknown environments, based on a continuous end-to-end deep reinforcement learning approach. The agent implicitly solves three tasks simultaneously, namely \textit{mapping} of the environment, \textit{planning} a coverage path, and \textit{navigating} the planned path while avoiding obstacles. For a scalable solution, we propose to use multi-scale maps, which is made viable through frontier maps that preserve the existence of non-covered space in coarse scales. We propose a novel reward term based on total variation, which significantly improves coverage time when applied locally, but not globally. The result is a task- and robot-agnostic CPP method that can be applied to different applications, which has been demonstrated on the exploration and lawn mowing tasks.

Furthermore, we transfer the trained CPP policies to a semi-virtual environment in order to reduce the sim-to-real gap. We address the three challenges of sim-to-real transfer presented in the introduction. (1)~The mismatch between simulated and real dynamics is reduced by a high inference frequency and fine-tuning the model in a real setting. (2)~Non-Markovian dynamics are accounted for by using action observations, and incorporating realistic accelerations and delays into the simulation. (3)~Computational delays are addressed by parallelizing the environment interactions and the model updates. In our experiments, we find that training the model in simulation assuming a first-order Markov process transfers well without fine-tuning, as long as the inference frequency is sufficiently high. Meanwhile, a higher-order policy can be further improved through fine-tuning, and can be deployed at a lower inference frequency, which lowers the computational requirements for a deployed system.

\section{Limitations and Future Work}

Limitations and future work relate to accurate pose, simulated sensor data, and static environments with no moving obstacles, which we discuss here.

\textbf{Accurate pose.} The real-world experiments are conducted with highly accurate pose estimates from an indoor positioning system, which may not be available, depending on the application. In such cases it needs to be estimated by other means, such as SLAM, which introduces higher levels of noise. However, our simulation experiments show that even under high levels of localization and sensing noise, the performance of our method is not impacted to a significant extent. Thus, we believe that our conclusions likely extend to real environments with higher noise levels.

\textbf{Sensing.} The sim-to-real gap is mainly reduced with respect to robot kinematics and dynamics, while the sensor data remains simulated. This was done to evaluate the sim-to-real transfer capability of the learning algorithm in a controlled manner. Nevertheless, we believe that our conclusions likely extend to real sensors as well, for the same reason as above.

\textbf{Dynamic obstacles.} While dynamic obstacles are not explicitly modeled in the training framework, they can be accounted for by including, for instance, an object detector, to update the map based on detected moving obstacles. When an object is detected, the corresponding location in the obstacle map is replaced with non-free space, and when the object is no longer detected, it is replaced with free space again. This was briefly tested in our experiments, with the following results. The robot did not collide with detected obstacles, but rather it planned around them. When the objects were moved away, and the robot explored the same region again, it planned its path through that region if necessary. In future work, we have to consider such experiments more systematically.



\appendices

\section{\break Agent Architecture Details}
\label{supp_sec_agent_architecture}

As proposed for soft actor-critic (SAC) \cite{haarnoja2018soft_2}, we use two Q-functions that are trained independently, together with accompanying target Q-functions whose weights are exponentially moving averages. No separate state-value network is used. All networks, including actor network, Q-functions, and target Q-functions share the same network architecture, which is either a multilayer perceptron (MLP), a naive convolutional neural network (CNN), or our proposed scale-grouped convolutional neural network (SGCNN). These are described in the following paragraphs. The MLP contains a total of $3.2$M parameters, where most are part of the initial fully connected layer, which takes the entire observation vector as input. Meanwhile, the CNN-based architectures only contain $0.8$M parameters, as the convolution layers process the maps with fewer parameters.

\textbf{MLP.} As mentioned in the main paper, the map and sensor feature extractors for MLP simply consist of flattening layers that restructure the inputs into vectors. The flattened multi-scale coverage maps, obstacle maps, frontier maps, and lidar detections are concatenated and fed to the fusion module. For the Q-functions, the predicted action is appended to the input of the fusion module. The fusion module consists of two fully connected layers with ReLU and 256 units each. A final linear prediction head is appended, which predicts the mean and standard deviation for sampling the action in the actor network, or the Q-value in the Q-functions.

\textbf{Naive CNN.} This architecture is identical to SGCNN, see below, except that it convolves all maps simultaneously in one go, without utilizing grouped convolutions.

\textbf{SGCNN.} Our proposed SGCNN architecture uses the same fusion module as MLP, but uses additional layers for the feature extractors, including convolutional layers for the image-like maps. Due to the simple nature of the distance readings for the lidar sensor, the lidar feature extractor only uses a single fully connected layer with ReLU. It has the same number of hidden neurons as there are lidar rays. The map feature extractor consists of a $2 \times 2$ convolution with stride $2$ to reduce the spatial resolution, followed by three $3 \times 3$ convolutions with stride 1 and without padding. We use $24$ total channels in the convolution layers. Note that the maps are grouped by scale, and convolved separately by their own set of filters. The map features are flattened into a vector and fed to a fully connected layer of $256$ units, which is the final output of the map feature extractor. ReLU is used as the activation function throughout.


\textbf{Computational efficiency.} As our goal is to perform fine-tuning and inference on edge hardware in real time, the computational efficiency of our model is highly important. In this regard, our SGCNN architecture is suitable, with only $0.8$M parameters. This is lightweight compared to popular ImageNet computer vision models, such as MobileNet~\cite{howard2019mobilenetv3}, ResNet~\cite{he2016resnet}, ConvNeXt~\cite{liu2022convnext}, SwinTransformer~\cite{liu2021swin, liu2022swinv2}, and VisionTransformer~\cite{dosovitskiy2021vit}. In Table \ref{supp_table_inference_time}, we compare our model against these architectures in terms of number of parameters and inference time. The inference time for the ImageNet models were averaged over $1000$ forward passes on $64 \times 64 \times 3$-dimensional input images, as this corresponds to the same size as our $32 \times 32 \times 4 \times 3$-dimensional input maps. However, the VisionTransformer expects a $224 \times 224 \times 3$-dimensional input, so this was used instead. As we can see from the table, our SGCNN architecture measures the lowest inference time with the fewest parameters.

\begin{table}[h]
    \centering
    \caption{Model comparison in terms of number of parameters and inference time between our proposed SGCNN architecture and existing ImageNet computer vision models. The inference time is measured on two laptops; L1 (i5-520M CPU, no GPU) and L2 (i7-13700H CPU, 4060 GPU).}
    \begin{tabular}{lccc}
        \toprule
        Model & \#params & Time (L1) & Time (L2) \\
        \midrule
        MobileNet-V3-Small & 2.5M & 17 ms & 8 ms \\
        MobileNet-V3-Large & 5.5M & 25 ms & 10 ms \\
        ResNet-18 & 12M & 31 ms & 5 ms \\
        ResNet-50 & 26M & 77 ms & 8 ms \\
        ConvNeXt-Tiny & 29M & 82 ms & 15 ms \\
        Swin-V2-T & 29M & 175 ms & 35 ms \\
        ViT-Base-16 & 86M & 1194 ms & 116 ms \\
        \midrule
        \textbf{SGCNN (ours)} & \textbf{0.8M} & \textbf{5 ms} & \textbf{2 ms} \\
        \bottomrule
    \end{tabular}
    \label{supp_table_inference_time}
\end{table}

\section{\break Choice of Hyperparameters}
\label{supp_sec_hyperparameters}

\textbf{Multi-scale map parameters.} We started by considering the size and resolution in the finest scale, such that sufficiently small details could be represented in the nearest vicinity. We concluded that a resolution of $0.0375$ meters per pixel was a good choice as it corresponds to $8$ pixels for a robot with a diameter of $30$ cm. Next, we chose a size of $32 \times 32$ pixels, as this results in the finest scale spanning $1.2 \times 1.2$ meters. Next, we chose the scale factor $s = 4$, as this allows only a few maps to represent a relatively large region, while maintaining sufficient detail in the second finest scale. Finally, the training needs to be done with a fixed number of scales. The most practical way is to simply train with sufficiently many scales to cover the maximum size for a particular use case. Thus, we chose $m = 4$ scales, which in total spans a $76.8 \times 76.8$~m region, and can contain any of the considered training and evaluation environments. If the model is deployed in a small area, the larger scales would not contain any frontier points in the far distance, but the agent can still cover the area by utilizing the smaller scales. For larger use cases, increasing the size of the represented area by adding more scales is fairly cheap, as the computational cost is $\mathcal{O}(\log n)$.

\textbf{Reward parameters.} We started by fixing $\lambda_\mathrm{area} = 1$ and tuned the learning rate accordingly. Due to the normalization of the area and TV rewards, they are given approximately the same weight for $\lambda_\mathrm{area} = 1$ and $\lambda_\mathrm{TV} = 1$. Thus, we used this as a starting point, which seemed to work well for the lawn mowing problem, while a lower weight for the TV rewards was advantageous for exploration. Here, we made a parameter sweep over $\lambda_\mathrm{TV}^\mathrm{I}$ and $\lambda_\mathrm{TV}^\mathrm{G}$, with values in $\{ 0.1, 0.2, 0.5, 1, 2 \}$, and found $0.2$ to be the best choice for exploration. For the collision reward, $R_\mathrm{coll}$, we found $-10$ to work best from $\{ 0, -1, -10 \}$. For the constant reward, $R_\mathrm{const}$, a value of $-0.1$ was chosen out of $\{ 0, -0.1, -1 \}$.

\textbf{Episode truncation.} For large environments, we found it important not to truncate the episodes too early, as this would hinder learning. If the truncation parameter $\tau$ was set too low, the agent would not be forced to learn to cover an area completely, as the episode would simply truncate whenever the agent could not progress, without a large penalty. With the chosen value of $\tau = 1000$, the agent would be greatly penalized by the constant reward $R_\mathrm{const}$ for not reaching the goal coverage, and would be forced to learn to cover the complete area in order to maximize the reward.

\section{\break Progressive Training}
\label{supp_sec_progressive_training}

To improve the convergence in the early parts of training, we apply curriculum learning \cite{bengio2009curriculum}, which has shown to be effective in reinforcement learning \cite{narvekar2020curriculum}. We use simple maps and increase their difficulty progressively. To this end, we rank the fixed training maps by difficulty, and assign them into tiers, see Fig. \ref{supp_fig_train_maps_sim}. The maps in the lower tiers have smaller sizes and simpler obstacles. For the higher tiers, the map size is increased together with the complexity of the obstacles.

\begin{table*}[t]
    \setlength{\tabcolsep}{5pt}
    \centering
    \caption{\textbf{First-order policy.} Time in minutes for reaching $90\%$ and $99\%$ coverage. \textit{Env}: Evaluation environment. \textit{Steps}: Fine-tuning steps. Inference step size is in ms. Note: Training step size is always 500 ms.}
    \begin{tabular}{lllcccccccccccccc}
        \toprule
        & & & \multicolumn{2}{c}{Map 1} & \multicolumn{2}{c}{Map 2} & \multicolumn{2}{c}{Map 3} & \multicolumn{2}{c}{Map 4} & \multicolumn{2}{c}{Map 5} & \multicolumn{2}{c}{Map 6} & \multicolumn{2}{c}{Average} \\
        \cmidrule(l{\tabcolsep}r{\tabcolsep}){4-5}
        \cmidrule(l{\tabcolsep}r{\tabcolsep}){6-7}
        \cmidrule(l{\tabcolsep}r{\tabcolsep}){8-9}
        \cmidrule(l{\tabcolsep}r{\tabcolsep}){10-11}
        \cmidrule(l{\tabcolsep}r{\tabcolsep}){12-13}
        \cmidrule(l{\tabcolsep}r{\tabcolsep}){14-15}
        \cmidrule(l{\tabcolsep}r{\tabcolsep}){16-17}
        Env & Steps & $\Delta t$ & $T_{90}$ & $T_{99}$ & $T_{90}$ & $T_{99}$ & $T_{90}$ & $T_{99}$ & $T_{90}$ & $T_{99}$ & $T_{90}$ & $T_{99}$ & $T_{90}$ & $T_{99}$ & $T_{90}$ & $T_{99}$ \\
        \midrule
        Sim & 0 & 500 & 6.1 & 8.0 & 5.2 & 7.8 & 5.7 & 9.1 & 5.9 & 9.6 & 5.2 & 8.1 & 8.1 & 11.8 & 6.0 & 9.1 \\
        Real & 0 & 500 & 5.7 & 9.5 & 6.5 & 10.5 & 6.5 & 9.7 & 9.0 & 10.9 & 6.0 & 9.1 & 8.5 & 13.8 & 7.0 & 10.6 \\
        \midrule
        Real & 0 & 15 & 5.3 & 6.9 & 5.1 & 7.6 & 5.5 & 10.5 & 5.0 & 6.8 & 5.2 & 6.8 & 6.5 & 11.0 & 5.4 & 8.3 \\
        Real & 80k & 15 & 6.4 & 8.4 & 6.1 & 11.0 & 5.6 & 11.6 & 5.7 & 9.0 & 6.2 & 10.4 & 7.5 & 15.1 & 6.3 & 10.9 \\
        \bottomrule
    \end{tabular}
    \label{table_mowing_1_ext}
\end{table*}

\begin{table*}[h]
    \setlength{\tabcolsep}{5pt}
    \centering
    \caption{\textbf{Higher-order policy.} Time in minutes for reaching $90\%$ and $99\%$ coverage. \textit{Env}: Evaluation environment. \textit{Steps}: Fine-tuning steps. Inference step size is in ms. Note: Training step size is always 500 ms.}
    \begin{tabular}{lllcccccccccccccc}
        \toprule
        & & & \multicolumn{2}{c}{Map 1} & \multicolumn{2}{c}{Map 2} & \multicolumn{2}{c}{Map 3} & \multicolumn{2}{c}{Map 4} & \multicolumn{2}{c}{Map 5} & \multicolumn{2}{c}{Map 6} & \multicolumn{2}{c}{Average} \\
        \cmidrule(l{\tabcolsep}r{\tabcolsep}){4-5}
        \cmidrule(l{\tabcolsep}r{\tabcolsep}){6-7}
        \cmidrule(l{\tabcolsep}r{\tabcolsep}){8-9}
        \cmidrule(l{\tabcolsep}r{\tabcolsep}){10-11}
        \cmidrule(l{\tabcolsep}r{\tabcolsep}){12-13}
        \cmidrule(l{\tabcolsep}r{\tabcolsep}){14-15}
        \cmidrule(l{\tabcolsep}r{\tabcolsep}){16-17}
        Env & Steps & $\Delta t$ & $T_{90}$ & $T_{99}$ & $T_{90}$ & $T_{99}$ & $T_{90}$ & $T_{99}$ & $T_{90}$ & $T_{99}$ & $T_{90}$ & $T_{99}$ & $T_{90}$ & $T_{99}$ & $T_{90}$ & $T_{99}$ \\
        \midrule
        Sim & 0 & 500 & 5.6 & 8.6 & 5.7 & 8.7 & 5.2 & 8.2 & 5.3 & 7.6 & 5.2 & 7.3 & 7.7 & 10.3 & 5.8 & 8.5 \\
        Real & 0 & 500 & 6.8 & 10.5 & 7.9 & 10.7 & 6.7 & 10.9 & 7.0 & 9.1 & 6.0 & 8.3 & 9.4 & 12.5 & 7.3 & 10.3 \\
        Real & 60k & 500 & 6.7 & 9.2 & 6.5 & 9.5 & 6.1 & 11.0 & 6.2 & 9.1 & 6.2 & 9.2 & 8.4 & 13.0 & 6.7 & 10.2 \\
        \bottomrule
    \end{tabular}
    \label{table_mowing_2_ext}
\end{table*}

\begin{table}[h]
    \setlength{\tabcolsep}{5pt}
    \centering
    \caption{The progressive training levels contain maps of increasingly higher tiers and goal coverage percentages. At the highest levels, generated maps with random floor plans and obstacles are also used.}
    \begin{tabular}{ccccccc}
        \toprule
        & \multicolumn{3}{c}{Exploration} & \multicolumn{3}{c}{Lawn mowing} \\
        \cmidrule(l{\tabcolsep}r{\tabcolsep}){2-4}
        \cmidrule(l{\tabcolsep}r{\tabcolsep}){5-7}
        & Map & Random & Goal & Map & Random & Goal \\
        Level & tiers & maps & coverage & tiers & maps & coverage \\
        \midrule
        1 & 1,2   & & $90\%$ & 0    & & $90\%$ \\
        2 & 1,2,4 & & $90\%$ & 0,1  & & $90\%$ \\
        3 & 1,2,4 & & $95\%$ & 0,1  & & $95\%$ \\
        4 & 1,2,4 & & $97\%$ & 0--2 & & $95\%$ \\
        5 & 1,2,4 & & $99\%$ & 0--2 & & $97\%$ \\
        6 & 1--4  & & $99\%$ & 0--2 & & $99\%$ \\
        7 & 1--4  & \checkmark & $99\%$ & 0--3 & & $99\%$ \\
        8 & 1--5  & \checkmark & $99\%$ & 0--3 & \checkmark & $99\%$ \\
        \bottomrule
    \end{tabular}
    \label{supp_table_levels}
\end{table}

Furthermore, we define levels containing certain sets of maps and specific goal coverage percentages, see Table \ref{supp_table_levels}. The agent starts at level 1, and progresses through the levels during training. To progress to the next level, the agent has to complete each map in the current level, by reaching the specified goal coverage. Note that randomly generated maps are also used in the higher levels, in which case the agent has to additionally complete a map with random floor plans, and a map with randomly placed circular obstacles to progress. Whenever random maps are used in a level, either a fixed map or a random map is chosen with a $50 \%$ probability each at the start of every episode.

\section{\break Random Map Generation}
\label{supp_sec_random_map_generation}

Inspired by procedural environment generation for reinforcement learning \cite{justesen2018illuminating}, we use randomly generated maps during training to increase the variation in the training data and to improve generalization. We consider random floor plans and randomly placed circular obstacles. First, the side length of a square area is chosen uniformly at random, from the interval $[2.4, 7.5]$ meters for mowing and $[9.6, 15]$ meters for exploration. Subsequently, a random floor plan is created with a $70\%$ probability. Finally, with $70\%$ probability, a number of obstacles are placed such that they are far enough apart to guarantee that the agent can visit the entire free space, i.e.\ that they are not blocking the path between different parts of the free space. An empty map can be generated if neither a floor plan is created nor obstacles placed. In the following paragraphs, we describe the floor plan generation and obstacle placement in more detail.

\textbf{Random floor plans.} The random floor plans contain square rooms of equal size in a grid-like configuration, where neighboring rooms can be accessed through door openings. On occasion, a wall is removed completely or some openings are closed off to increase the variation further. First, floor plan parameters are chosen uniformly at random, such as the side length of the rooms from $[1.5, 4.8]$ meters, wall thickness from $[0.075, 0.3]$ meters, and door opening size from $[0.6, 1.2]$ meters. Subsequently, each vertical and horizontal wall spanning the whole map either top-to-bottom or left-to-right is placed with a $90\%$ probability. After that, door openings are created at random positions between each neighboring room. Finally, one opening is closed off at random for either each top-to-bottom spanning vertical wall or each left-to-right spanning horizontal wall, not both. This is to ensure that each part of the free space can be reached by the agent.

\textbf{Random circular obstacles.} Circular obstacles with radius $0.25$ meters are randomly scattered across the map, where one obstacle is placed every four square meters on average. If the closest distance between a new obstacle and any previous obstacle or wall is less than $0.6$ meters, the new obstacle is removed to ensure that large enough gaps are left for the agent to navigate through and that it can reach every part of the free space.

\section{\break Extended Fine-tuning Coverage Results}
\label{supp_sec_extended_coverage_real}

Tables \ref{table_mowing_1_ext} and \ref{table_mowing_2_ext} are extensions of Tables \ref{table_mowing_1} and \ref{table_mowing_2} in the main paper. They contain the $90\%$ and $99\%$ coverage times for each individual evaluation map, as well as the average times.

\begin{figure*}[!t]
    \setlength{\tabcolsep}{0.5pt}
    \setlength{\fboxsep}{0pt}%
    \setlength{\fboxrule}{0.5pt}%
    \begin{tabular}{p{1cm}cccccccc}
        Tier 0 &
        \fbox{\includegraphics[width=.0745\linewidth]{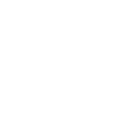}}
    \end{tabular} \\
    \begin{tabular}{p{1cm}cccccccc}
        Tier 1 &
        \fbox{\includegraphics[width=.0745\linewidth]{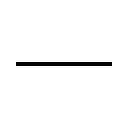}} &
        \fbox{\includegraphics[width=.0745\linewidth]{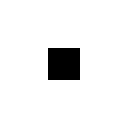}} &
        \fbox{\includegraphics[width=.0745\linewidth]{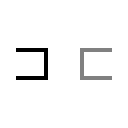}} &
        \fbox{\includegraphics[width=.0745\linewidth]{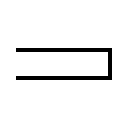}}
    \end{tabular} \\
    \begin{tabular}{p{1cm}cccccccc}
        Tier 2 &
        \fbox{\includegraphics[width=.09\linewidth]{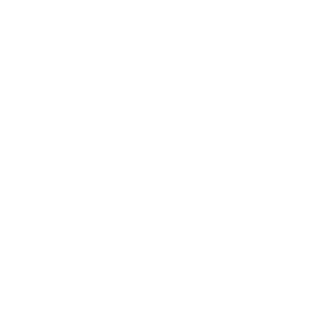}} &
        \fbox{\includegraphics[width=.09\linewidth]{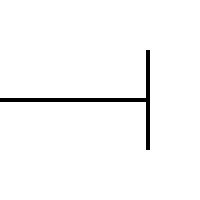}} &
        \fbox{\includegraphics[width=.09\linewidth]{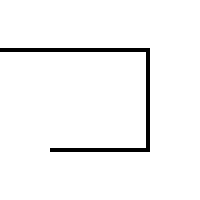}} &
        \fbox{\includegraphics[width=.09\linewidth]{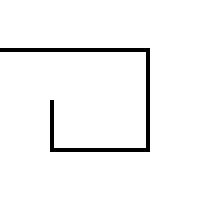}} &
        \fbox{\includegraphics[width=.09\linewidth]{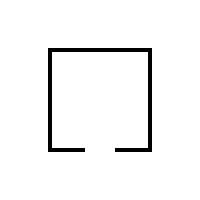}} &
        \fbox{\includegraphics[width=.09\linewidth]{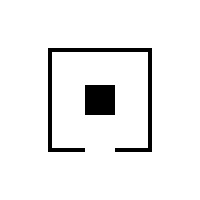}} &
        \fbox{\includegraphics[width=.09\linewidth]{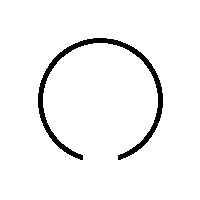}} &
        \fbox{\includegraphics[width=.09\linewidth]{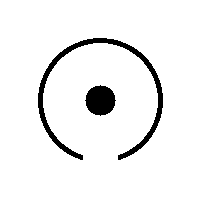}}
    \end{tabular} \\
    \begin{tabular}{p{1cm}cccccccc}
        Tier 3 &
        \fbox{\includegraphics[width=.1\linewidth]{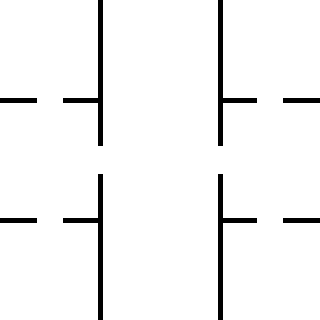}} &
        \fbox{\includegraphics[width=.1\linewidth]{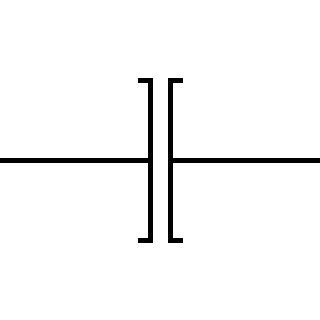}} &
        \fbox{\includegraphics[width=.1\linewidth]{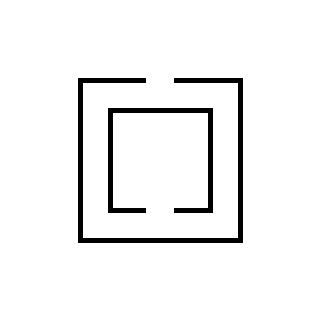}} &
        \fbox{\includegraphics[width=.1\linewidth]{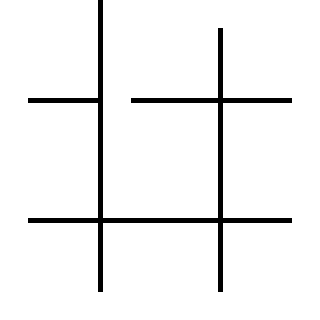}} &
        \fbox{\includegraphics[width=.1\linewidth]{figures/maps/train_3_5.png}} &
        \fbox{\includegraphics[width=.1\linewidth]{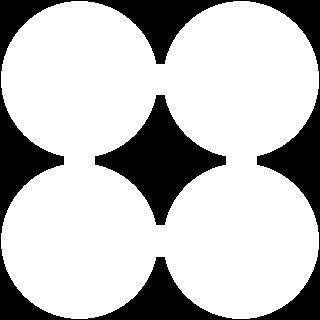}}
    \end{tabular} \\
    \begin{tabular}{p{1cm}cccccccc}
        Tier 4 &
        \fbox{\includegraphics[width=.12\linewidth]{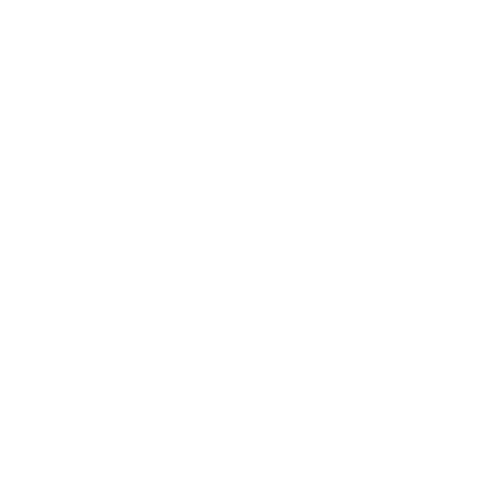}} &
        \fbox{\includegraphics[width=.12\linewidth]{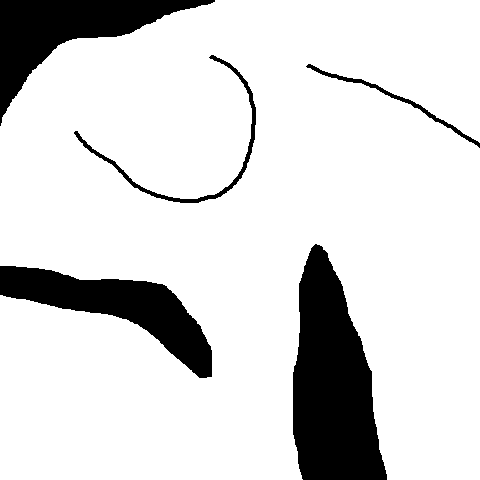}} &
        \fbox{\includegraphics[width=.12\linewidth]{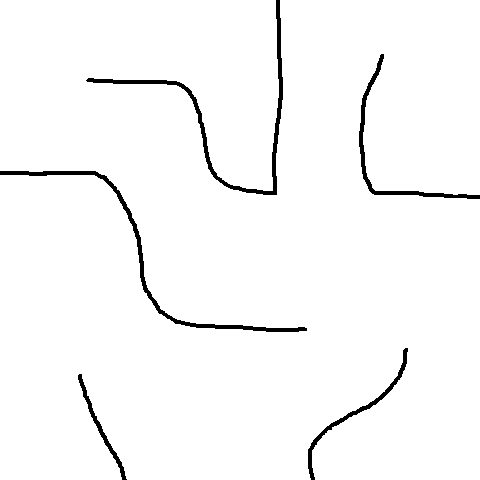}} &
        \fbox{\includegraphics[width=.12\linewidth]{figures/maps/train_4_3.png}} &
        \fbox{\includegraphics[width=.12\linewidth]{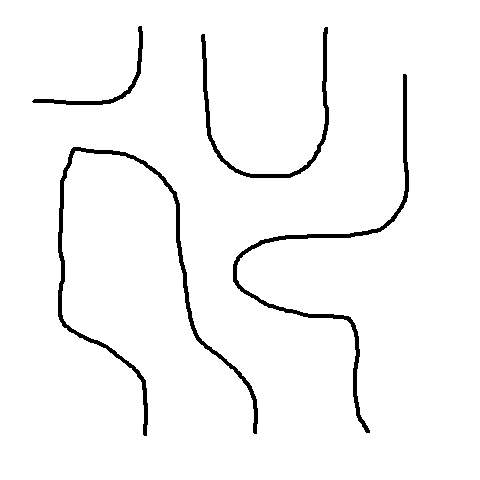}} &
        \fbox{\includegraphics[width=.12\linewidth]{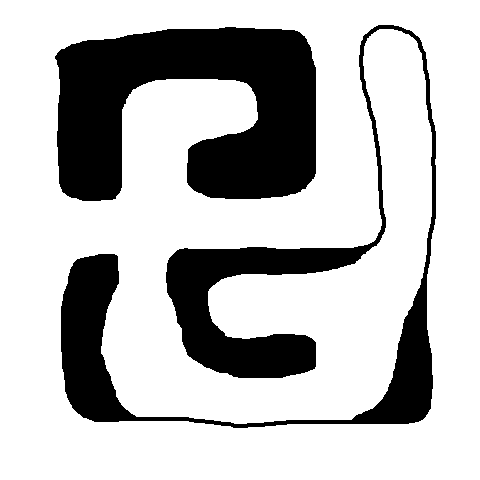}} &
        \fbox{\includegraphics[width=.12\linewidth]{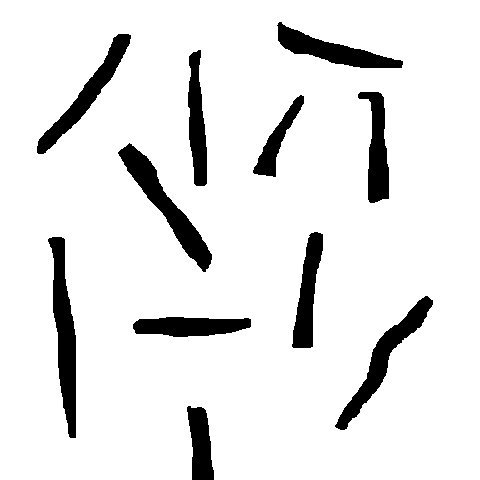}}
    \end{tabular} \\[12pt]
    \begin{tabular}{p{1cm}cccccccc}
        Tier 5 &
        \fbox{\includegraphics[width=.14\linewidth]{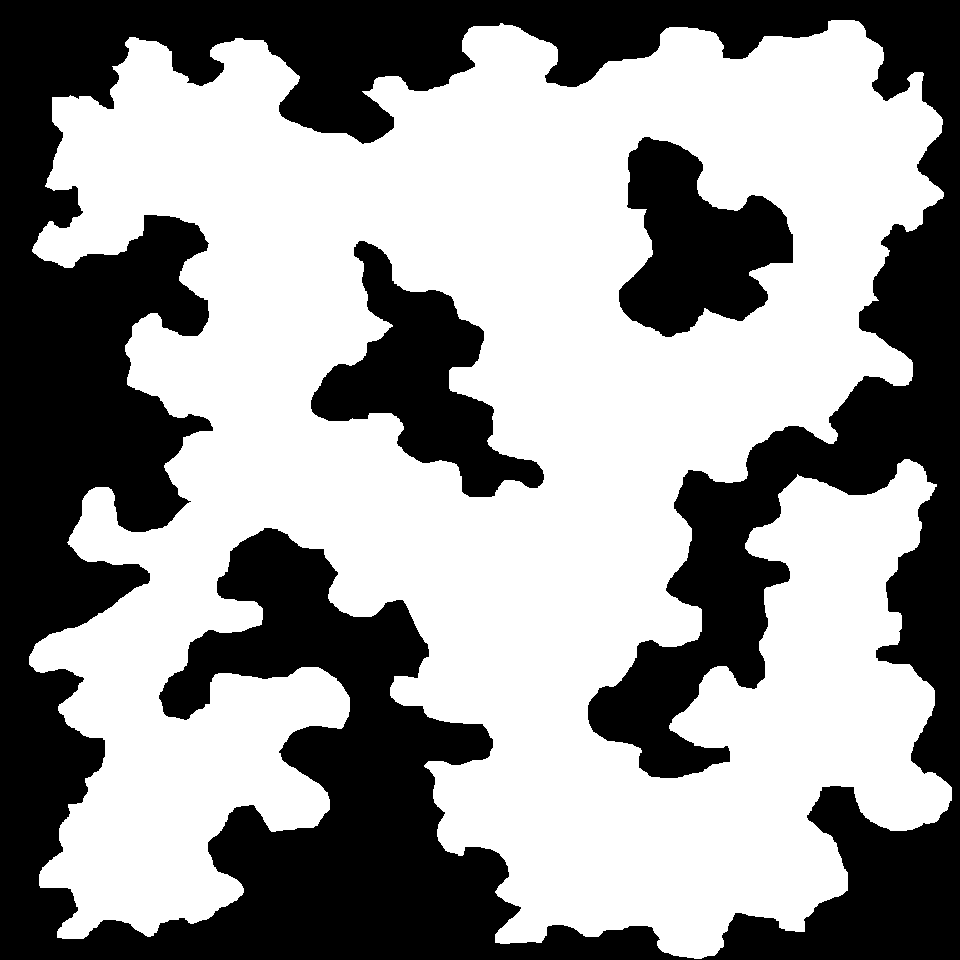}} &
        \fbox{\includegraphics[width=.14\linewidth]{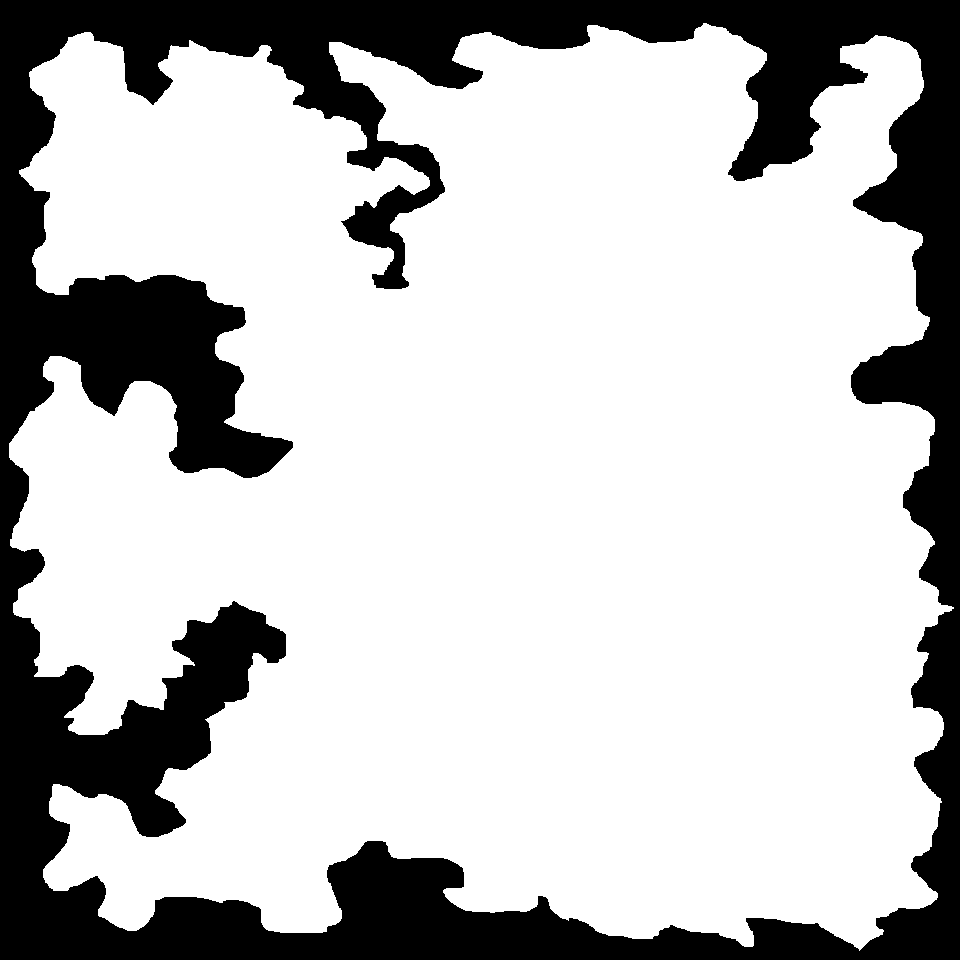}} &
        \fbox{\includegraphics[width=.14\linewidth]{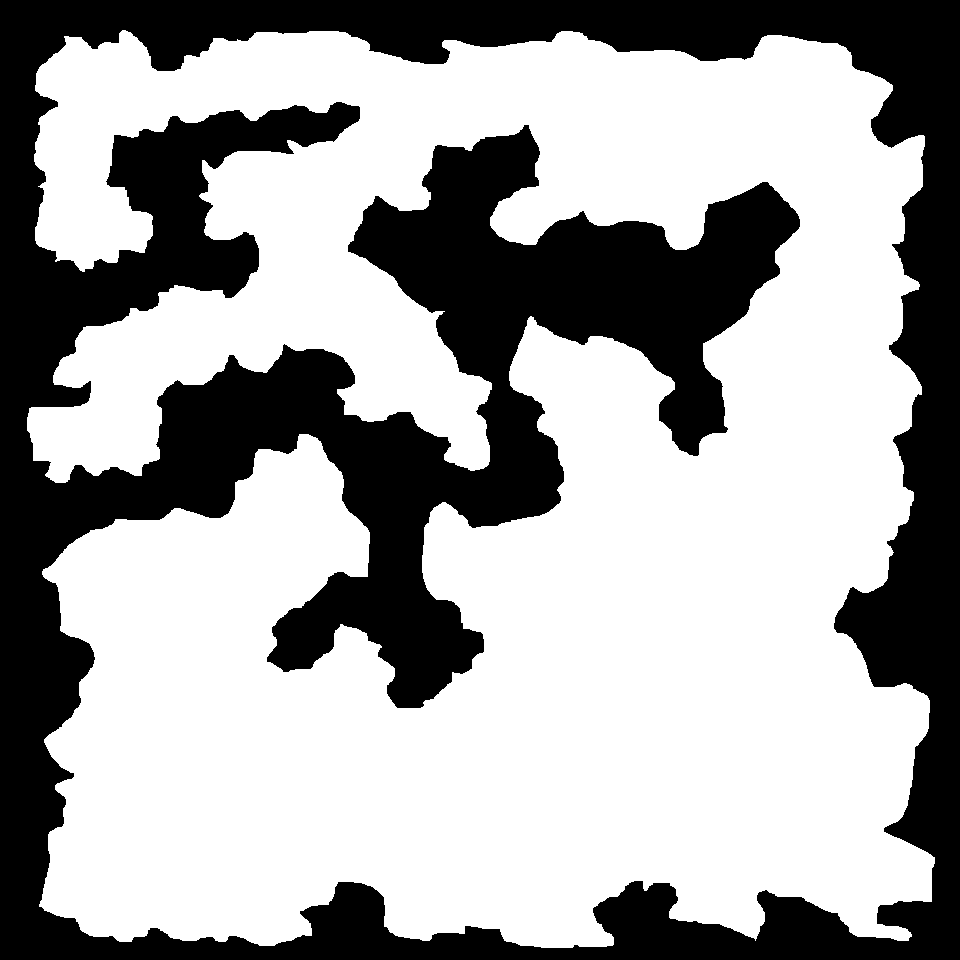}} &
        \fbox{\includegraphics[width=.14\linewidth]{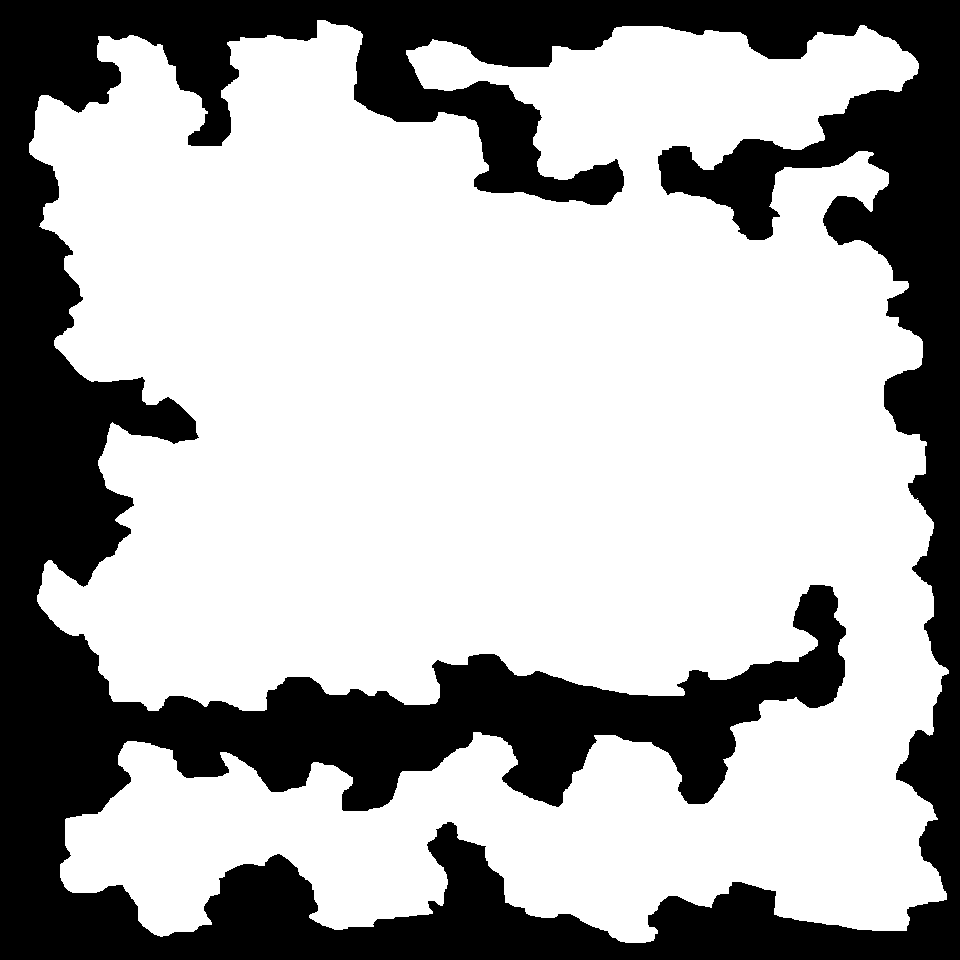}} &
        \fbox{\includegraphics[width=.14\linewidth]{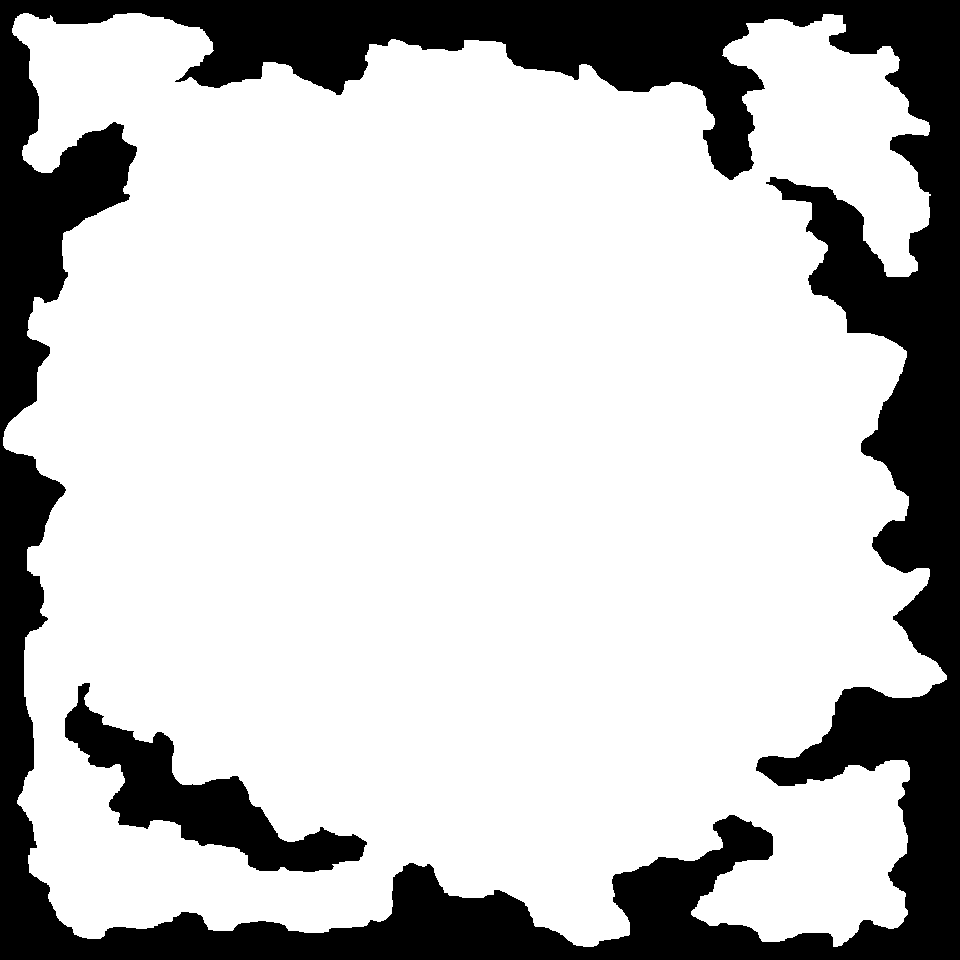}} &
        \fbox{\includegraphics[width=.14\linewidth]{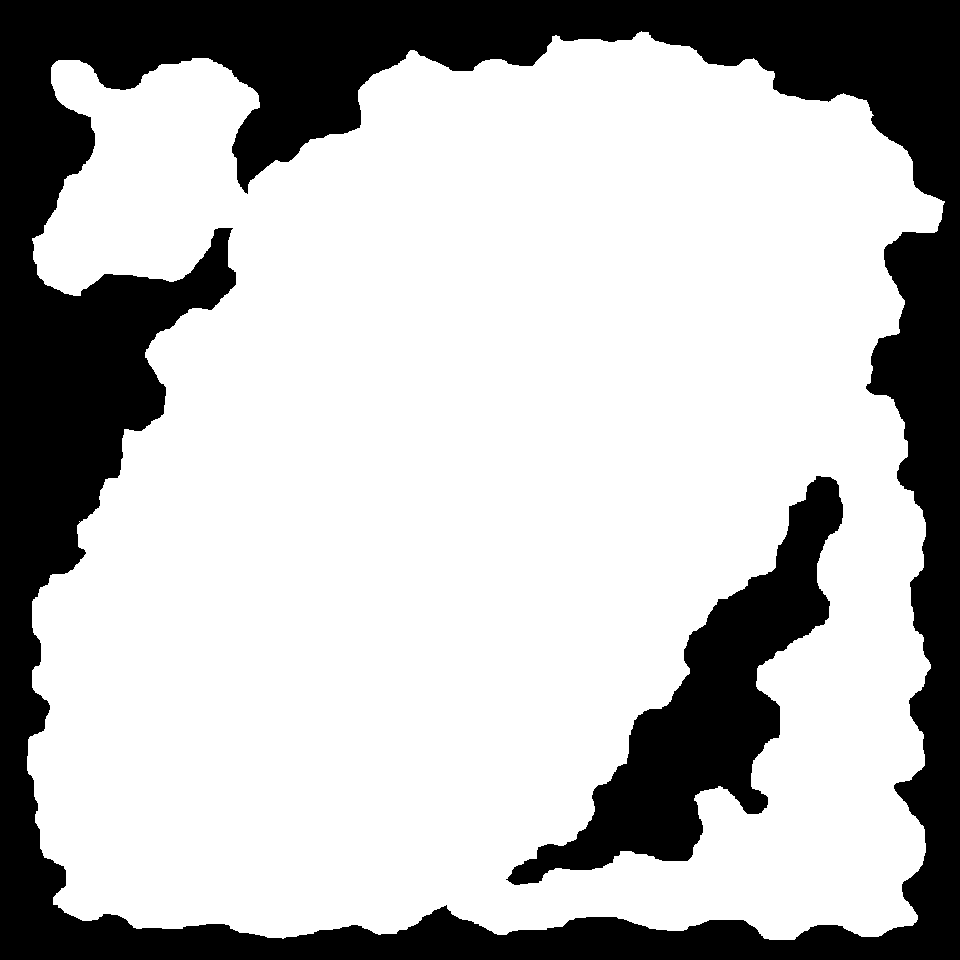}} \\
        &
        \fbox{\includegraphics[width=.14\linewidth]{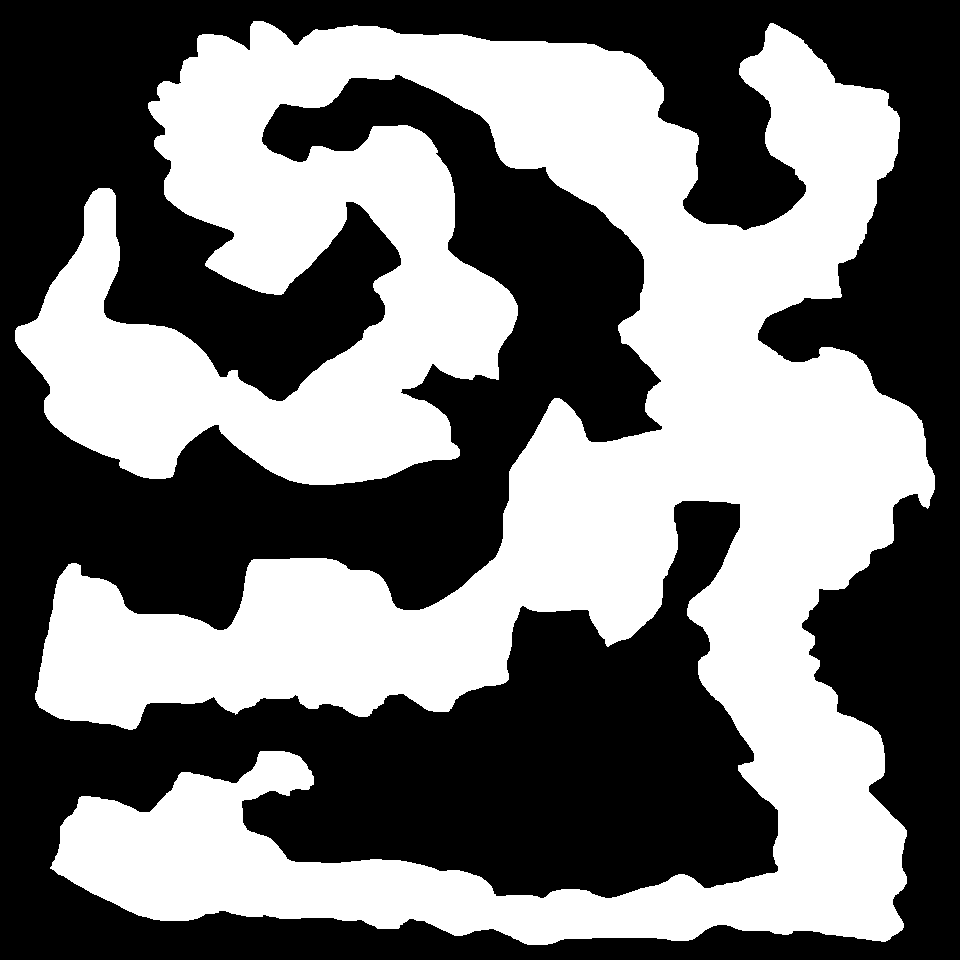}} &
        \fbox{\includegraphics[width=.14\linewidth]{figures/maps/train_5_8.png}} &
        \fbox{\includegraphics[width=.14\linewidth]{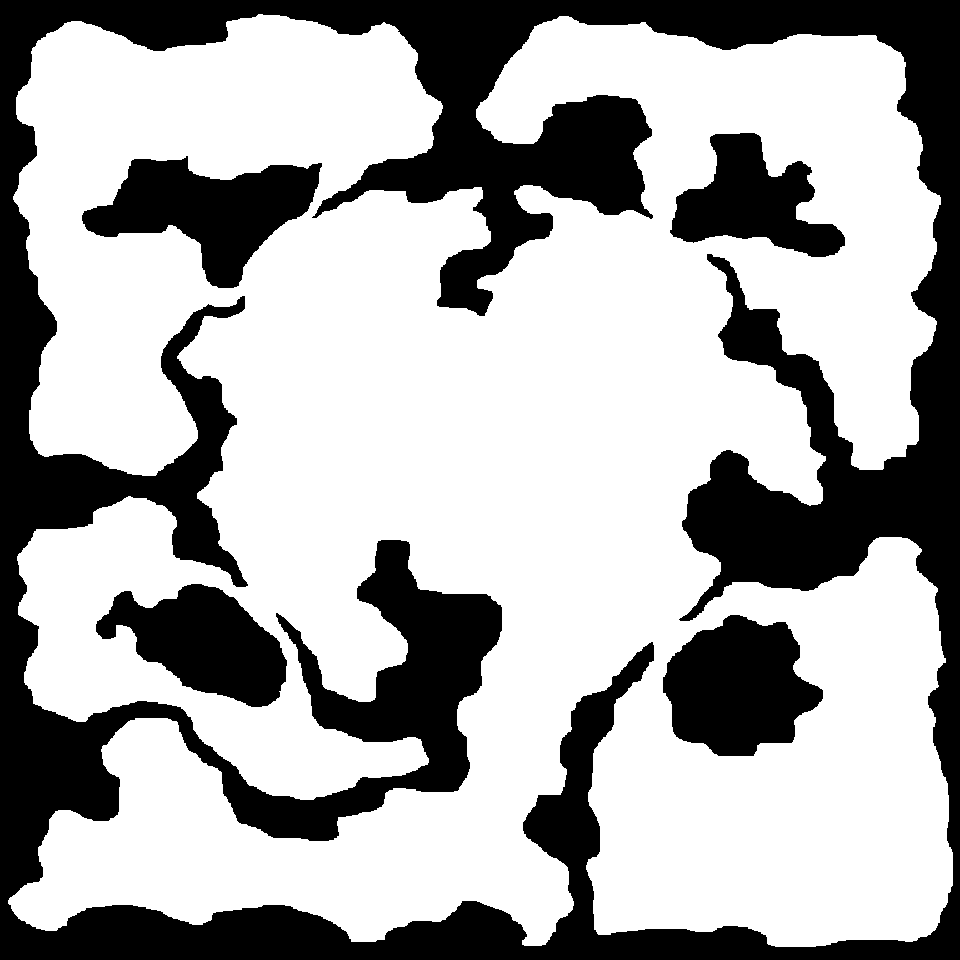}} &
        \fbox{\includegraphics[width=.14\linewidth]{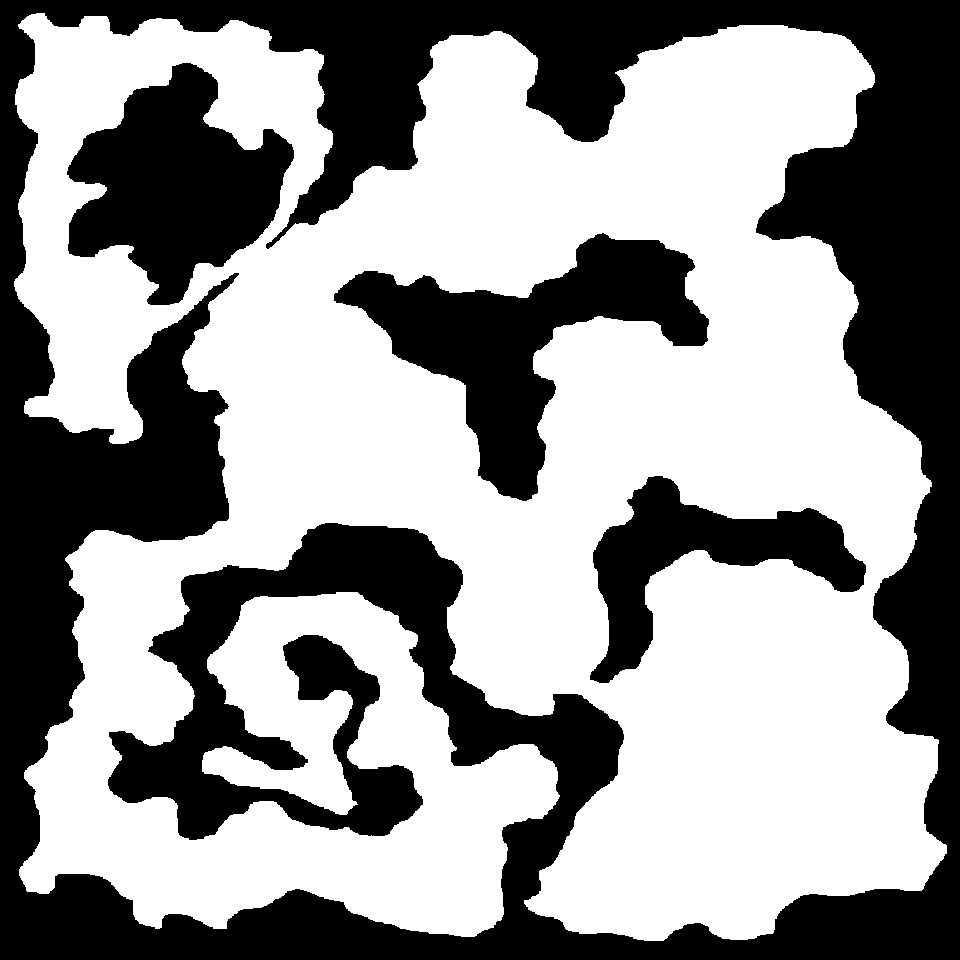}} &
        \fbox{\includegraphics[width=.14\linewidth]{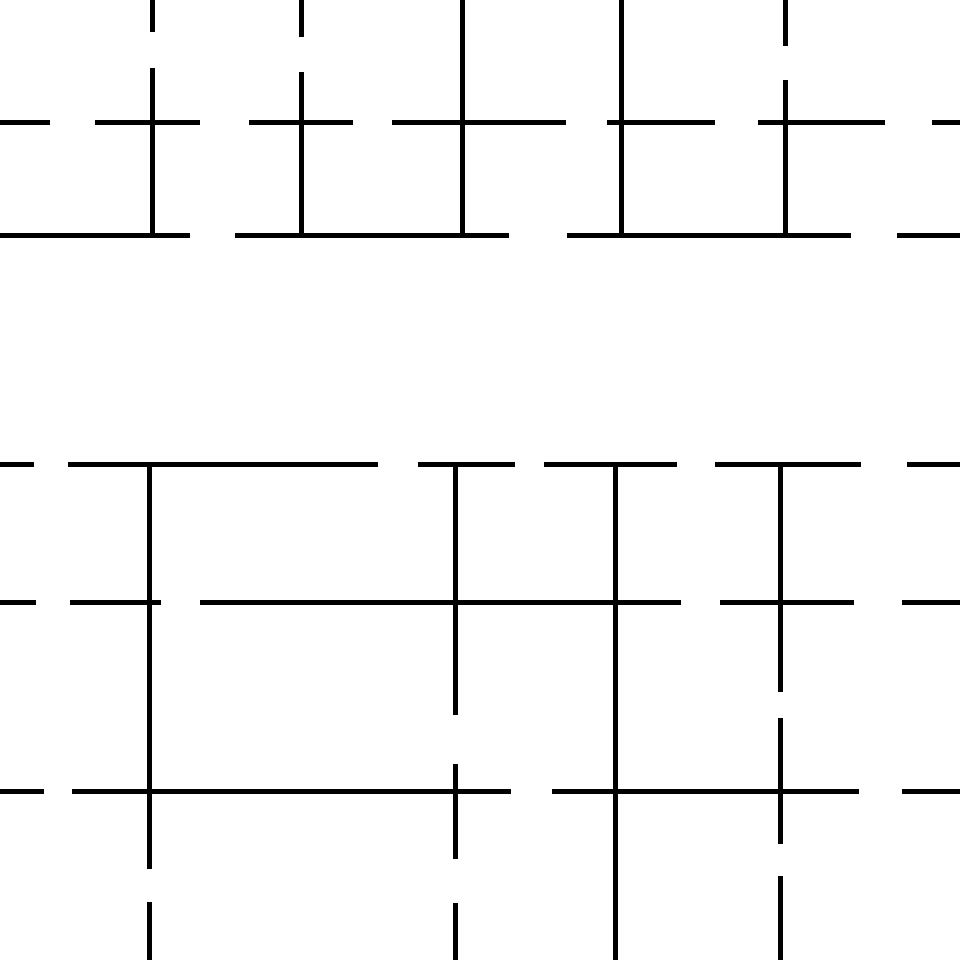}} &
        \fbox{\includegraphics[width=.14\linewidth]{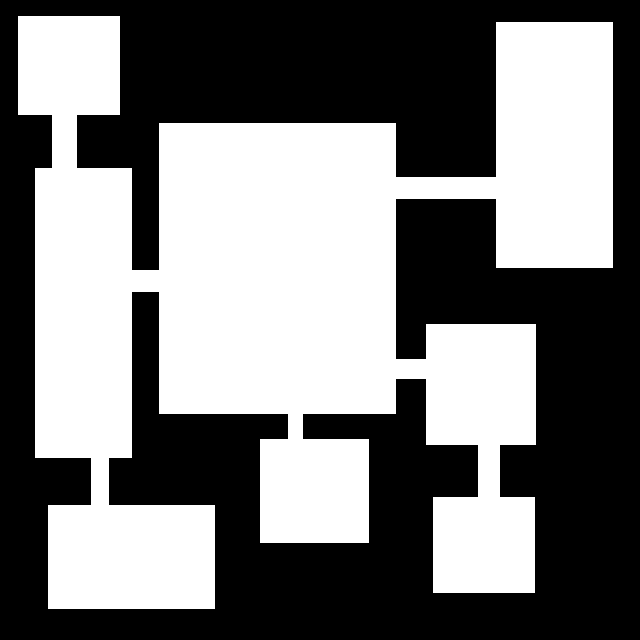}} \\
        &
        \fbox{\includegraphics[width=.14\linewidth]{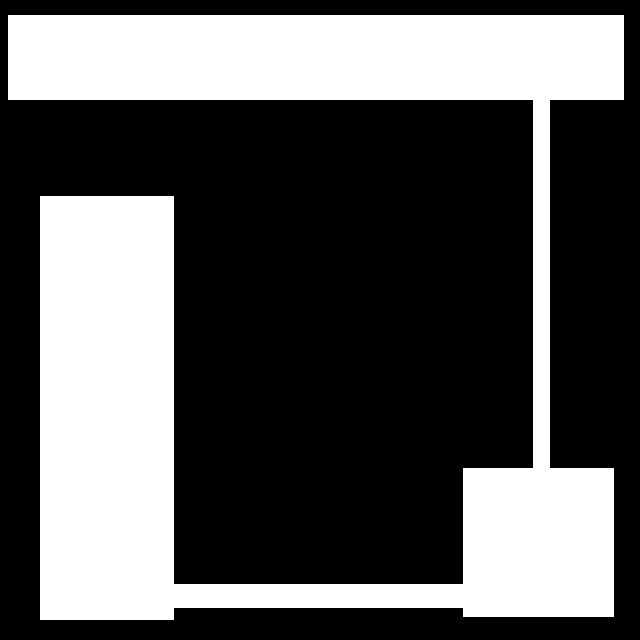}} &
        \fbox{\includegraphics[width=.14\linewidth]{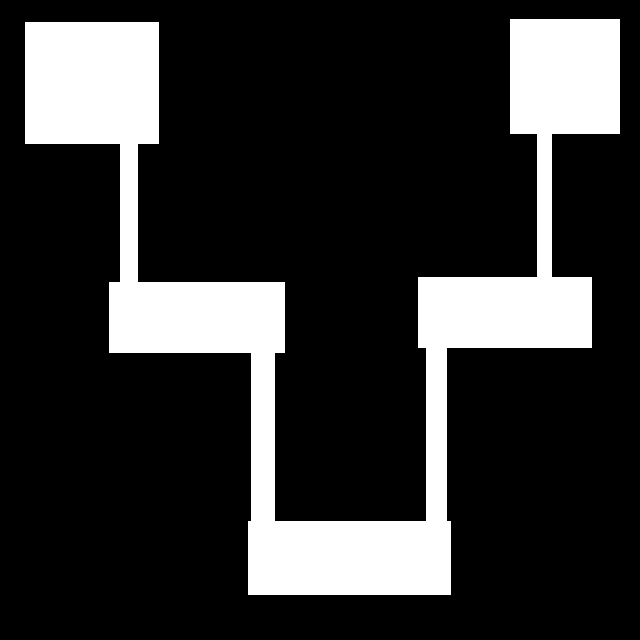}} &
        \fbox{\includegraphics[width=.14\linewidth]{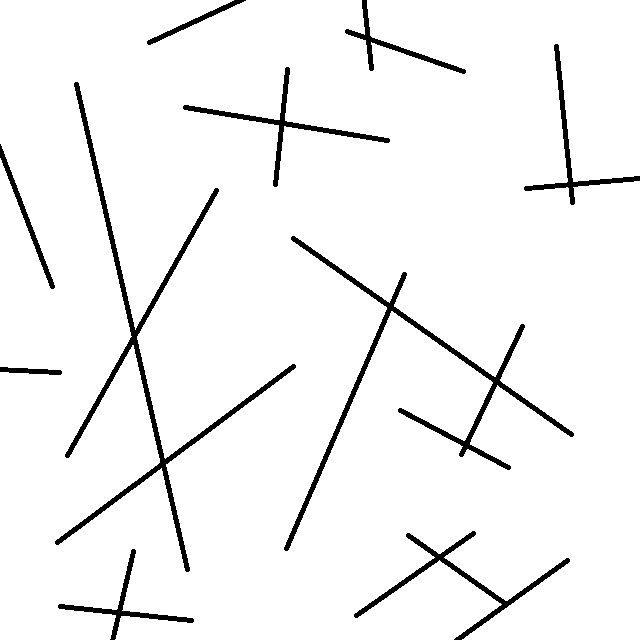}} &
        \fbox{\includegraphics[width=.14\linewidth]{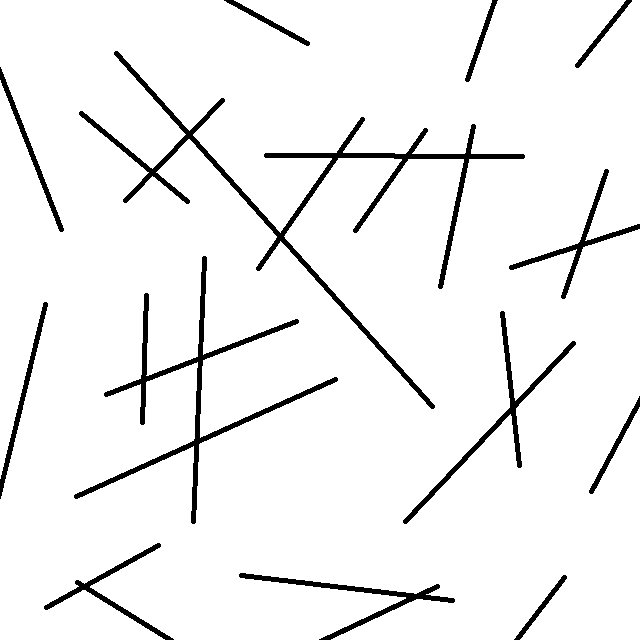}} &
        \fbox{\includegraphics[width=.14\linewidth]{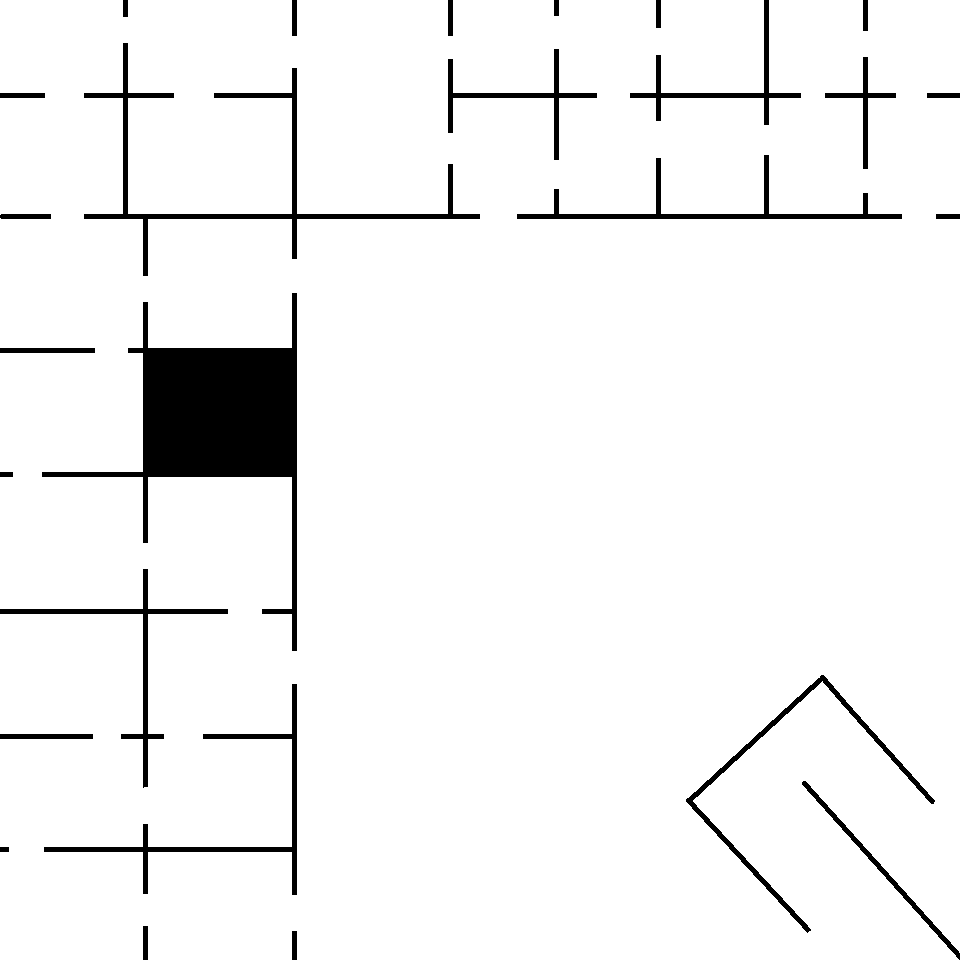}} &
        \fbox{\includegraphics[width=.14\linewidth]{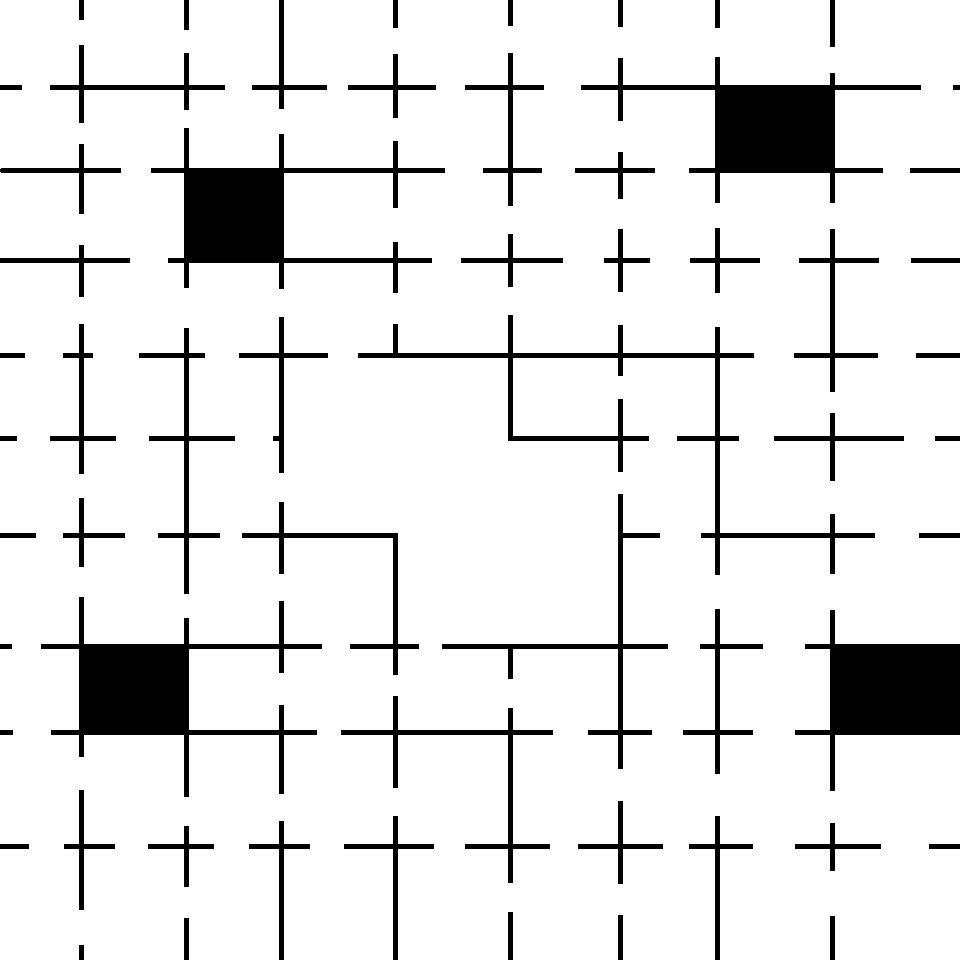}}
    \end{tabular}
    \caption{The fixed training maps used in simulation are grouped into tiers by difficulty, depending on their size and complexity of the obstacles.}
    \label{supp_fig_train_maps_sim}
\end{figure*}

\begin{figure*}[!t]
    \centering
    \setlength{\tabcolsep}{0.5pt}
    \setlength{\fboxsep}{0pt}%
    \setlength{\fboxrule}{0.5pt}%
    \begin{tabular}{ccccccccccccc}
        \fbox{\includegraphics[width=.1\linewidth]{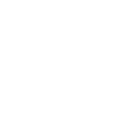}} &
        \fbox{\includegraphics[width=.1\linewidth]{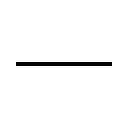}} &
        \fbox{\includegraphics[width=.1\linewidth]{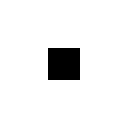}} &
        \fbox{\includegraphics[width=.1\linewidth]{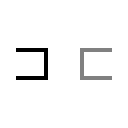}} &
        \fbox{\includegraphics[width=.1\linewidth]{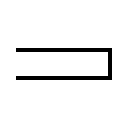}} &
        \fbox{\includegraphics[width=.1\linewidth]{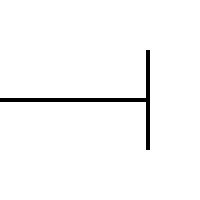}} \\
        \fbox{\includegraphics[width=.1\linewidth]{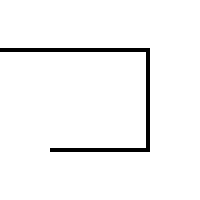}} &
        \fbox{\includegraphics[width=.1\linewidth]{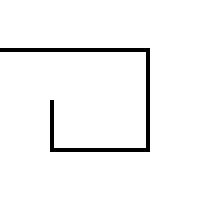}} &
        \fbox{\includegraphics[width=.1\linewidth]{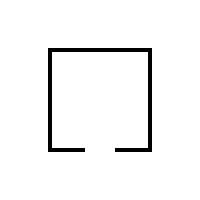}} &
        \fbox{\includegraphics[width=.1\linewidth]{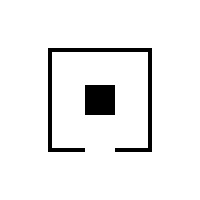}} &
        \fbox{\includegraphics[width=.1\linewidth]{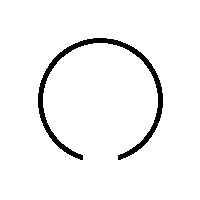}} &
        \fbox{\includegraphics[width=.1\linewidth]{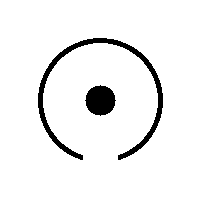}}
    \end{tabular} \\
    \caption{Training maps used during real-world fine-tuning.}
    \label{supp_fig_train_maps_real}
\end{figure*}

\begin{figure*}[!t]
    \setlength{\tabcolsep}{0.5pt}
    \setlength{\fboxsep}{0pt}%
    \setlength{\fboxrule}{0.5pt}%
    \begin{tabular}{p{1cm}cccccccc}
        (a) &
        \fbox{\includegraphics[width=.14\linewidth]{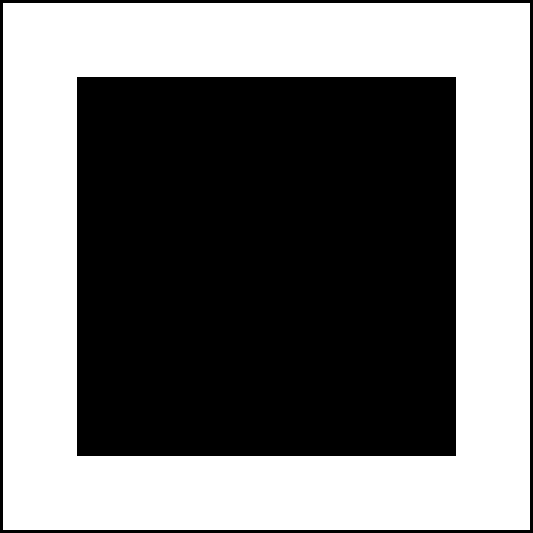}} &
        \fbox{\includegraphics[width=.14\linewidth]{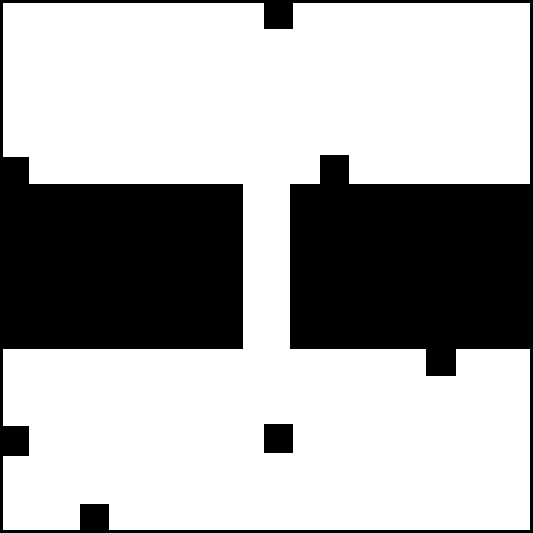}} &
        \fbox{\includegraphics[width=.14\linewidth]{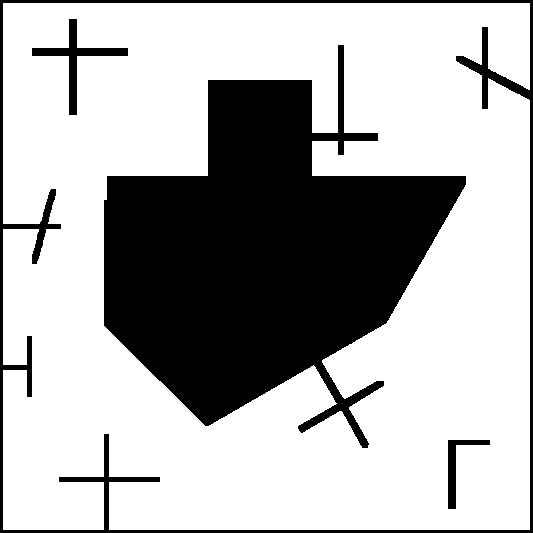}} &
        \fbox{\includegraphics[width=.14\linewidth]{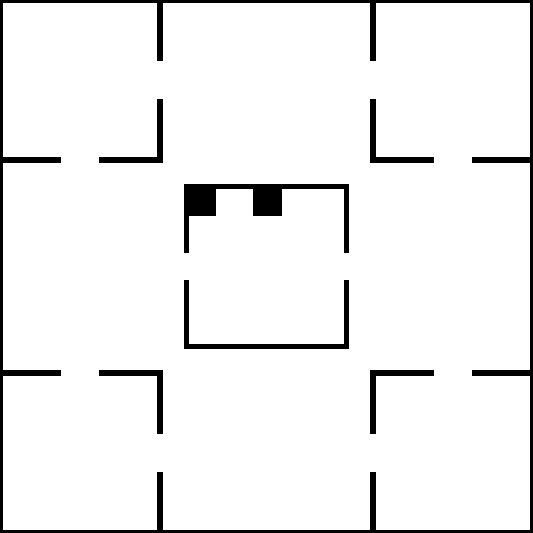}} &
        \fbox{\includegraphics[width=.14\linewidth]{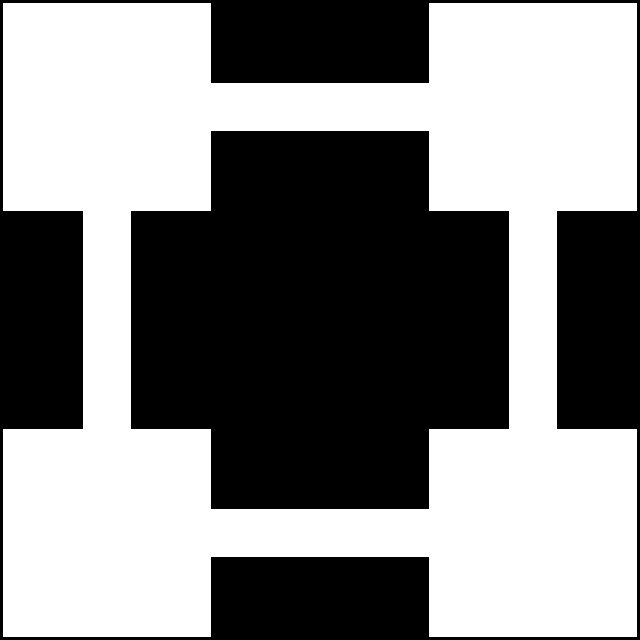}} &
        \fbox{\includegraphics[width=.14\linewidth]{figures/maps/eval_exploration_21.png}}
    \end{tabular} \\
    \begin{tabular}{p{1cm}cccccccc}
        (b) &
        \fbox{\includegraphics[width=.14\linewidth]{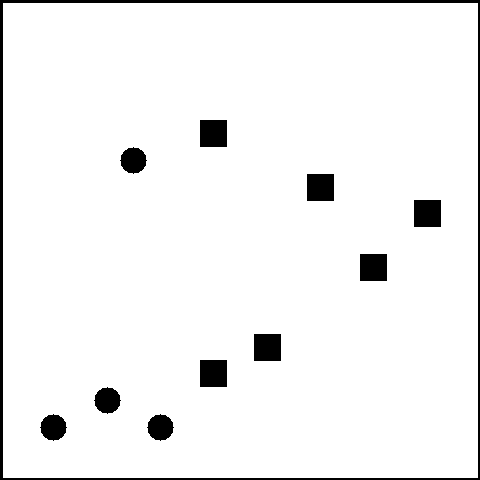}} &
        \fbox{\includegraphics[width=.14\linewidth]{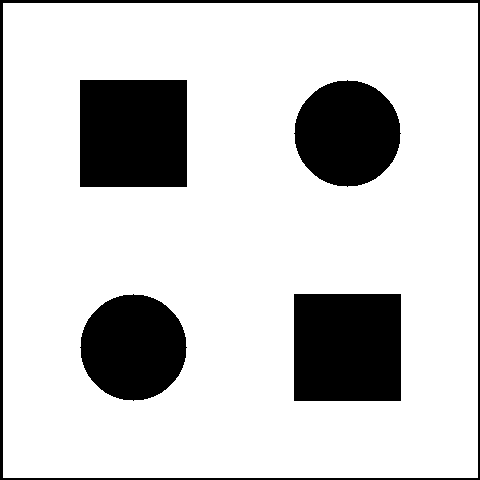}} &
        \fbox{\includegraphics[width=.14\linewidth]{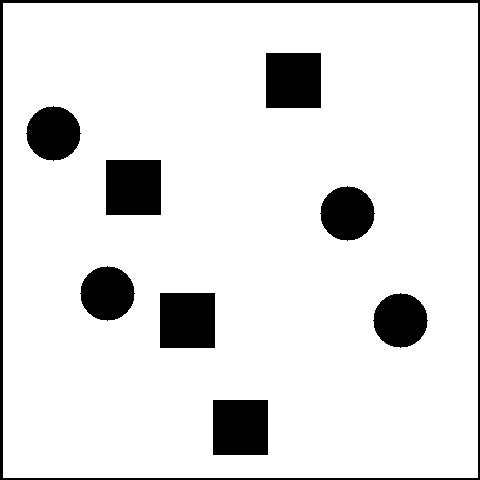}} &
        \fbox{\includegraphics[width=.14\linewidth]{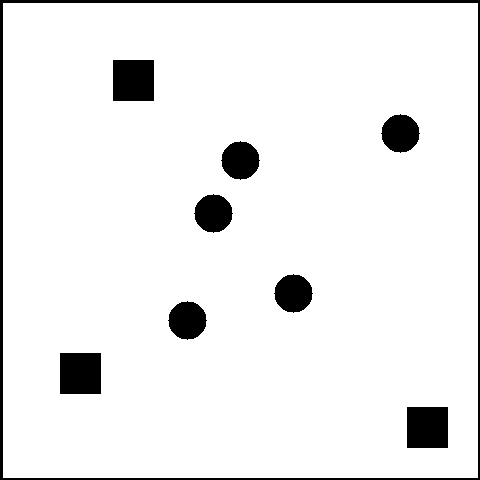}} &
        \fbox{\includegraphics[width=.14\linewidth]{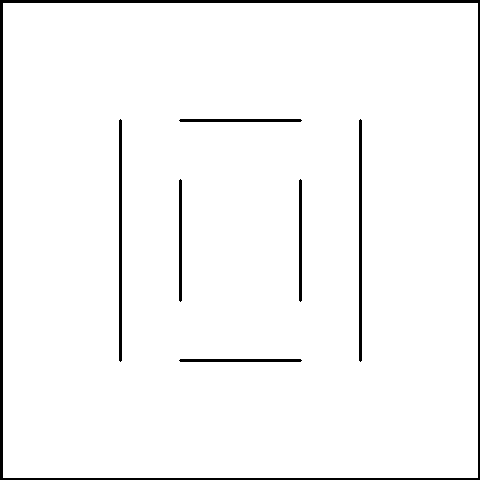}} &
        \fbox{\includegraphics[width=.14\linewidth]{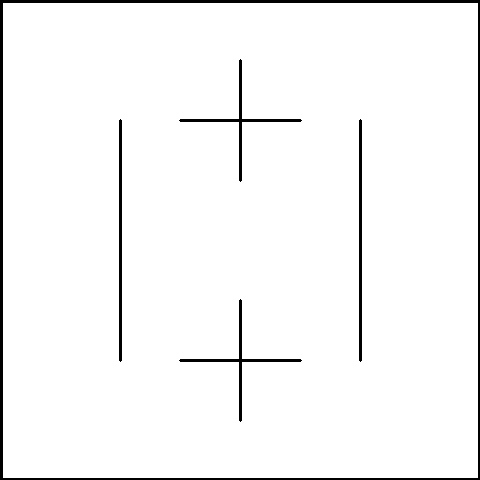}}
    \end{tabular} \\
    \begin{tabular}{p{1cm}cccccccc}
        (c) &
        \fbox{\includegraphics[width=.14\linewidth]{figures/maps/eval_mowing_9.png}} &
        \fbox{\includegraphics[width=.14\linewidth]{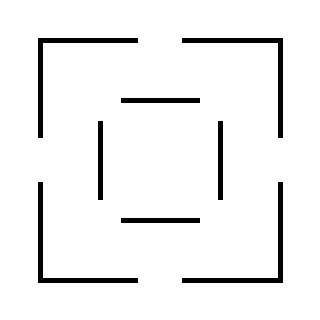}} &
        \fbox{\includegraphics[width=.14\linewidth]{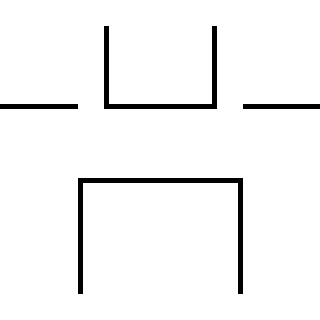}} &
        \fbox{\includegraphics[width=.14\linewidth]{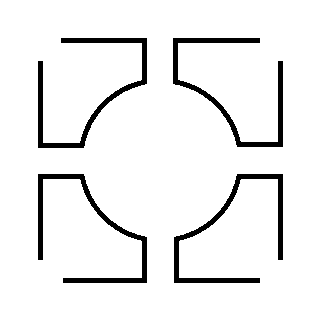}} &
        \fbox{\includegraphics[width=.14\linewidth]{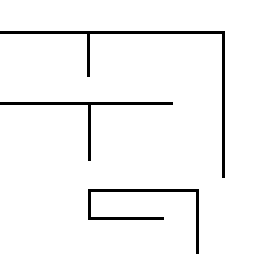}} &
        \fbox{\includegraphics[width=.14\linewidth]{figures/maps/eval_mowing_14.png}}
    \end{tabular}
    \begin{tabular}{p{1cm}cccccccccc}
        (d) &
        \fbox{\includegraphics[width=.085\linewidth]{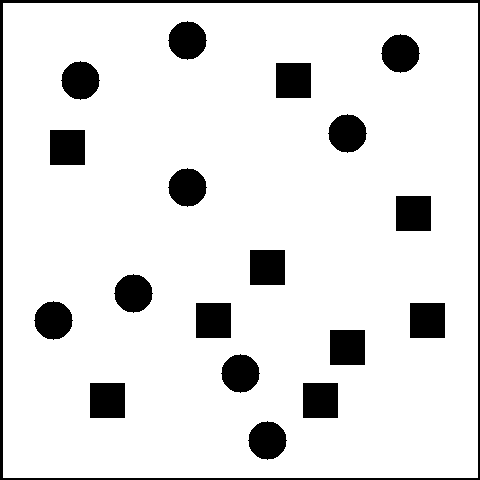}} &
        \fbox{\includegraphics[width=.085\linewidth]{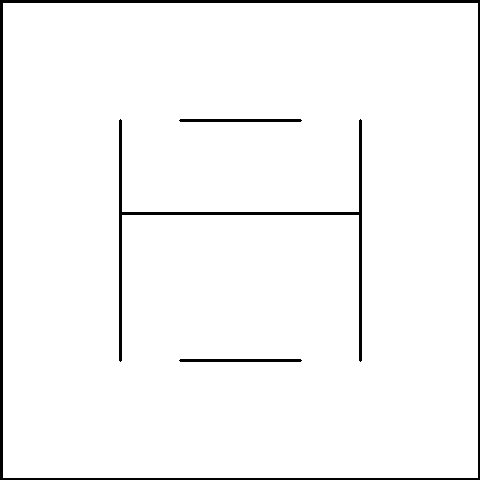}} &
        \fbox{\includegraphics[width=.085\linewidth]{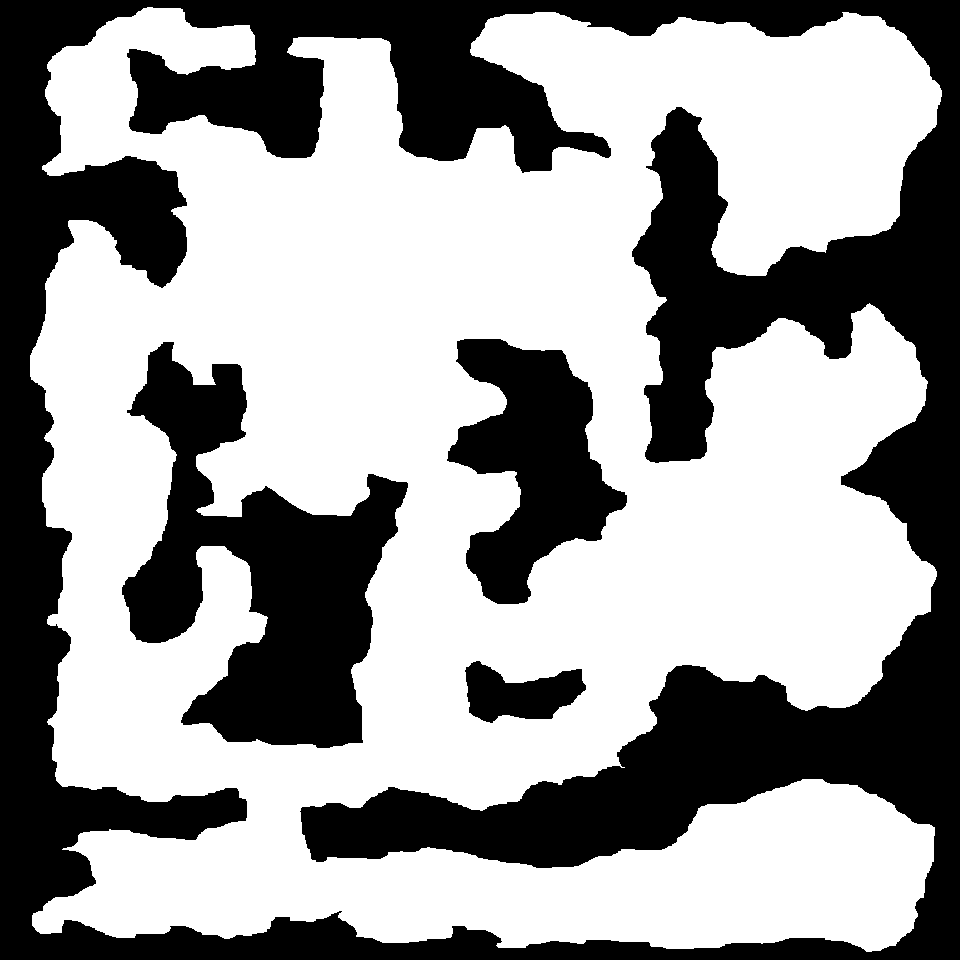}} &
        \fbox{\includegraphics[width=.085\linewidth]{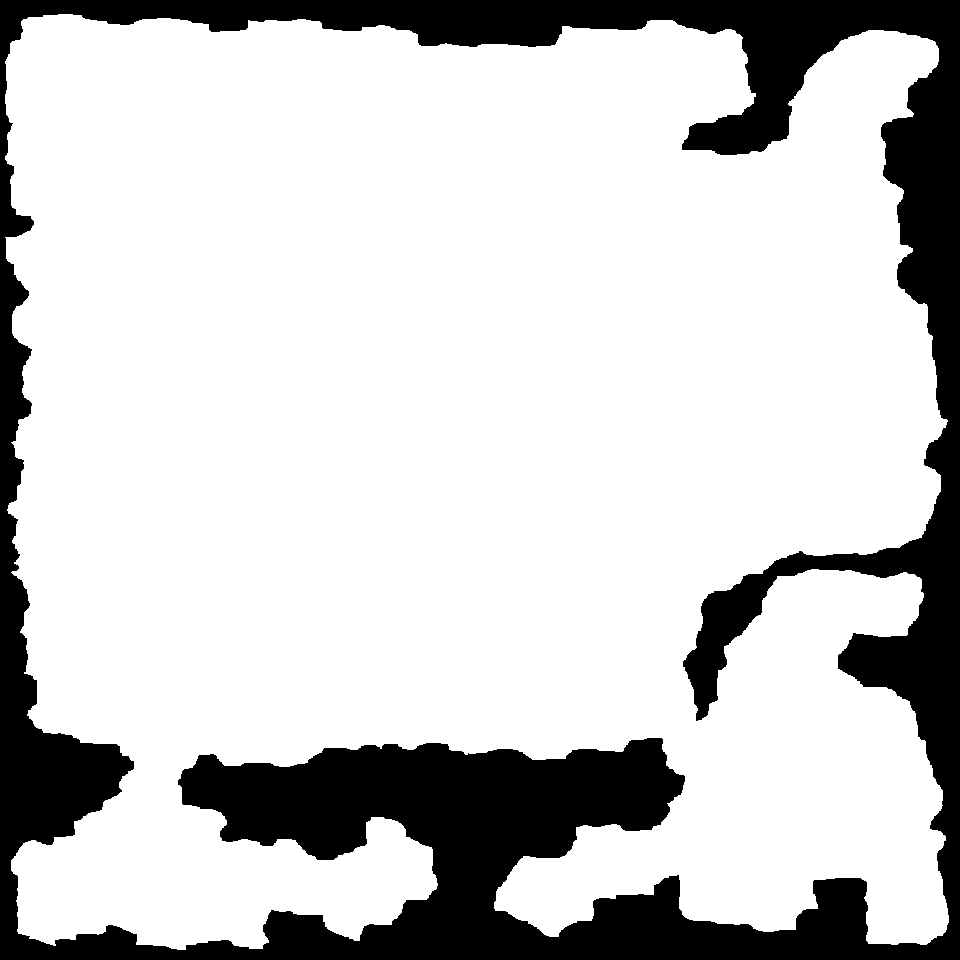}} &
        \fbox{\includegraphics[width=.085\linewidth]{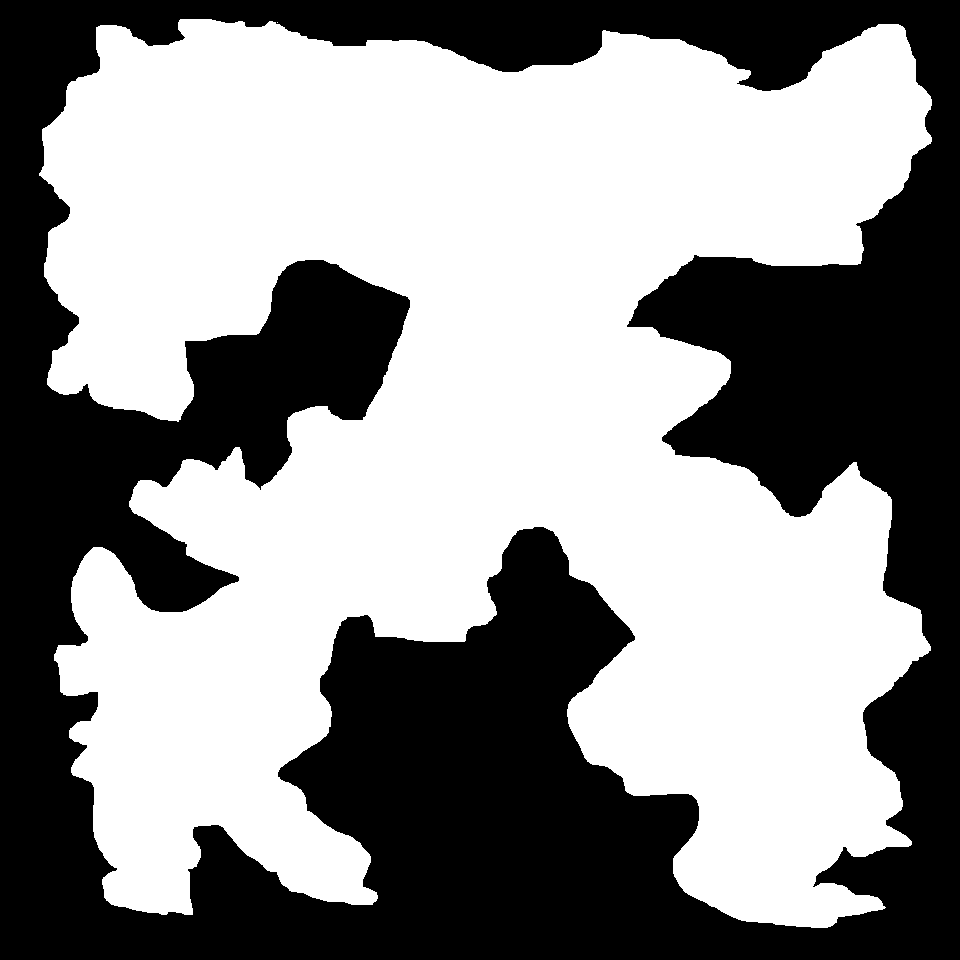}} &
        \fbox{\includegraphics[width=.085\linewidth]{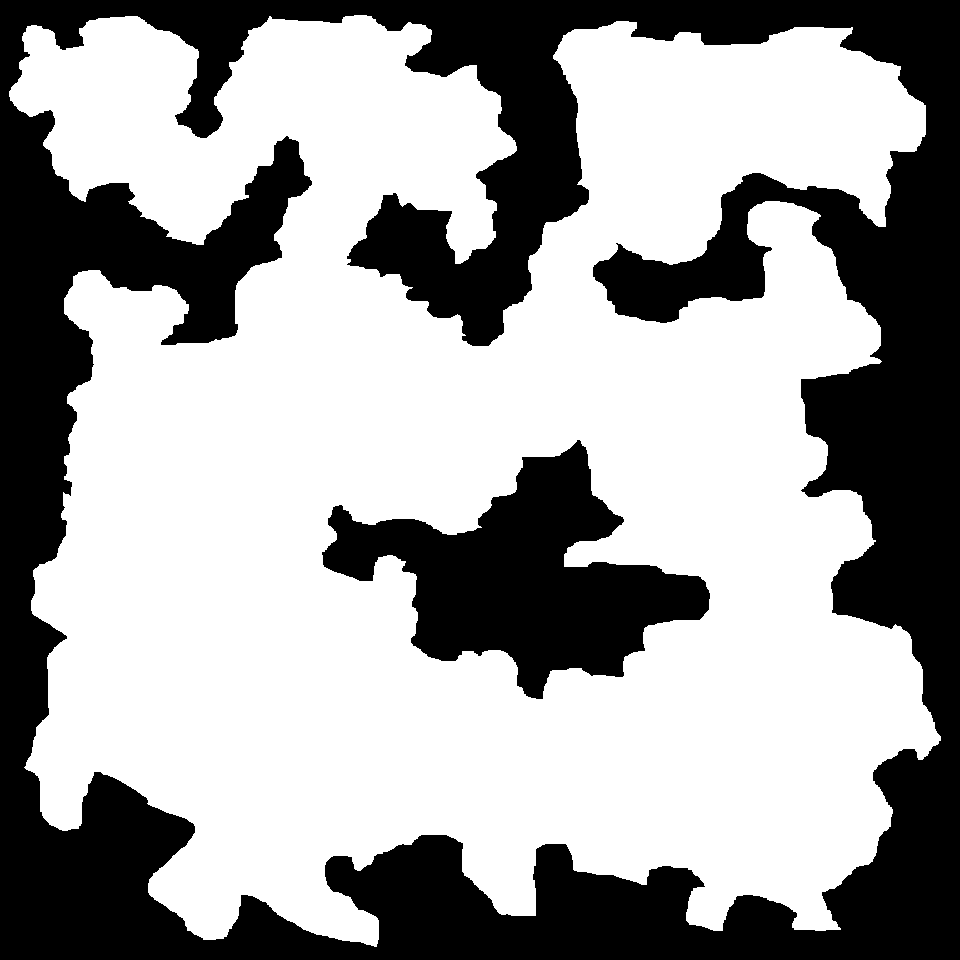}} &
        \fbox{\includegraphics[width=.085\linewidth]{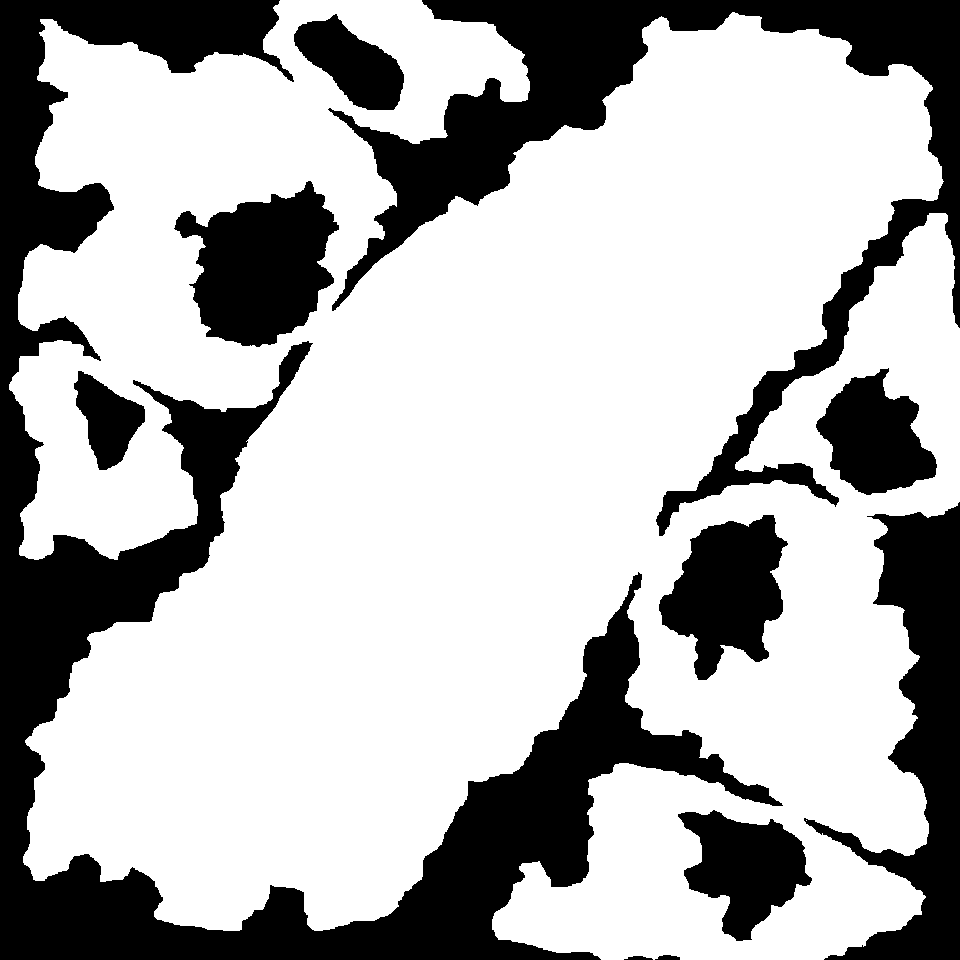}} &
        \fbox{\includegraphics[width=.085\linewidth]{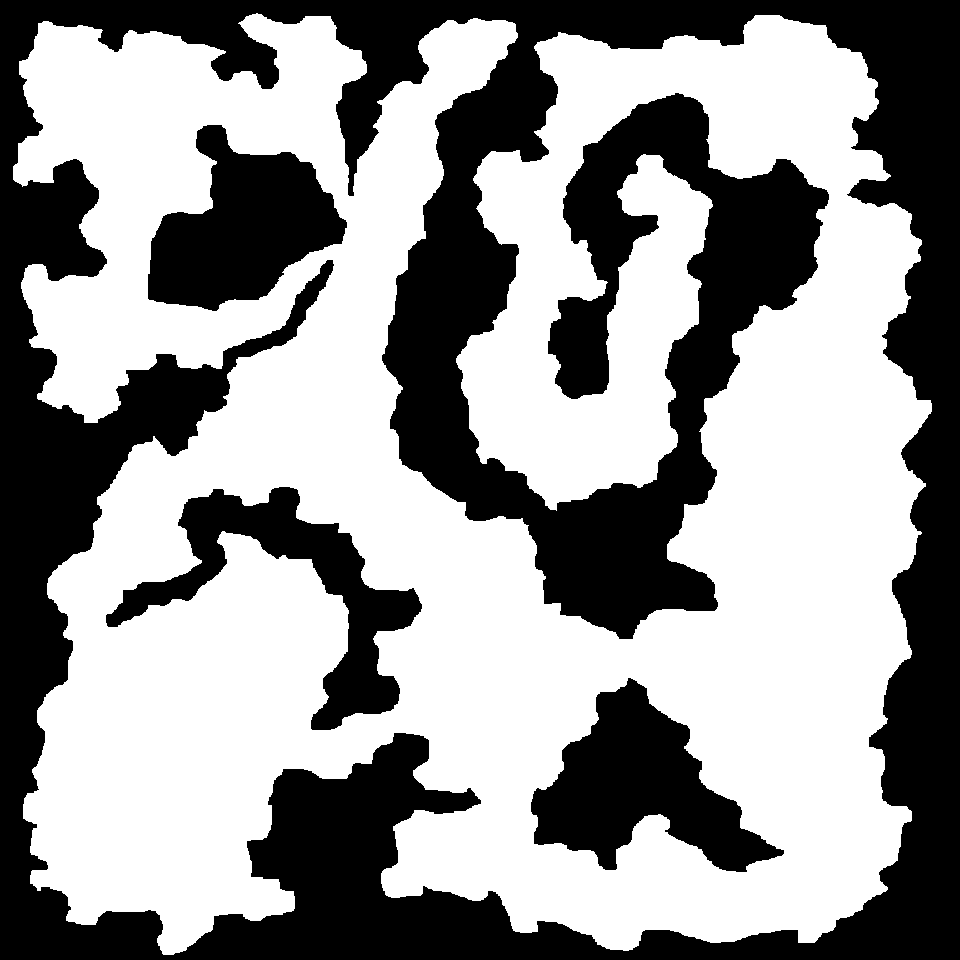}} &
        \fbox{\includegraphics[width=.085\linewidth]{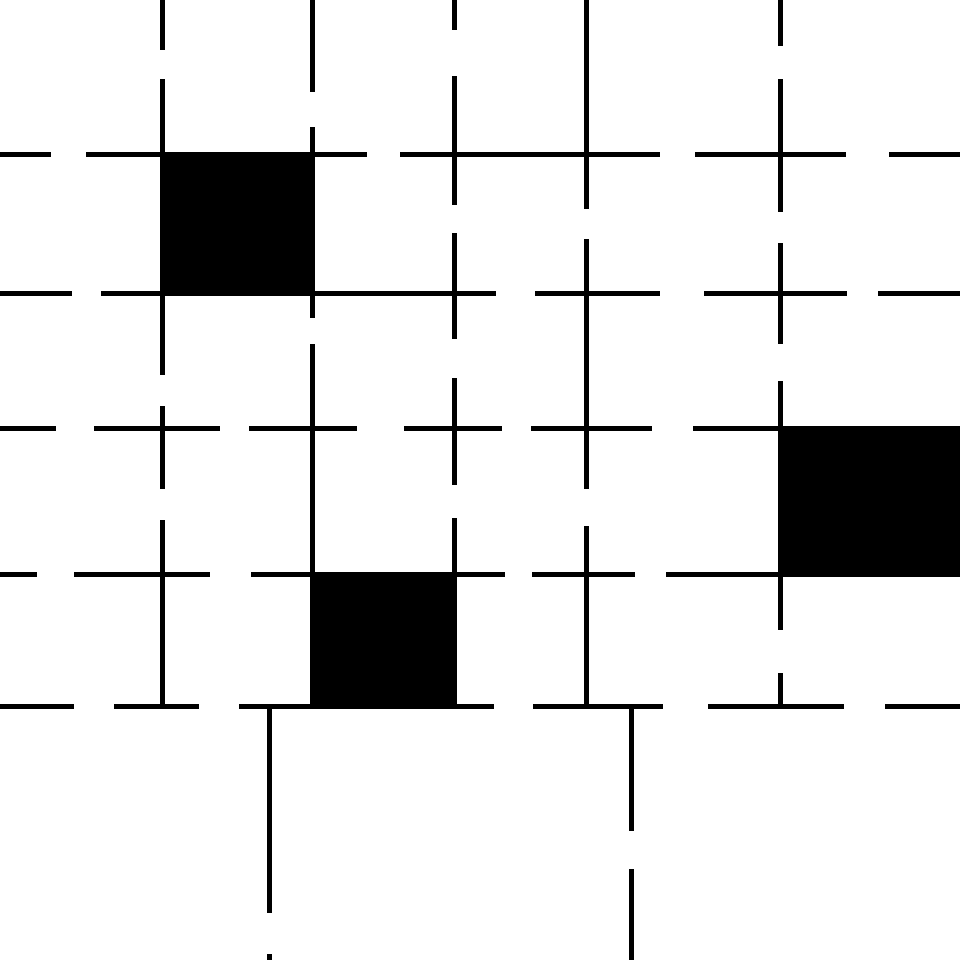}} &
        \fbox{\includegraphics[width=.085\linewidth]{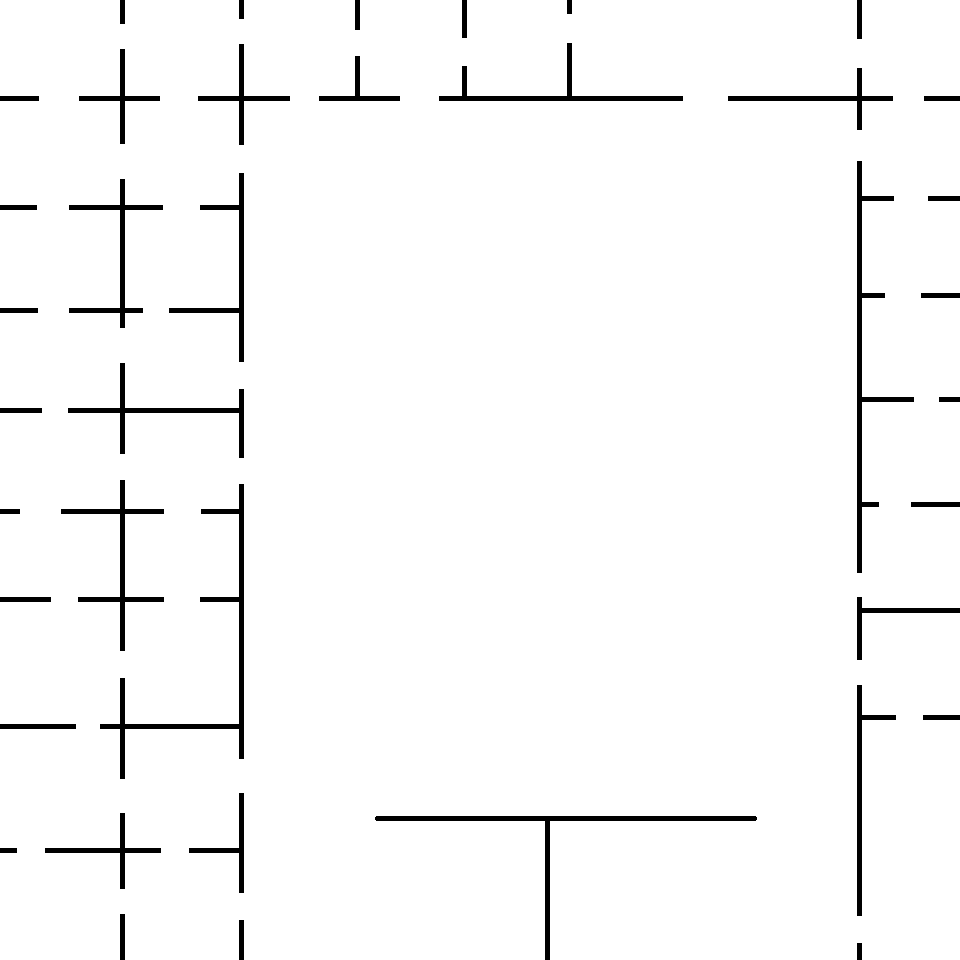}}
    \end{tabular} \\
    \begin{tabular}{p{1cm}cccccccc}
        (e) &
        \fbox{\includegraphics[width=.1\linewidth]{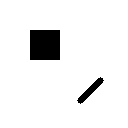}} &
        \fbox{\includegraphics[width=.1\linewidth]{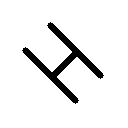}} &
        \fbox{\includegraphics[width=.1\linewidth]{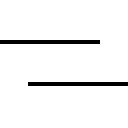}} &
        \fbox{\includegraphics[width=.1\linewidth]{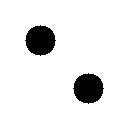}} &
        \fbox{\includegraphics[width=.1\linewidth]{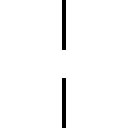}} &
        \fbox{\includegraphics[width=.1\linewidth]{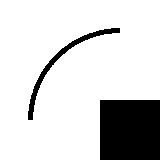}} &
        \fbox{\includegraphics[width=.1\linewidth]{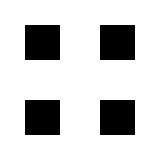}}
    \end{tabular}
    \caption{Evaluation maps used in simulation for (a) omnidirectional exploration, (b) non-omnidirectional exploration, and (c) lawn mowing. The maps in the last two rows were used additionally for ablations in (d) exploration and (e) lawn mowing.}
    \label{supp_fig_eval_maps_sim}
\end{figure*}

\begin{figure*}[!t]
    \centering
    \setlength{\tabcolsep}{0.5pt}
    \setlength{\fboxsep}{0pt}%
    \setlength{\fboxrule}{0.5pt}%
    \begin{tabular}{ccccccccccccc}
        \fbox{\includegraphics[width=.12\linewidth]{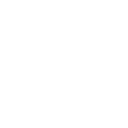}} &
        \fbox{\includegraphics[width=.12\linewidth]{figures/maps/eval_mowing_2.png}} &
        \fbox{\includegraphics[width=.12\linewidth]{figures/maps/eval_mowing_3.png}} &
        \fbox{\includegraphics[width=.12\linewidth]{figures/maps/eval_mowing_4.png}} &
        \fbox{\includegraphics[width=.12\linewidth]{figures/maps/eval_mowing_5.png}} &
        \fbox{\includegraphics[width=.12\linewidth]{figures/maps/eval_mowing_8.png}}
    \end{tabular} \\
    \caption{The evaluation maps used for the real-world experiments.}
    \label{supp_fig_eval_maps_real}
\end{figure*}

\section{\break Maps}
\label{supp_sec_maps}

In Fig. \ref{supp_fig_train_maps_sim}, \ref{supp_fig_train_maps_real}, \ref{supp_fig_eval_maps_sim}, and \ref{supp_fig_eval_maps_real}, we present the maps used in our experiments. Fig. \ref{supp_fig_train_maps_sim} and \ref{supp_fig_train_maps_real} show the training maps used in simulation and the real-world, respectively. Fig. \ref{supp_fig_eval_maps_sim} and \ref{supp_fig_eval_maps_real} show the evaluation maps used in simulation and the real-world, respectively. The maps are ordered from left to right in the same order as they appear in the respective tables in the main paper.

\section{\break Details on the Compared Methods}
\label{supp_sec_implementation_compared_methods}

\textbf{TSP-based solution with A*.} For the TSP baseline we subdivide the environment into square grid cells of comparable size to the agent, and apply a TSP solver to find the shortest path to visit the center of each cell, which are the nodes. In order to guarantee that each cell is fully covered by a circular agent, the side length for each grid cell is set to $\sqrt{2}r$, where $r$ is the agent radius. Note that this leads to overlap and thus increases the time to reach full coverage. In the offline case, we compute the weight between each pair of nodes using the shortest path algorithm, A* \cite{hart1968formal}. However, due to the size of our environments, the computation time was infeasible. Thus, we implemented a supremum heuristic for distant nodes, where they would be assigned the largest possible path length instead of running A*. This improved the runtime considerably, while not affecting the path length to a noticeable extent. For the online case we applied the TSP solver on all visible nodes, executed the path while observing new nodes, and repeated until all nodes were covered. In this case, the TSP execution time on an Epyc 7742 64-core CPU was included in the coverage time, as the path would need to be replanned online in a realistic setting. However, the execution time was relatively small compared to the time to navigate the path. We also tried replanning in shorter intervals, but the gain in time due to a decreased path length was smaller than the increase in computation time.

\textbf{Code repositories.} Table \ref{supp_table_github_links} lists the code implementations used for the methods under comparison. Note that for omnidirectional exploration, we evaluate our method under the same settings as in Explore-Bench, for which Xu \textit{et al.} \cite{xu2022explore} report the performance of the compared methods. We report these results in Table \ref{table_explore_bench}.

\begin{table}[h]
    \centering
    \caption{Links to code implementations used for the methods under comparison.}
    \begin{tabular}{l}
        \toprule
        \textbf{Omnidirectional exploration} \\
        Official Explore-Bench implementation \cite{xu2022explore}: \\
        \scriptsize{\url{https://github.com/efc-robot/Explore-Bench}} \\
        \midrule
        \textbf{Non-omnidirectional exploration} \\
        Official Frontier RL implementation \cite{hu2020tovt}: \\
        \scriptsize{\url{https://github.com/hanlinniu/turtlebot3_ddpg_collision_avoidance}} \\
        Frontier RL implementation in simulation: \\
        \tiny{\url{https://github.com/Peace1997/Voronoi_Based_Multi_Robot_Collaborate_Exploration_Unknow_Enviroment}} \\
        \midrule
        \textbf{Lawn mowing} \\
        Implementation of BSA \cite{gonzalez2005icra}: \\
        \scriptsize{\url{https://github.com/nobleo/full_coverage_path_planner}} \\
        \bottomrule
    \end{tabular}
    \label{supp_table_github_links}
\end{table}

\bibliography{egbib}
\bibliographystyle{IEEEtran}

\newpage

\begin{IEEEbiography}[{\includegraphics[width=1in,height=1.25in,clip,keepaspectratio]{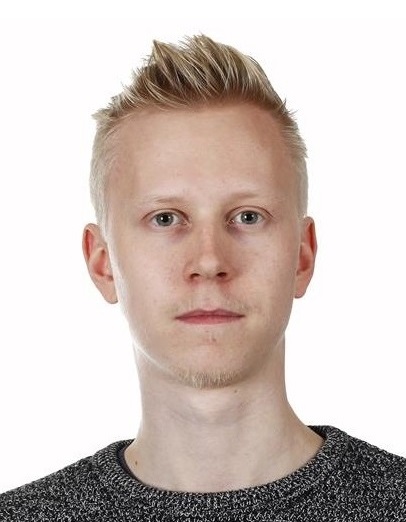}}]{Arvi Jonnarth} received the B.S. degree in mathematics in 2017, and the M.S. degree in engineering physics in 2018, both from
Uppsala University, Sweden. He received his Ph.D. degree in electrical engineering from Linköping University, Sweden, in 2024.

From 2018 to 2019, he was a Software Engineer with Husqvarna Group. Between 2024 and 2025 he was a Vision and Machine Learning Specialist with Husqvarna Group, Sweden. Currently, he is an Engineering Lead with Manta Systems, Sweden. His research interests include semantic segmentation, weakly-supervised learning, and robot vision.

\end{IEEEbiography}

\begin{IEEEbiography}[{\includegraphics[width=1in,height=1.25in,clip,keepaspectratio]{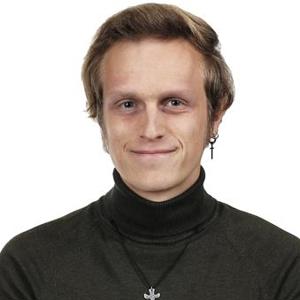}}]{Ola Johansson} received the B.S. degree in electrical engineering in 2020, and the M.S degree in systems control and robotics in 2022, both from KTH Royal Institute of Technology, Sweden. He is currently a Research Engineer at the Department of Electrical Engineering, Linköping University, Sweden. His research interests include localisation and motion planning in robotics.
\end{IEEEbiography}

\begin{IEEEbiography}[{\includegraphics[width=1in,height=1.25in,clip,keepaspectratio]{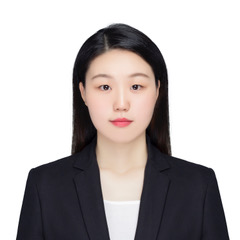}}]{Jie Zhao} received the B.E. degree in information management and information systems from Northwest A\&F University, China, in 2018, and received the Ph.D. degree in signal and information processing from Dalian University of Technology (DUT), China, in 2023. She is currently a postdoctoral with the school of information and communication engineering, DUT. Her research interests include visual object tracking, and robot vision.
\end{IEEEbiography}

\begin{IEEEbiography}[{\includegraphics[width=1in,height=1.25in,clip,keepaspectratio]{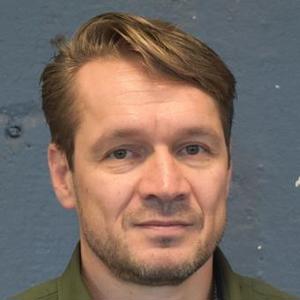}}]{Michael Felsberg} received the Ph.D. degree from Kiel University, Germany,
in 2002, and the docent degree from Linköping University, in 2005. He
is a full professor with Linköping University, Sweden, since 2008. He
received the DAGM Olympus award in 2005 and is fellow of the IAPR, 
ELLIS, and AAIA.

His research on visual learning theory includes video object and instance segmentation,
classification, segmentation, and registration of point clouds, as well
as efficient machine learning techniques for incremental, few-shot, and
long-tailed settings.
\end{IEEEbiography}

\EOD

\end{document}